\documentclass[sigconf]{acmart}

\copyrightyear{2020}
\acmYear{2020}
\setcopyright{acmcopyright}
\acmConference[MM '20] {28th ACM International Conference on Multimedia}{October 12--16, 2020}{Seattle, WA, USA.}
\acmBooktitle{28th ACM International Conference on Multimedia (MM '20), October 12--16, 2020, Seattle, WA, USA.}
\acmPrice{15.00}
\acmDOI{10.1145/3394171.3413820}
\acmISBN{978-1-4503-7988-5/20/10}
\usepackage{epstopdf}
\usepackage{multirow}
\begin{document}
\fancyhead{}
\title{DCSFN: Deep Cross-scale Fusion Network for Single Image Rain Removal}


%

\author{Cong Wang}
\authornote{Both authors contributed equally to this research.}
\affiliation{%
  \institution{Dalian University of Technology}}
\email{supercong94@gmail.com}

\author{Xiaoying Xing}
\authornotemark[1]
\affiliation{\institution{Tsinghua University}}
\email{xing-xy17@mails.tsinghua.edu.cn}

\author{Zhixun Su}
\affiliation{\institution{Dalian University of Technology}}
\email{zxsu@dlut.edu.cn}

\author{Junyang Chen}
\authornote{The corresponding author is Junyang Chen.}
\affiliation{\institution{University of Macau}}
\email{yb77403@umac.mo}
%
%
%
%
%
%

\def\etal{{\emph{et al}}}
\begin{abstract}
  Rain removal is an important but challenging computer vision task as rain streaks can severely degrade the visibility of images that may make other visions or multimedia tasks fail to work.
  Previous works mainly focused on feature extraction and processing or neural network structure, while the current rain removal methods can already achieve remarkable results, training based on single network structure without considering the cross-scale relationship may cause information drop-out.
  In this paper, we explore the cross-scale manner between networks and inner-scale fusion operation to solve the image rain removal task.
  Specifically, to learn features with different scales, we propose a multi-sub-networks structure, where these sub-networks are fused via a cross-scale manner by Gate Recurrent Unit to inner-learn and make full use of information at different scales in these sub-networks.
  Further, we design an inner-scale connection block to utilize the multi-scale information and features fusion way between different scales to improve rain representation ability and we introduce the dense block with skip connection to inner-connect these blocks.
  Experimental results on both synthetic and real-world datasets have demonstrated the superiority of our proposed method, which outperforms over the state-of-the-art methods.
  The source code will be available at \url{https://supercong94.wixsite.com/supercong94}.
\end{abstract}





\maketitle

\section{Introduction}
Images with rain captured by outdoor surveillance equipment significantly degrade the performance of some existing computer vision systems and also result in pool visual experience for some multimedia applications.
Hence, removing rain from a single image is important for these multimedia applications.
In this paper, we aim to solve the image deraining problem.

A rainy image $\mathbf{O}$ can be expressed as the linear sum between a background image $\mathbf{B}$ and a rain layer $\mathbf{R}$:
\begin{equation}
\mathbf{O} = \mathbf{B} + \mathbf{R},
\label{eq:rain model}
\end{equation}
Deraining is a highly ill-posed problem because there are numerous $\mathbf{B}$ and $\mathbf{R}$ pairs for a given rainy image.
To restrain the solution space, many priors about the background image or rain layer are proposed~\cite{derain_id_kang,derain_dsc_luo,derain_lowrank,derain_lp_li}.
Although these methods solve the deraining task to some extent, they always fail to work when the priors are invalid due to the simper assumption on shapes, directions, and sizes of rain.
Hence, a more effective deraining approach, which can deal with various rain in most conditions, is needed.

Benefiting from stronger feature representation ability, Convolution Neural Networks (CNNs, also called Deep Learning) has achieved great success in many computer vision problems, e.g., object detection~\cite{objectdetection_wang}, object tracking~\cite{tracking_zhang}, pose estimation~\cite{pyramid_Pose_Estimation}, optical flow~\cite{Pyramid_Ranjan_Optical_Flow}, semantic segmentation~\cite{semanticsegmentation_fcn}, deblurring~\cite{deblur_Mustaniemi}, dehazing~\cite{dehaze_mscnn_ren,dehaze_dehazenet_cai,dehaze_li_AOD,dehaze_zhang_dcdpn}, super-resolution~\cite{SISR_dong,SISR_cui,SISR_yu} and also deraining~\cite{derain_clearing_fu,derain_ddn_fu,derain_jorder_yang,derain_cgan_zhang,derain_rescan_li,derain_nledn_li,derain_zhang_did,derain_fu_deeptree,derain_Physics_gan,dual_cnn,derain_GRN,derain_2019_CVPR_spa,derain_prenet_Ren_2019_CVPR,derain_Heavy_Li_2019_CVPR,derain_pami_yang,light_weight}.
These deraining algorithms consider different manners to remove rain.
Researchers design various deep network structures to solve the image deraining problem.
To improve the computational efficiency and decrease model sizes, light-weight networks~\cite{light_weight,derain_GRN} came out in turn.
By learning rain mask or estimating rain density to guide the deraining process, rain detection and removal~\cite{derain_jorder_yang} and multi-stream network~\cite{derain_zhang_did} are proposed and present a better deraining performance.
And other various network modules are developed by designing kinds of mechanisms, e.g., channel attention by squeeze-and-excitation~\cite{se}, pixel-wise attention using nonlocal mean network~\cite{nonlocalnetwork_wang} and adversarial learning~\cite{derain_cgan_zhang,derain_Heavy_Li_2019_CVPR,derain_Physics_gan}.
Although these deraining methods have achieved better performance, they still neglect some important details.
Previous works mainly focused on feature extraction and processing or neural network structure, while the current rain removal methods can already achieve remarkable results, training based on single network structure without considering the cross-scale relationship may cause information drop-out.

To address this problem, we propose a deep network with cross-scale learning architecture to remove rain from a single image.
Firstly, we design an inner-scale connection block to fuse features with different scales by building the correlation between each scales so that better learn the rain features.
Secondly, to maximize information flow and enable the computation of long-range spatial dependencies as well as efficient usage of the feature activation of proceeding layers, we introduce encoder and decoder with dense connection structure and skip connection to inner-connect these blocks.
Lastly, we propose a cross-scale fusion network to learn features at different scales, where the proposed cross-scale manner connects features at different scales by Gate Recurrent Unit (GRU) which makes full use of information at different scales.

In summary, this paper has the following contributions:
\begin{itemize}
\item We propose an inner-scale connection block by building the correlation between each scale so that it better learns the rain features.
\item We introduce encoder and decoder with dense connection structure and skip connection to inner-connect these blocks to improve the deraining performance.
\item We propose a cross-scale fusion network to learn features at different scales, where the proposed cross-scale manner connects features at different scales by Gate Recurrent Unit that makes full use of information at different scales.
\item Experimental results on both synthetic and real-world datasets have demonstrated the superiority of our proposed method, which outperforms over the state-of-the-art methods.
\end{itemize}

\section{Related Work}
In this section, we review some image deraining methods and the multi-scale learning.

\subsection{Single Image Deraining}
Single image rain removal can be more difficult compared to video-based methods~\cite{video_derain_chen_cvpr18,video_derain_cvpr18_li,video_derain_cvpr18_liu,video_derain_Yang_2019_CVPR}.
Video-based deraining can be promoted by extracting the features from serial frames that contain correlated information, while single image deraining is appropriate for an independent single input so that it is more difficult to restore the background image without frame information.
In this section, we only review the image deraining problem.
\newline
\textbf{Prior-Based Methods:} Many early works attempt to solve the problem with image priors~\cite{derain_id_kang,derain_dsc_luo,derain_lp_li,derain_lowrank,derain_nonlocalfilter_kim,derain_zhang_Sparse_and_Low-Rank}.
Considering that rain streaks dominant in high-frequency structure, Kang~\etal.~\cite{derain_id_kang} decompose the input image into high frequency and low-frequency layers, removing the high-frequency rain streaks by dictionary learning.
Based on the observation that rain is sparse, Luo~\etal.~\cite{derain_dsc_luo} propose a discriminative sparse coding structure to separate the rain layer from a clean image.
Moreover, since rain in a local patch is low-rank, some methods restrain the solution space from this point.
Chen~\etal.~\cite{derain_lowrank} propose a low-rank representation-based method which promotes the deraining performance by taking advantage of the low-rank model.
In \cite{derain_nonlocalfilter_kim}, they apply kernel regression to the deraining framework by using a non-local mean filter.
Wang~\etal.~\cite{derain_Wang_Hierarchical} propose a hierarchical structure for single image deraining and desnowing.
Zhu~\etal.~\cite{derain_zhu_bilayer} ingeniously consider rain direction and propose a joint optimization process.
\newline
\textbf{Deep learning-Based Methods:}
Deep learning-based methods achieve considerably better performance than prior-based methods, which verifies the efficiency of deep learning on computer vision tasks.
Fu~\etal.~\cite{derain_ddn_fu,derain_clearing_fu} first apply deep learning to single image deraining.
They manage to extract high-frequency structures via a guide filter and use a residual network to obtain the rain layer.
The learned rain layer is used to obtain a clean image via Eq.1.
Yang~\etal.~\cite{derain_jorder_yang} consider the hazy condition in some rainy situations and apply a dehazing-deraining-dehazing algorithm. They propose joint rain detection and removal method to deal with some complex situations.
Li~\etal.~\cite{derain_nledn_li} design a multi-scale non-local enhanced encoder-decoder network, which uses the pixel-wise attention mechanism to learn the residual rain.
Considering the guide role, Fan~\etal.~\cite{derain_GRN} design a light-weight residual-guide network for deraining in a stage-wise manner.
Li~\etal.~\cite{derain_rescan_li} believe spatial contextual information is important for single image deraining that they propose dilated convolution to capture more contextual information and remodel the rainy model by utilizing the squeeze-and-excitation operation.
Zhang~\etal.~\cite{derain_zhang_did} realize that some approaches have the performance of excessive deraining or residual rain.
Based on this, they propose a multi-stream densely connected convolutional neural network and estimate rain density to guide the deraining procedure.
Ren~\etal.~\cite{derain_prenet_Ren_2019_CVPR} present a better baseline model by investigating the input, output, and loss function
focus on network architecture and achieve better results than previous works.
Wang~\etal.~\cite{derain_2019_CVPR_spa} design a spatial attentive network to remove rain streaks in a local-to-global manner.
Li~\etal.~\cite{derain_Comprehensive_Benchmark_Li_2019_CVPR} study and evaluate existing single image deraining methods elaborately.
They develop a new large-scale benchmark consisted of both synthetic and real-world rainy images by considering diverse possible situations.
\subsection{Multi-scale Learning}
\begin{figure*}[!t]
\begin{center}
\begin{tabular}{c}
\includegraphics[width = 0.99\linewidth]{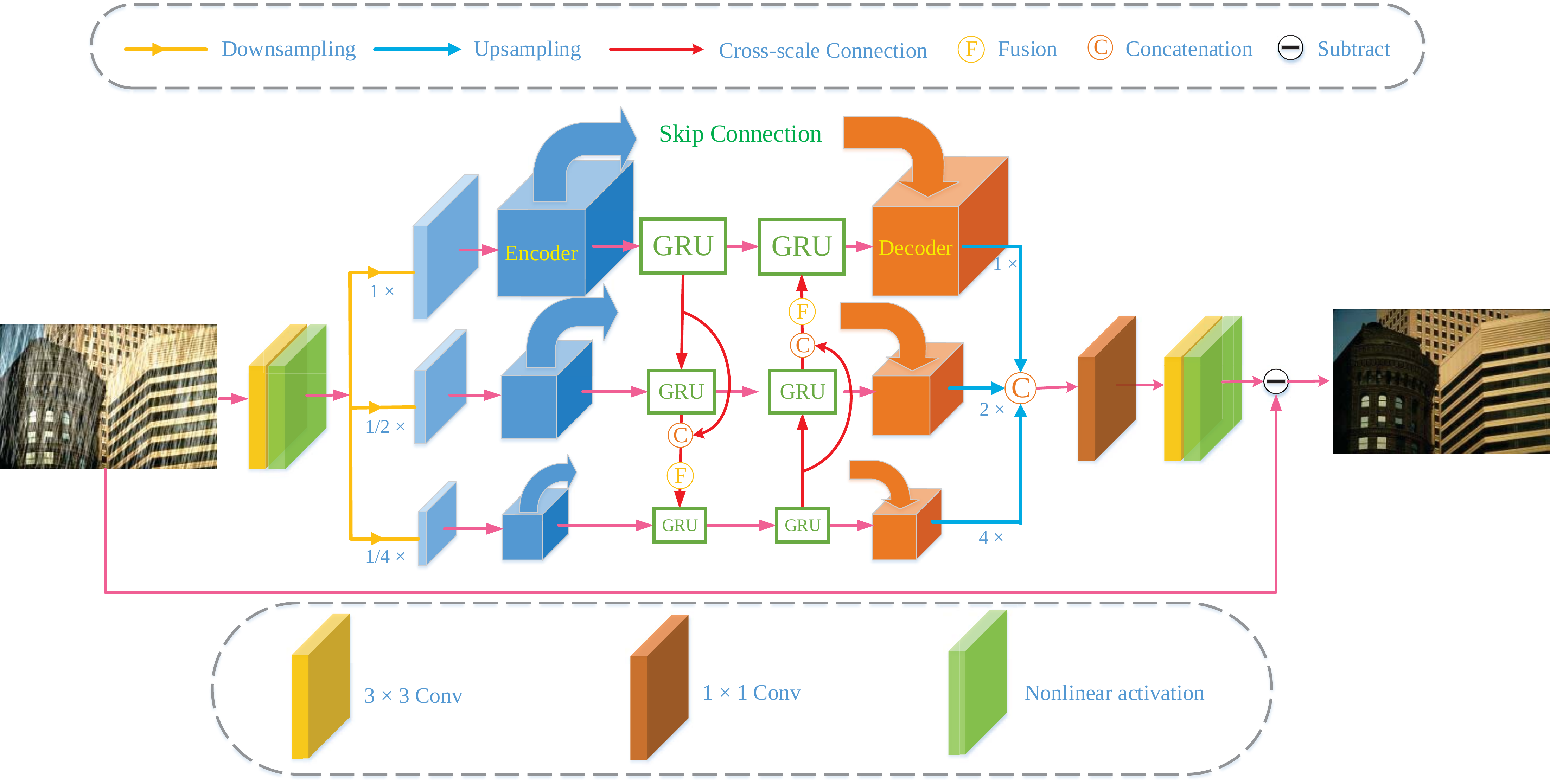}
\end{tabular}
\end{center}
\caption{The proposed network.
The network consists of three sub-networks with different scales.
At each scale, they have the same structure with densely connected encoder and decoder, shown in Fig.~\ref{fig: encoder-decoder}.
At the end of each encoder, features at different scales are fused by GRU via a cross-scale manner to make full use of information at
different scales.
At last, all features at different scales are fused to generate estimated rain and we gain the final estimated rain-free images via Eq.~\ref{eq:rain model}.
}
\label{fig: Overall Framework}
\end{figure*}
The hierarchy architecture can generate features of different scales that contain abundant layer information.
Moreover, exploiting the inner correlations of multi-scale features can lead to a deeper insight into image layout and boost the feature extraction performance to a great extent.
A representative architecture on multi-scale learning is proposed in~\cite{feature_pyramid_objectdetection}, which is a top-down convolutional network structure that is applied lateral connections for extracting multi-scale semantic features for the object detection task.
Inspired by the considerable performance on feature map construction, multi-scale learning had been applied on face image processing~\cite{multi-scale-face}, super-resolution~\cite{pyramid_superresolution}, and deraining~\cite{light_weight}\cite{derain-acmmm19-rehen} as well.
Yang~\etal.~\cite{derain-acmmm19-rehen} propose a recurrent hierarchy enhancement network considering the correlations between neighboring stages, where features extracted from the previous stage are transmitted to the later stage as guidance.
Although some works consider the correlations of different convolutional layers, the cross-scale information is still ignored and it should be explored.
\section{Proposed method}

\begin{figure*}[!t]
\begin{center}
\begin{tabular}{cc}
\includegraphics[width = 0.4\linewidth]{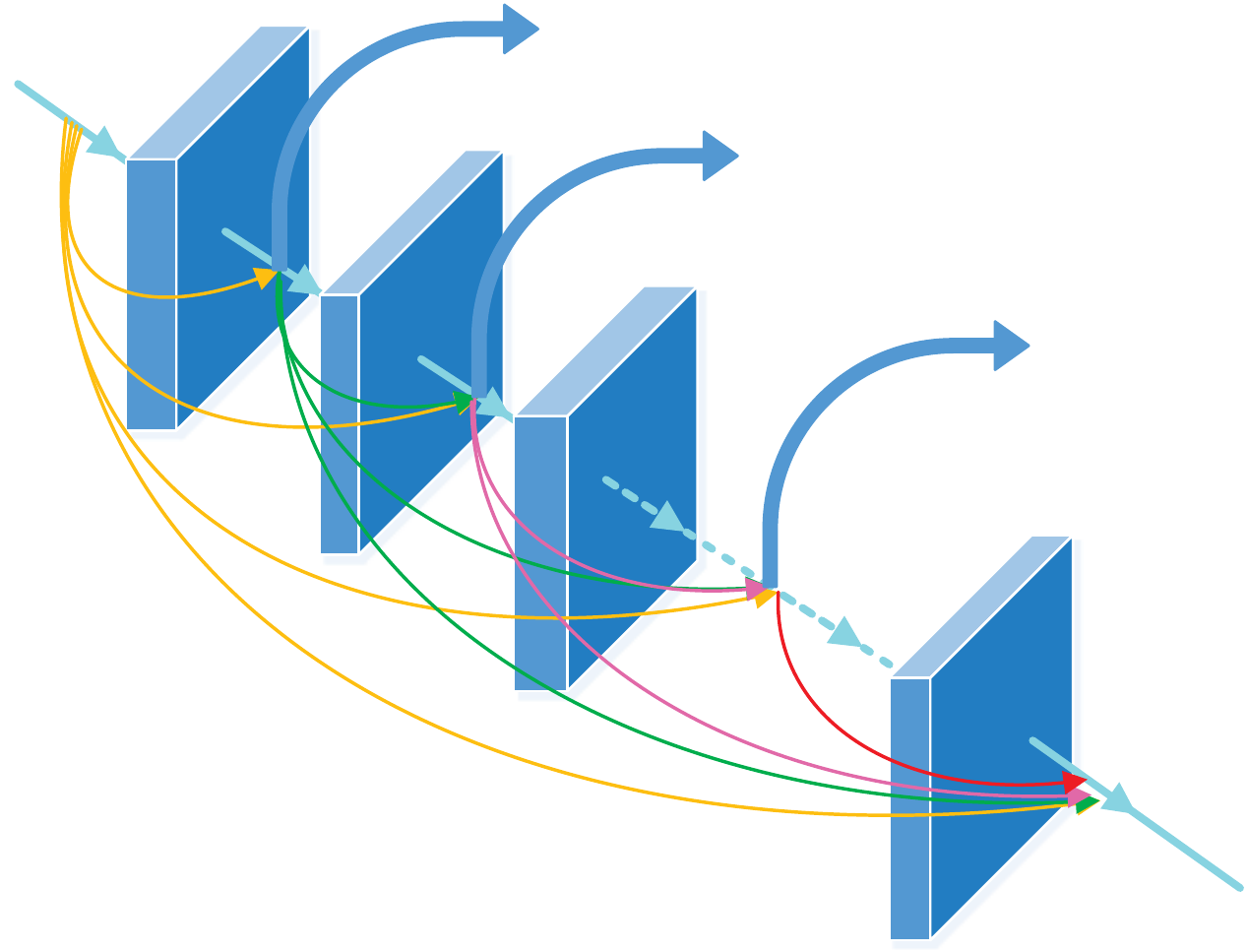}&\hspace{-4mm}
 \includegraphics[width = 0.4\linewidth]{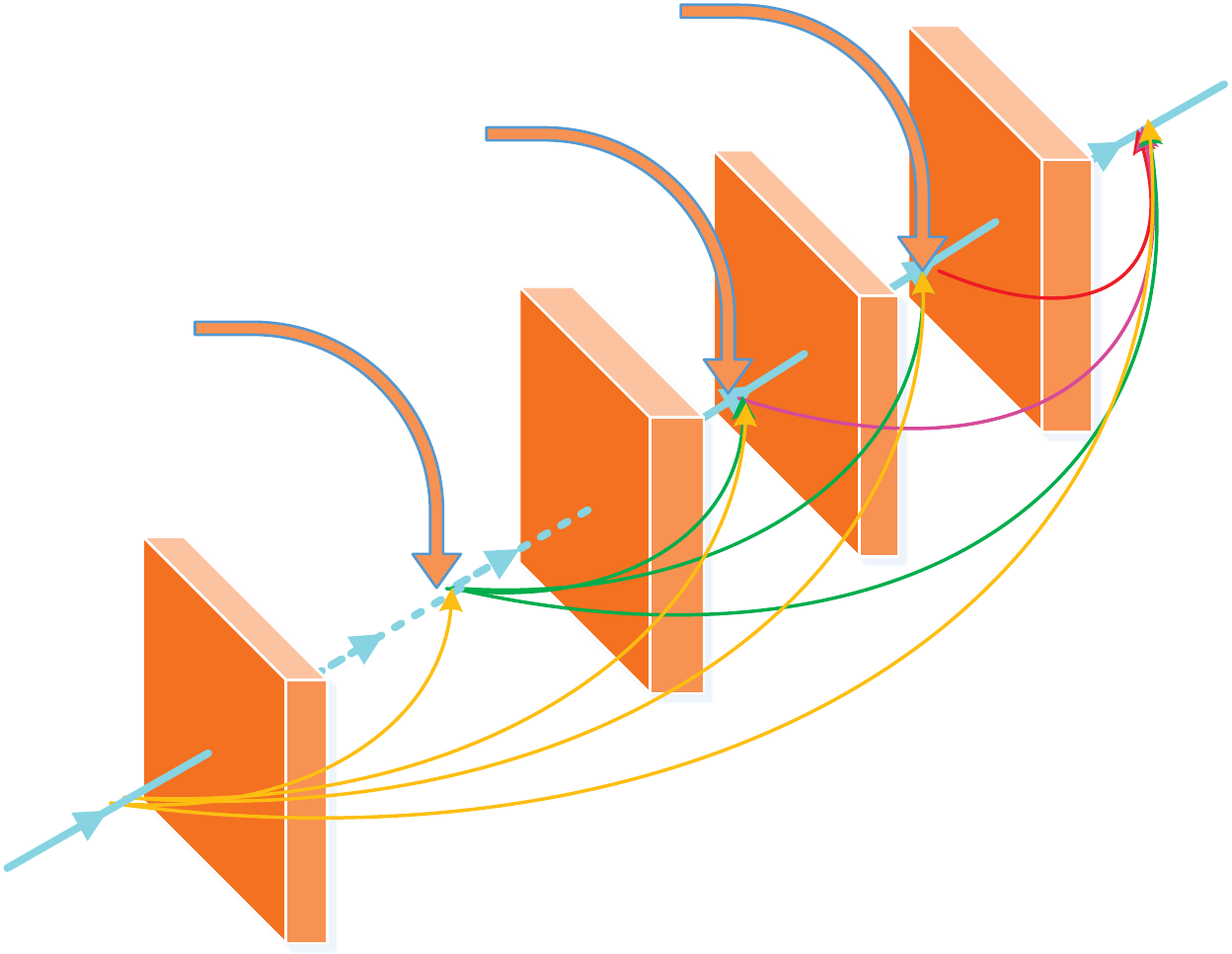}
\\
(a) Encoder &\hspace{-4mm} (b) Decoder
\\
\end{tabular}
\end{center}
\caption{(a) Encoder.
The encoder consists of a series of Inner-scale Connection Block (as illustrated in Fig.~\ref{fig: pyramid}) with dense connection.
(b) Decoder.
The Decoder has the same structure with encoder and skip connection is utilized to enable the computation of long-range spatial dependencies as well as efficient usage of the feature activation of proceeding layers.}
\label{fig: encoder-decoder}
\end{figure*}
We detail the proposed network in Sec.~\ref{sec:Proposed Network} and introduce the encoder and decoder in Sec.~\ref{sec:Encoder and Decoder with Dense Connection}.
The proposed Inner-scale Connection Block and applied loss function are described in Sec.~\ref{sec:Inner-scale Connection Block} and Sec.~\ref{sec: Loss Function}, respectively.
\subsection{Proposed Network (DCSFN)}
\label{sec:Proposed Network}
Our proposed network is illustrated in Fig.~\ref{fig: Overall Framework}.
The network consists of three subnetworks to learn information with different scales and they are fused via a cross-scale manner by Gate Recurrent Unit (GRU) at the end of each encoder to make full use of information at different scales.
We obtain the estimated rain by fusing features with different scales in these sub-networks and generate the final estimated rain-free image via Eq.~\ref{eq:rain model}.

\subsection{Encoder and Decoder with Dense Connection}
\label{sec:Encoder and Decoder with Dense Connection}
Dense connection is proposed in \cite{Densenetwork_huang}, which has achieved great success in many computer vision problems.
The dense connection structure alleviates the vanishing-gradient problem, strengthens feature propagation, encourages feature reuse.
One big advantage of dense connection is that it maximizes information and gradients flow throughout the network that makes it easy to train.
Moreover, as each layer has direct access to the gradients from the loss function and the original input signal, it leads to implicit deep supervision for these layers.
In this paper, we design densely connected encoder and decoder with skip connection as the style of our network.

Mathematically, the encoder can be expressed as:
\begin{equation}
\mathbf{F}_{l}^{E} = \mathcal{T}\Big (\mathcal{F}(\mathcal{C}[\mathbf{F}_{0}^{E}, \mathbf{F}_{1}^{E}, \cdots, \mathbf{F}_{l-1}^{E}])\Big),~~~l = 1, 2, \cdots, L,
\end{equation}
where $\mathbf{F}_{0}^{E}$ denotes the input features and $\mathbf{F}_{l}^{E}$ refers to the output of $l^{th}$ layer at the encoder stage.
$\mathcal{C}[\cdot]$ and $\mathcal{F}(\cdot)$ denotes the concatenation at channel dimension and fusion operation, respectively.
We use $1 \times 1$ convolution as the fusion operation.
$\mathcal{T}\Big (\cdot \Big )$ refers to the proposed inner-scale connection block detailed in Sec.~\ref{sec:Inner-scale Connection Block}.

In the decoder stage, skip connection between Encoder and Decoder is utilized to enable the computation of long-range spatial dependencies as well as efficient usage of the feature activation of proceeding layers:
\begin{equation}
\mathbf{F}_{l}^{D} = \mathcal{T}\Big (\mathcal{F}(\mathcal{C}[\mathbf{F}_{0}^{D}, \mathbf{F}_{1}^{D}, \cdots, \mathbf{F}_{l-1}^{D}])\Big) + \mathbf{F}_{l}^{E},~~~l = 1, 2, \cdots, L,
\end{equation}
where $\mathbf{F}_{0}^{D}$ denotes the GRU output of the current scale and $\mathbf{F}_{l}^{D}$ refers to the output of $l^{th}$ layer at the decoder stage.

\subsection{Inner-scale Connection Block}
\label{sec:Inner-scale Connection Block}
\begin{figure}[!h]
\begin{center}
\begin{tabular}{c}
\includegraphics[width = 0.99\linewidth]{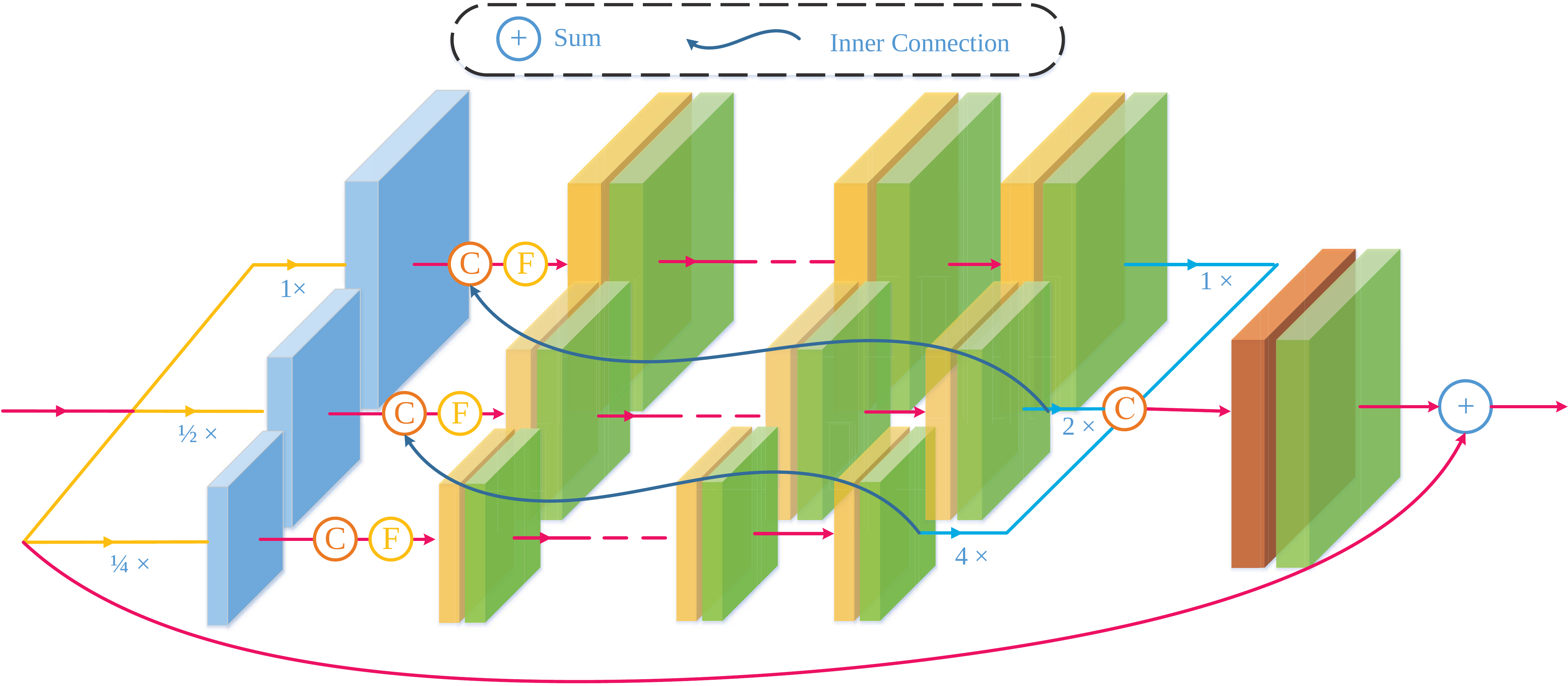}
\end{tabular}
\end{center}
\caption{Inner-scale Connection Block.
Firstly, \emph{Global Max Pooling} is utilized to obtain features with different scales.
Secondly, features after several convolutions are connected between different scales to boost the correlation information exploration.
Lastly, all features at different scales are fused to learn main features.}
\label{fig: pyramid}
\end{figure}

Multi-scale learning has been applied to a number of vision problems and obtains huge development.
Although various multi-scale manners are proposed and designed, the inner correlation between different scales never is explored.
In this paper, we explore the inner correlation between different scales by designing the proposed inner-scale connection block.
The proposed block is shown in Fig.~\ref{fig: pyramid}.

Firstly, we utilize \emph{Global Max Pooling} operation to obtain multi-scale features:
\begin{equation}
\mathbf{P}_{\frac{1}{k}} = \mathcal{P}_{k}(x),~~~k = 1, 2, \cdots, 2^{K-1},
\end{equation}
where $x$ is the input feature.
$K$ is the maximum scales we used.
$\mathcal{P}_{k}(\cdot)$ is \emph{Global Max Pooling} operation with $k \times k$ kernels and $k \times k$ strides.
$\mathbf{P}_{\frac{1}{k}}$ is the output features at $\frac{1}{k}$ scale.

Then, features at different scales are connected via inner way:
\begin{equation}
\tilde{\mathbf{P}}_{\frac{1}{k}} = \mathcal{F}\Big (\mathcal{C}[\tilde{\mathbf{P}}_{\frac{1}{2 \times k}}, \mathcal{U}_{2}(\mathbf{Y}_{\frac{1}{k}})]\Big ).
\end{equation}
Here, $\tilde{\mathbf{P}}_{\frac{1}{k}} = \mathbf{P}_{\frac{1}{k}}$ when $k = 1$.
$\mathbf{Y}_{\frac{1}{k}} = \mathcal{H}_{n}(\tilde{\mathbf{P}}_{\frac{1}{k}})$.
$\mathcal{H}_{n}(\cdot)$ denotes $3 \times 3$ convolution layer followed a nonlinear activation with $n$ groups.
$\mathcal{U}_{k}$ denotes the Up-sampling operation with $k \times$ factor.

At last, the features at different scales are fused and the residual learning is introduced:
\begin{equation}
\mathbf{Z} = \mathcal{F}\Big (\mathcal{C}[\mathbf{Y}_{1}, \cdots, \mathcal{U}_{2^{K-1}}(\mathbf{Y}_{\frac{1}{2^{K-1}}})]\Big ) + x.
\end{equation}
Here, $\mathbf{Z}$ is the output of our proposed Inner-scale Connection Block.
\subsection{Loss Function}
\label{sec: Loss Function}
SSIM loss has been proved to the best loss function in the deraining task~\cite{derain_prenet_Ren_2019_CVPR}.
In the network, we use it as our error loss:
\begin{equation}
\mathcal{L} = -SSIM(\hat{\mathbf{B}}, \mathbf{B}).
\label{eq:Loss}
\end{equation}
where $\hat{\mathbf{B}}$ and $\mathbf{B}$ are the de-rained image and the corresponding ground-truth background.

\section{Experiments Results}

\begin{table*}[!t]
	\caption{Quantitative experiments evaluated on three synthetic datasets.
		The best results are highlighted in boldface.}
	\centering
	\scalebox{0.99}{
		\begin{tabular}{ccccccccccccccccc}
			\toprule
			& \multicolumn{2}{c}{DDN~\cite{derain_ddn_fu}}                   & \multicolumn{2}{c}{RESCAN~\cite{derain_rescan_li}}    & \multicolumn{2}{c}{NLEDN~\cite{derain_nledn_li}}& \multicolumn{2}{c}{REHEN~\cite{derain-acmmm19-rehen}} & \multicolumn{2}{c}{PreNet~\cite{derain_prenet_Ren_2019_CVPR}}   & \multicolumn{2}{c}{SSIR~\cite{Derain-cvpr19-semi}}             & \multicolumn{2}{c}{Ours}      \\
			& \multicolumn{2}{c}{CVPR'17}         & \multicolumn{2}{c}{ECCV'18}    & \multicolumn{2}{c}{ACM MM'18}& \multicolumn{2}{c}{ACM MM'19}   & \multicolumn{2}{c}{CVPR'19}             &\multicolumn{2}{c}{CVPR'19}             & \multicolumn{2}{c}{}
			\\
			\midrule
			Dataset       & PSNR& SSIM & PSNR&SSIM & PSNR&SSIM & PSNR & SSIM& PSNR & SSIM  & PSNR & SSIM & PSNR & SSIM     \\
			\midrule
			Rain100H   &22.26&0.69    &25.92&0.84 & 28.42&0.88 & 27.52 & 0.86& 27.89&0.89  & 22.47 & 0.71& \textbf{28.81}&\textbf{0.90} \\
			\midrule
			Rain100L   &34.85&0.95    &36.12&0.97& 38.84&0.98 & 37.91 & 0.98&36.69&0.98    & 32.37 &0.92 &\textbf{38.96}&\textbf{0.99} \\
			\midrule
			Rain1200  &30.95& 0.86   &32.35&0.89 & 32.98&0.92 & 32.51 & 0.91&32.38&0.92 &29.32& 0.89 &\textbf{33.19}& \textbf{0.93} \\
			\bottomrule
	\end{tabular}}
	\label{tab: the results in synthetic datasets}
\end{table*}

\begin{figure*}[!t]
\begin{center}
\begin{tabular}{ccccccccc}
\includegraphics[width = 0.11\linewidth]{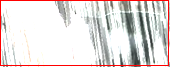} &\hspace{-4mm}
\includegraphics[width = 0.11\linewidth]{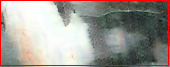} &\hspace{-4mm}
\includegraphics[width = 0.11\linewidth]{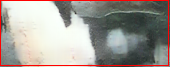} &\hspace{-4mm}
\includegraphics[width = 0.11\linewidth]{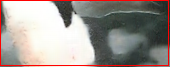} &\hspace{-4mm}
\includegraphics[width = 0.11\linewidth]{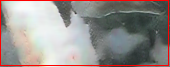} &\hspace{-4mm}
\includegraphics[width = 0.11\linewidth]{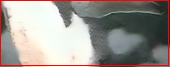} &\hspace{-4mm}
\includegraphics[width = 0.11\linewidth]{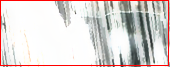} &\hspace{-4mm}
\includegraphics[width = 0.11\linewidth]{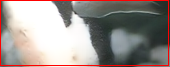} &\hspace{-4mm}
\includegraphics[width = 0.11\linewidth]{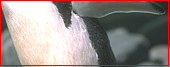}
\\
\includegraphics[width = 0.11\linewidth]{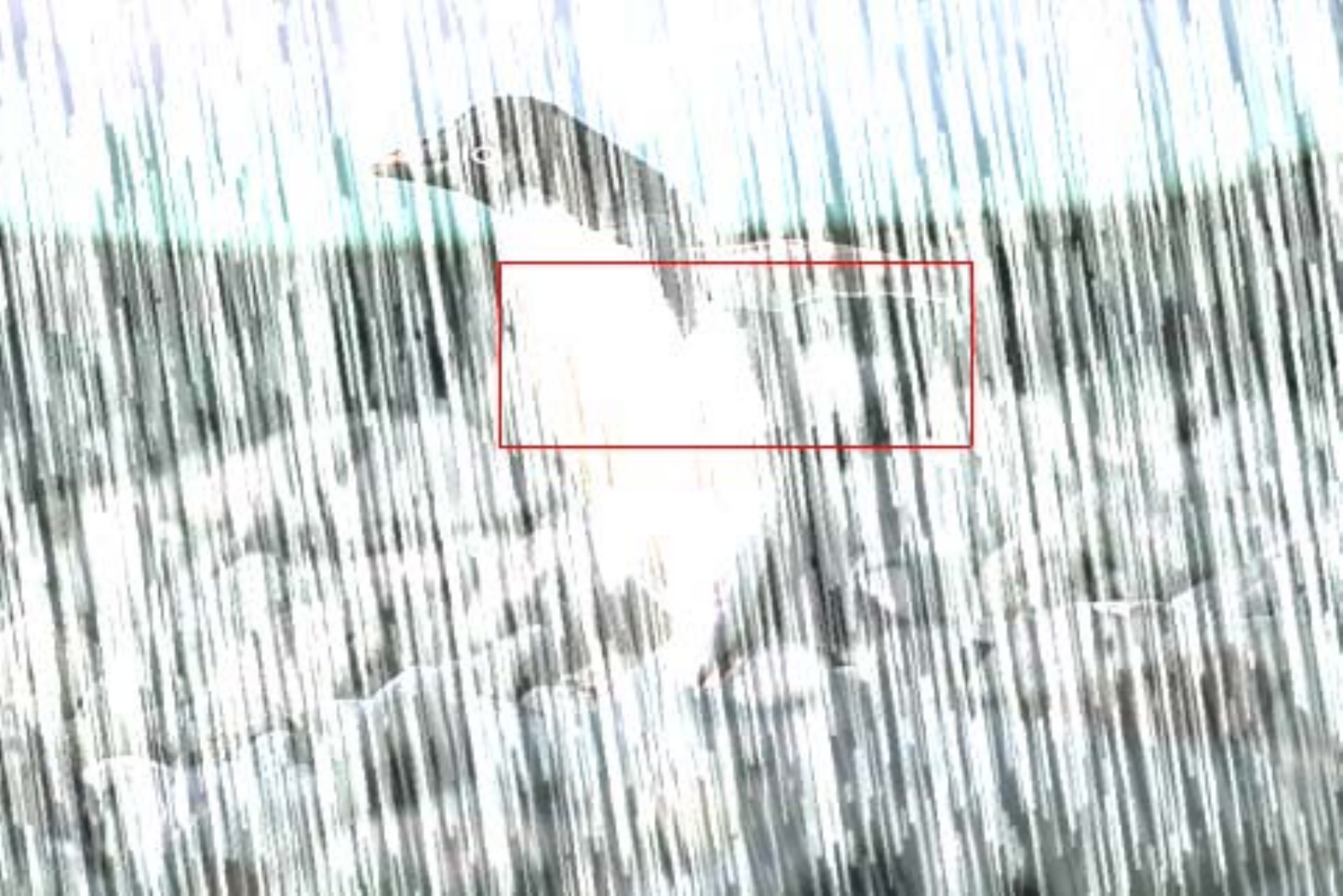} &\hspace{-4mm}
\includegraphics[width = 0.11\linewidth]{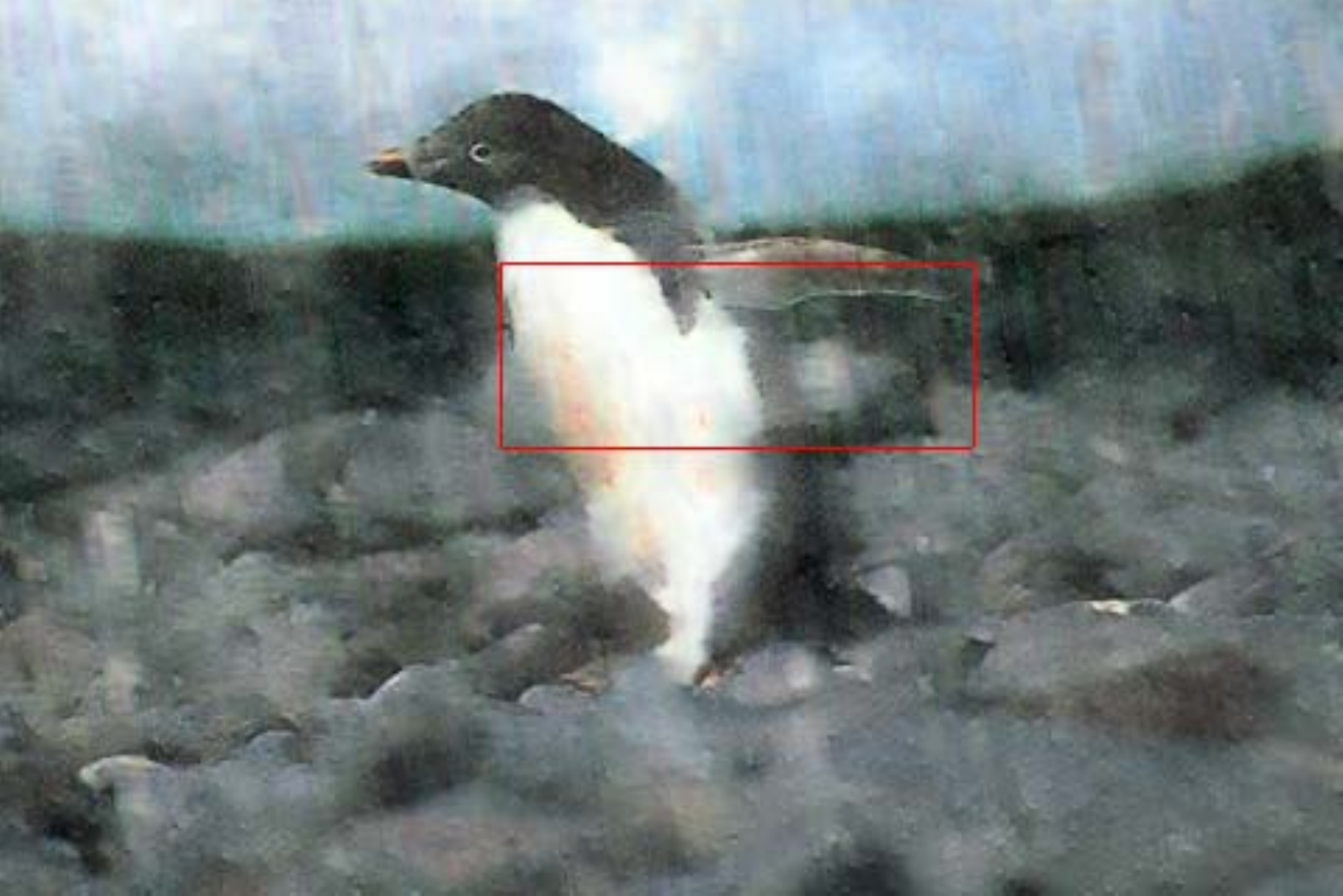} &\hspace{-4mm}
\includegraphics[width = 0.11\linewidth]{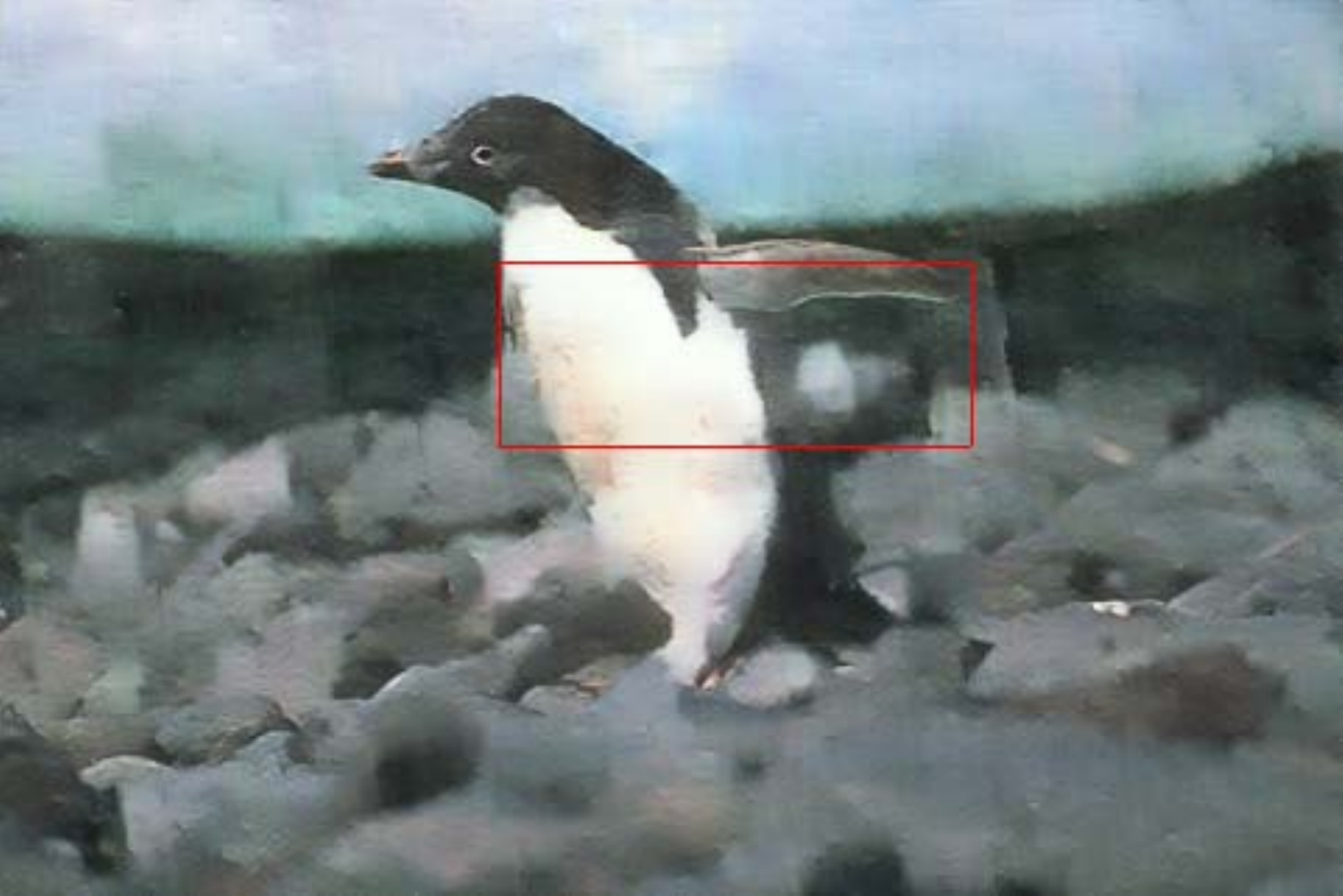} &\hspace{-4mm}
\includegraphics[width = 0.11\linewidth]{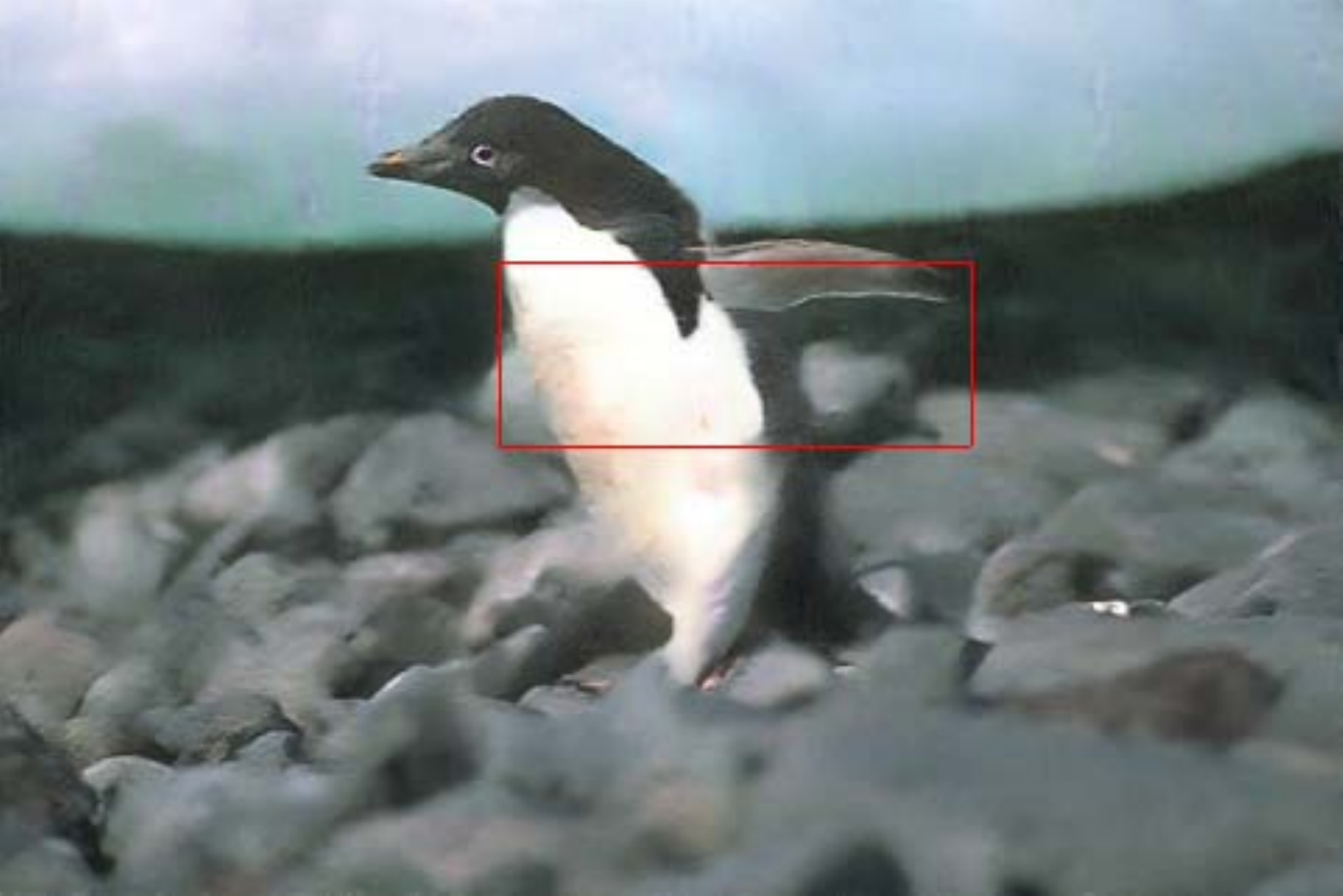} &\hspace{-4mm}
\includegraphics[width = 0.11\linewidth]{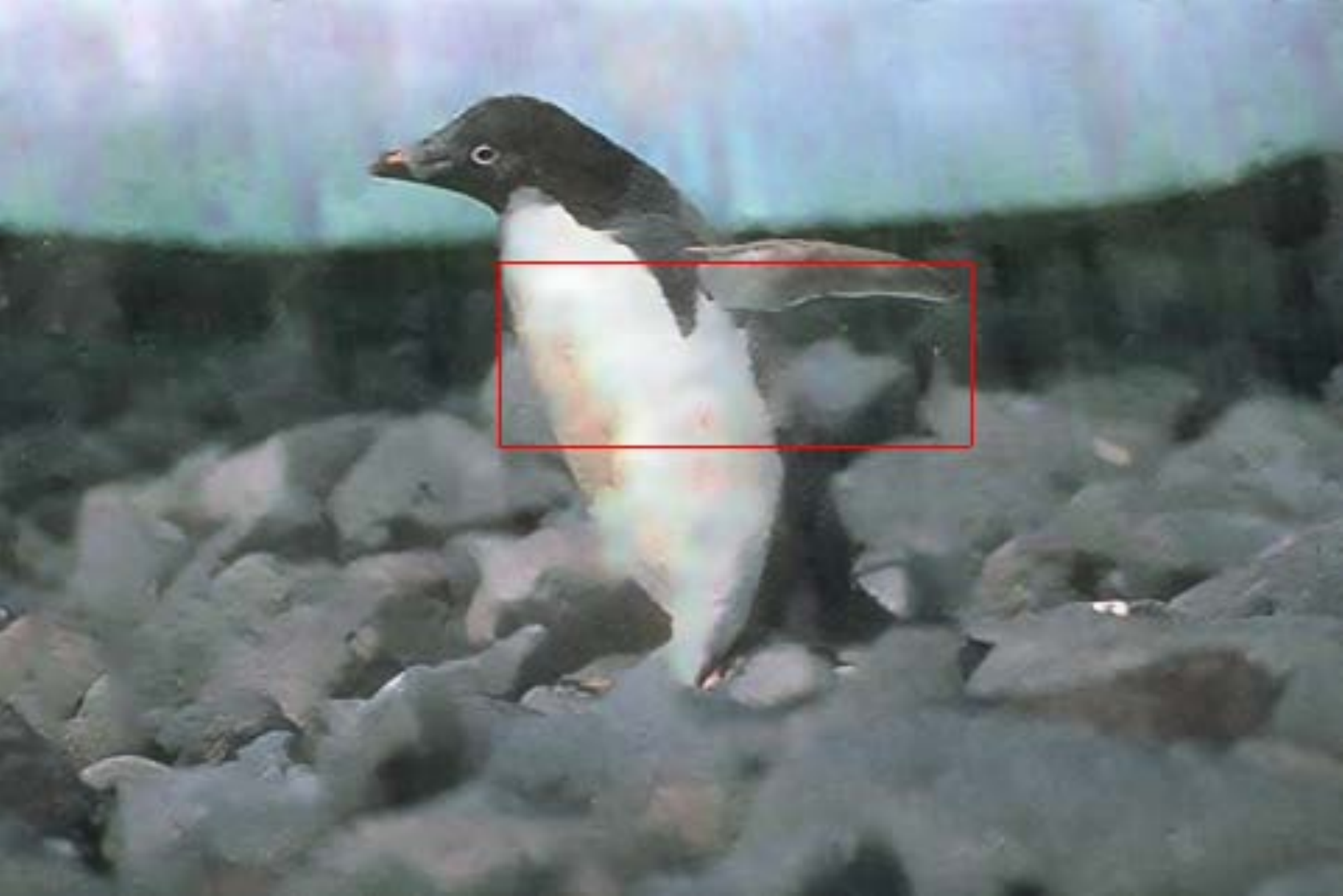} &\hspace{-4mm}
\includegraphics[width = 0.11\linewidth]{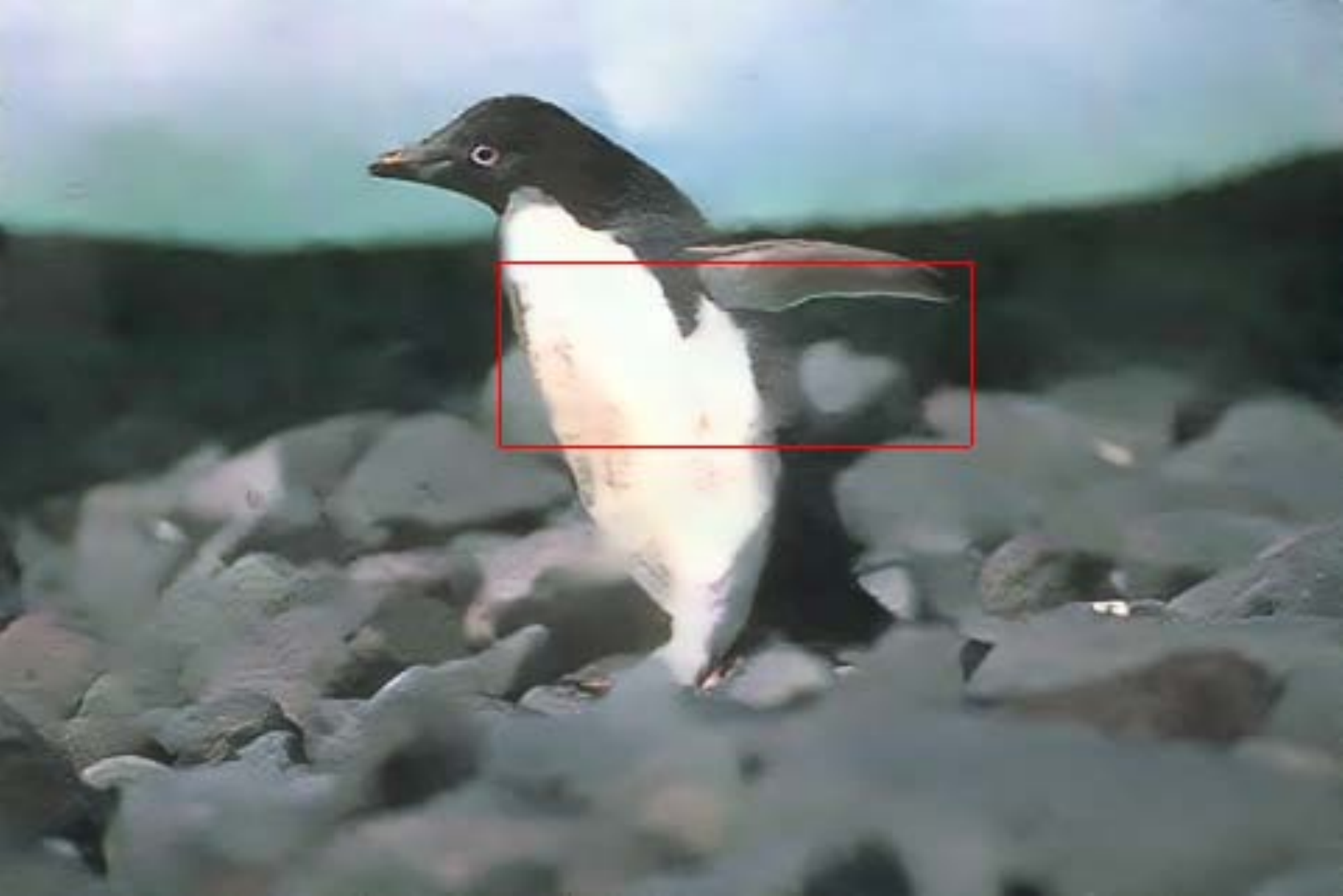} &\hspace{-4mm}
\includegraphics[width = 0.11\linewidth]{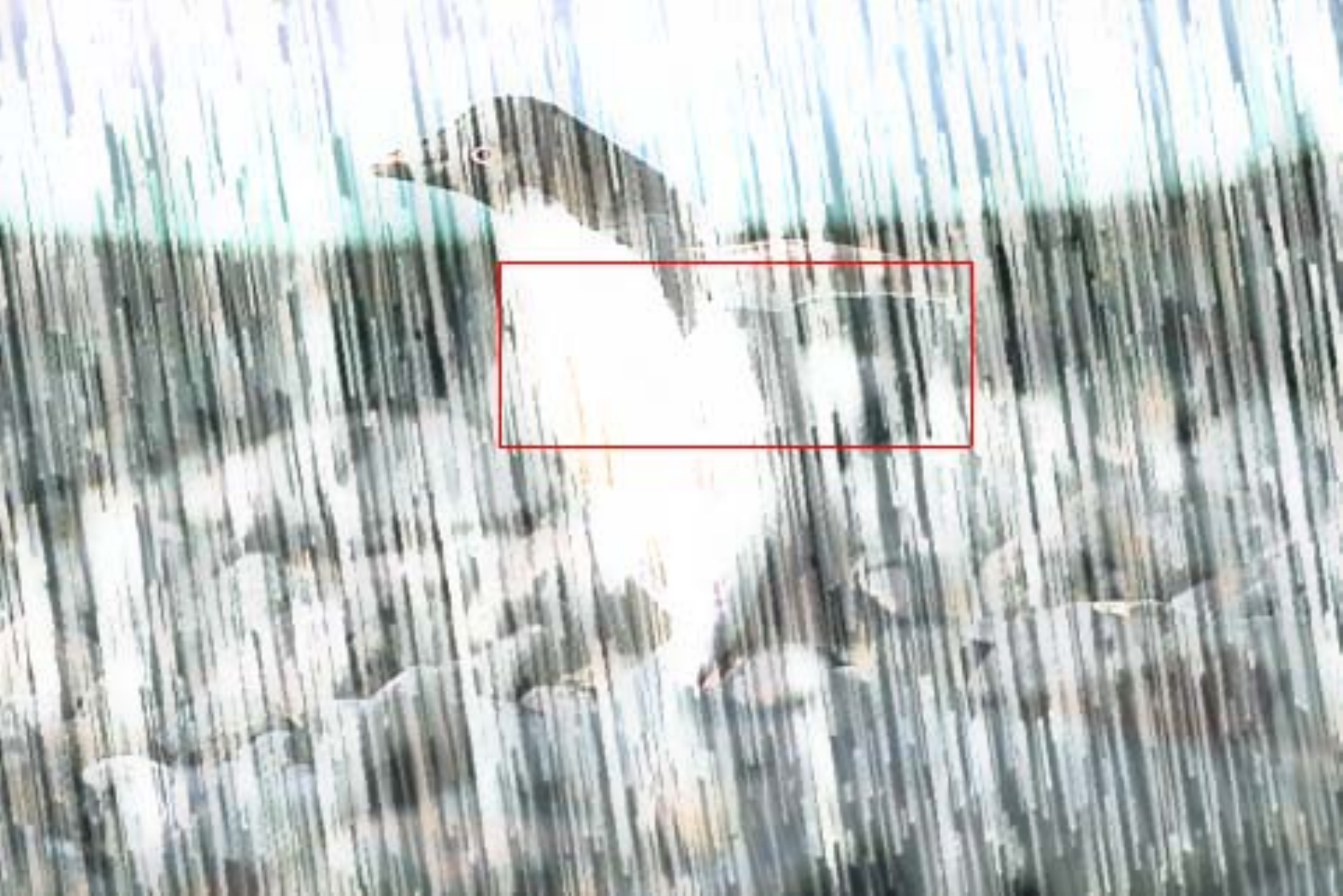} &\hspace{-4mm}
\includegraphics[width = 0.11\linewidth]{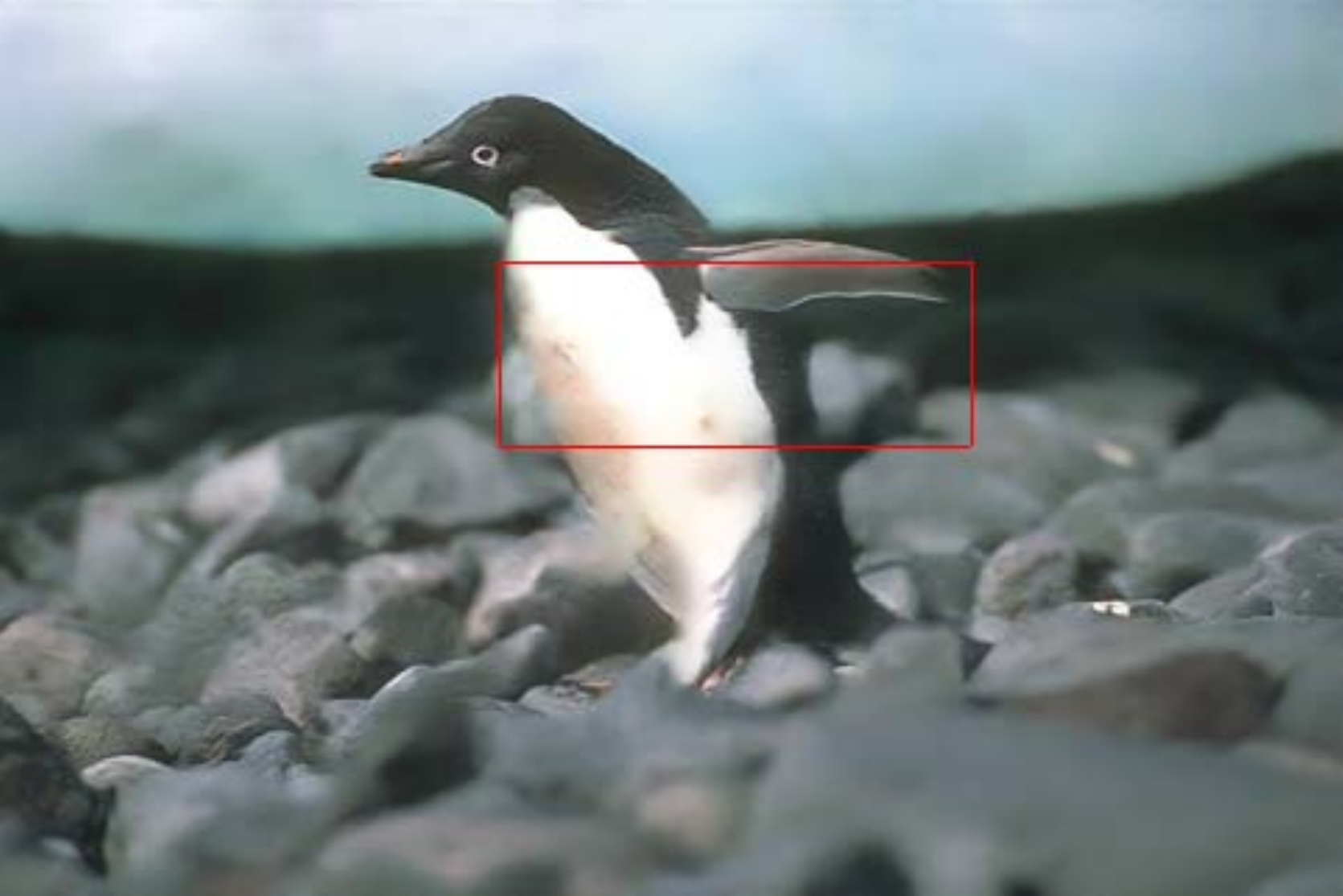} &\hspace{-4mm}
\includegraphics[width = 0.11\linewidth]{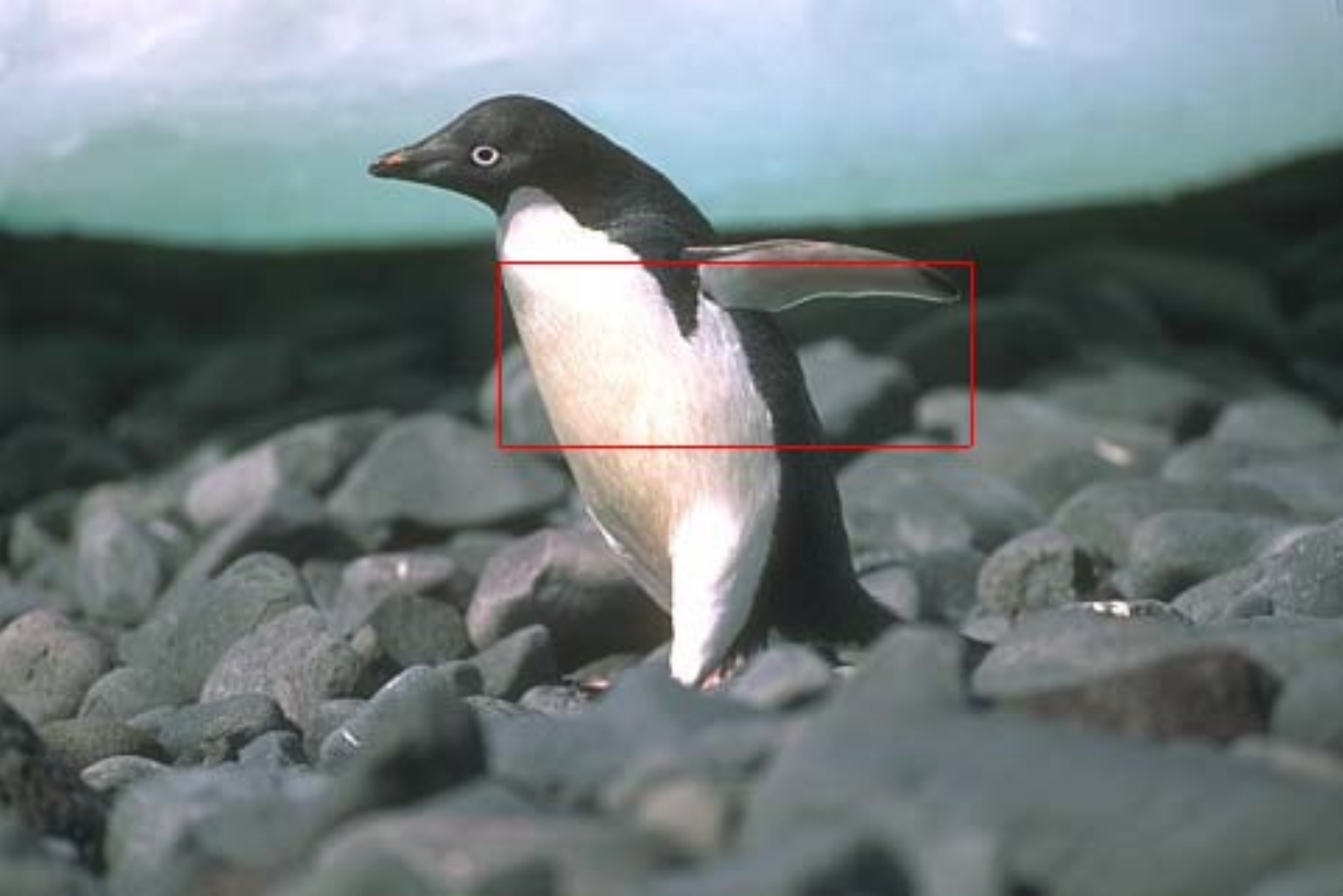}
\\
\includegraphics[width = 0.11\linewidth]{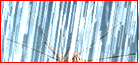} &\hspace{-4mm}
\includegraphics[width = 0.11\linewidth]{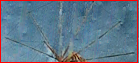} &\hspace{-4mm}
\includegraphics[width = 0.11\linewidth]{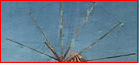} &\hspace{-4mm}
\includegraphics[width = 0.11\linewidth]{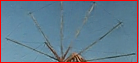} &\hspace{-4mm}
\includegraphics[width = 0.11\linewidth]{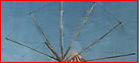} &\hspace{-4mm}
\includegraphics[width = 0.11\linewidth]{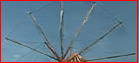} &\hspace{-4mm}
\includegraphics[width = 0.11\linewidth]{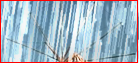} &\hspace{-4mm}
\includegraphics[width = 0.11\linewidth]{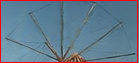} &\hspace{-4mm}
\includegraphics[width = 0.11\linewidth]{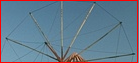}
\\
\includegraphics[width = 0.11\linewidth]{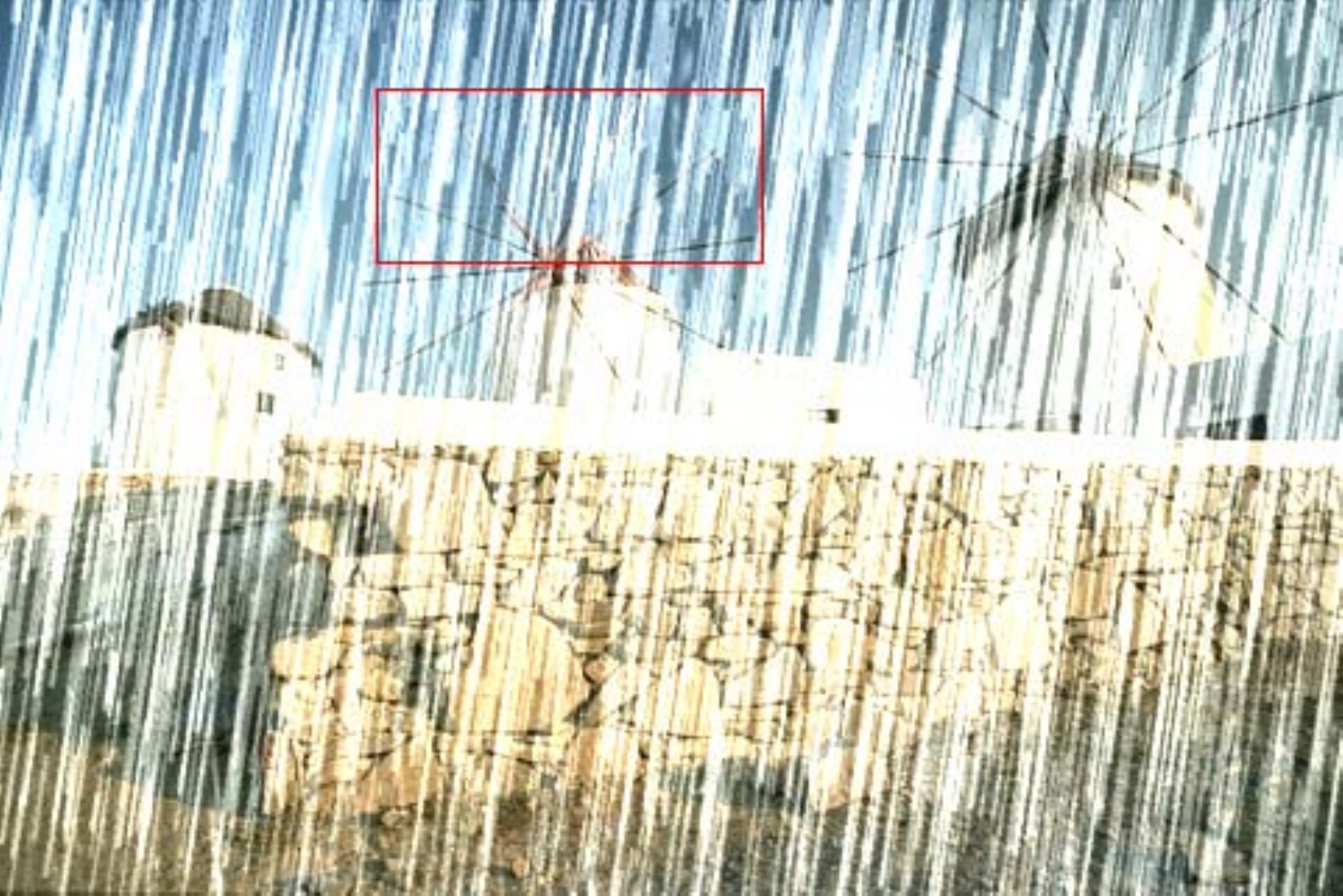} &\hspace{-4mm}
\includegraphics[width = 0.11\linewidth]{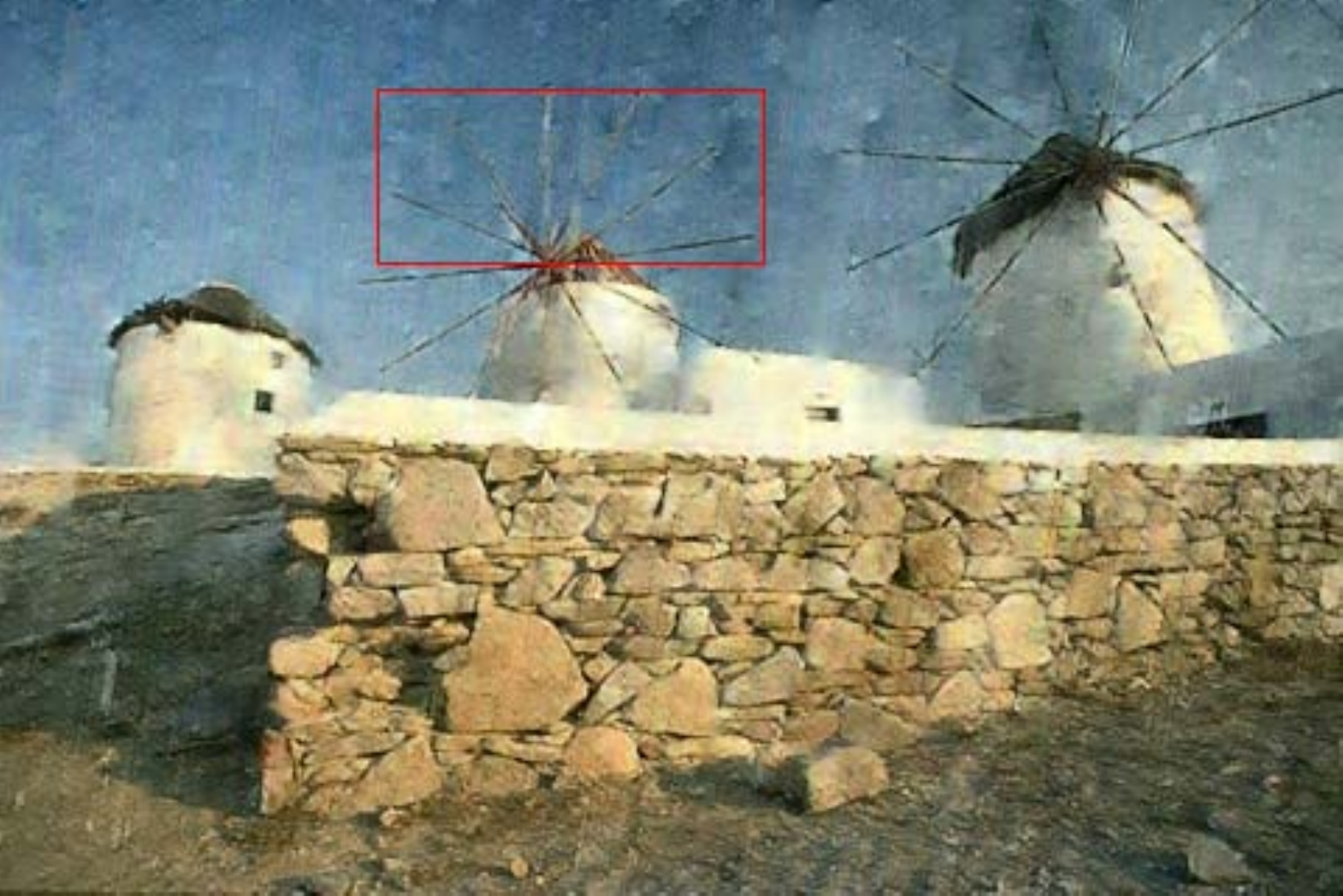} &\hspace{-4mm}
\includegraphics[width = 0.11\linewidth]{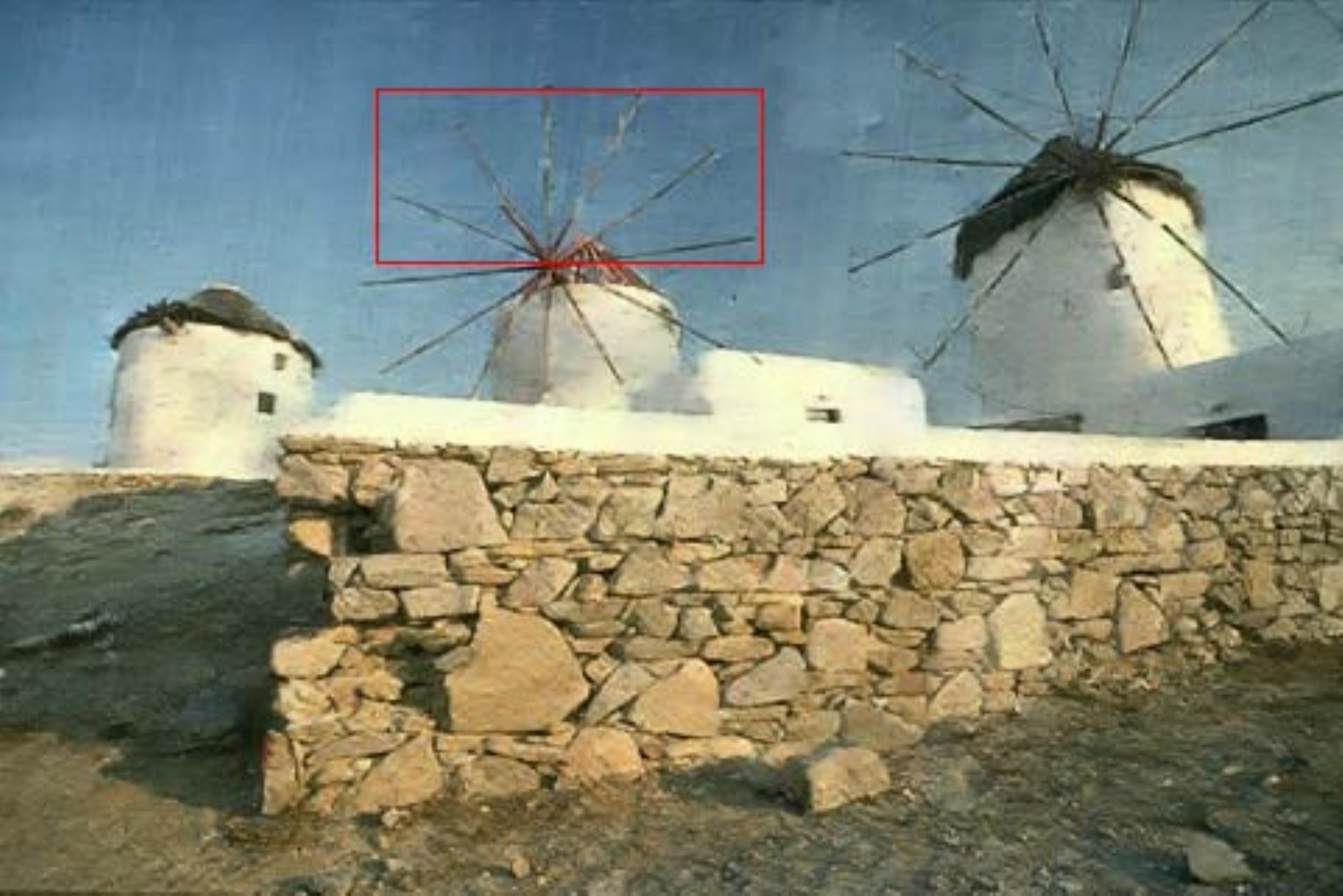} &\hspace{-4mm}
\includegraphics[width = 0.11\linewidth]{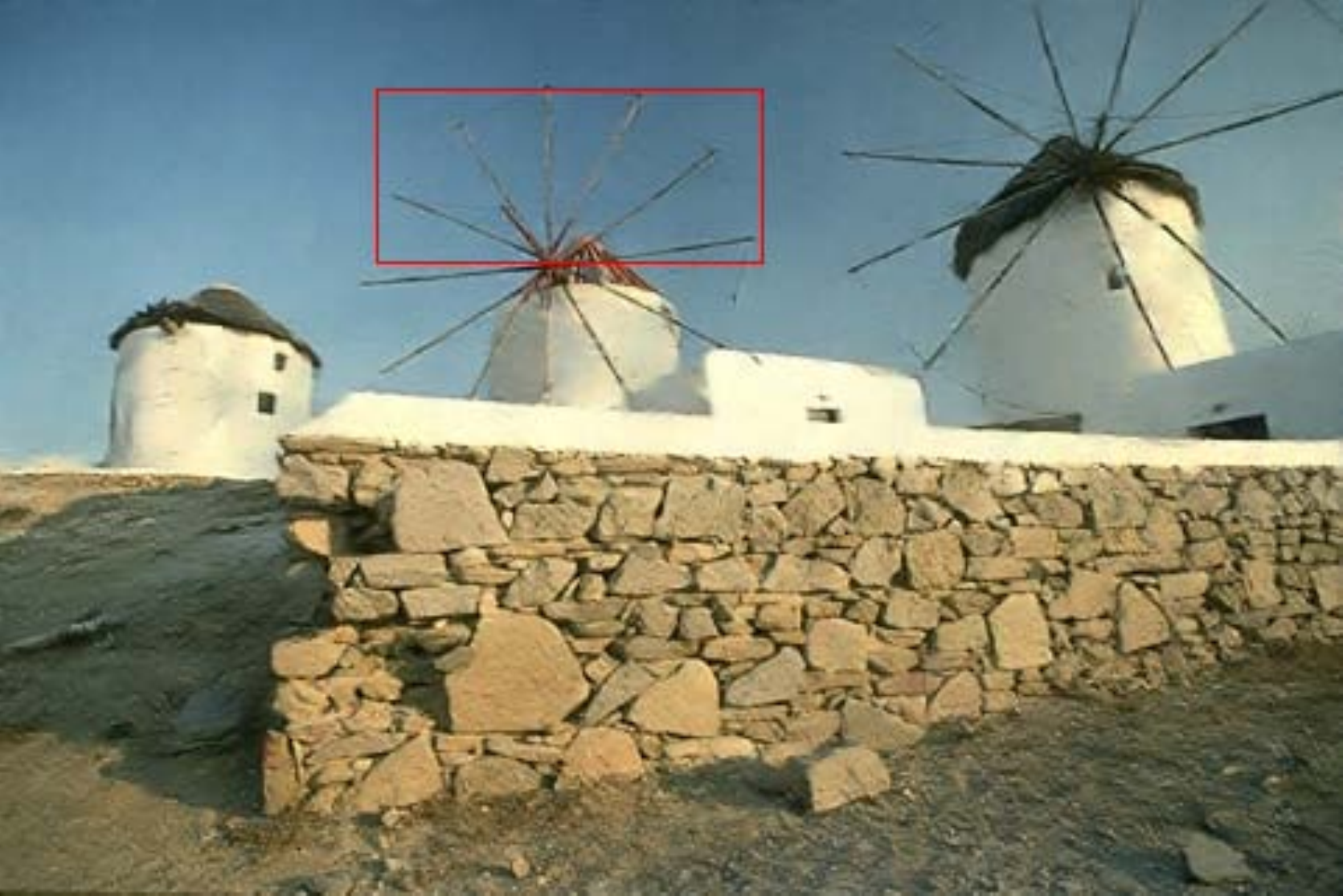} &\hspace{-4mm}
\includegraphics[width = 0.11\linewidth]{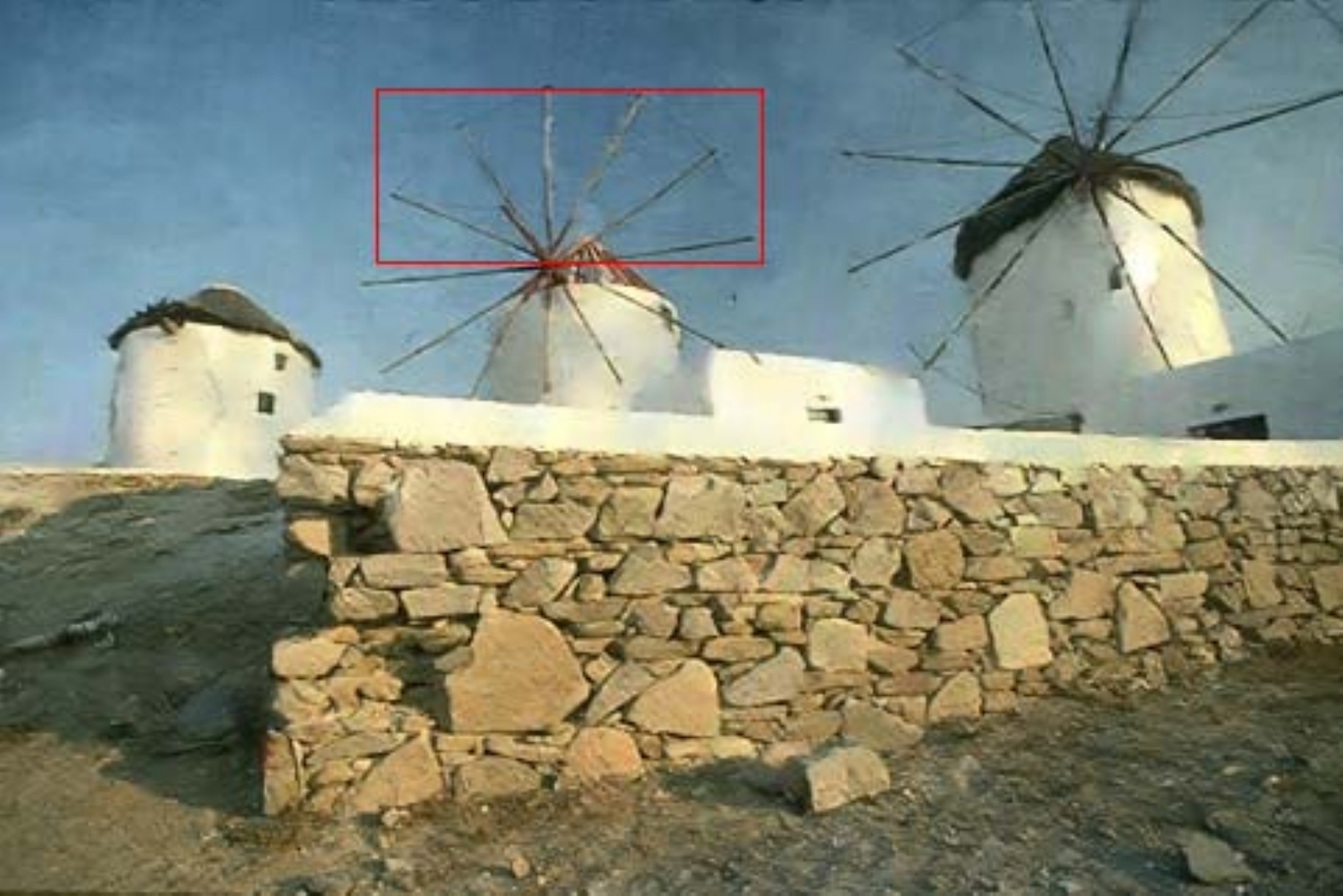} &\hspace{-4mm}
\includegraphics[width = 0.11\linewidth]{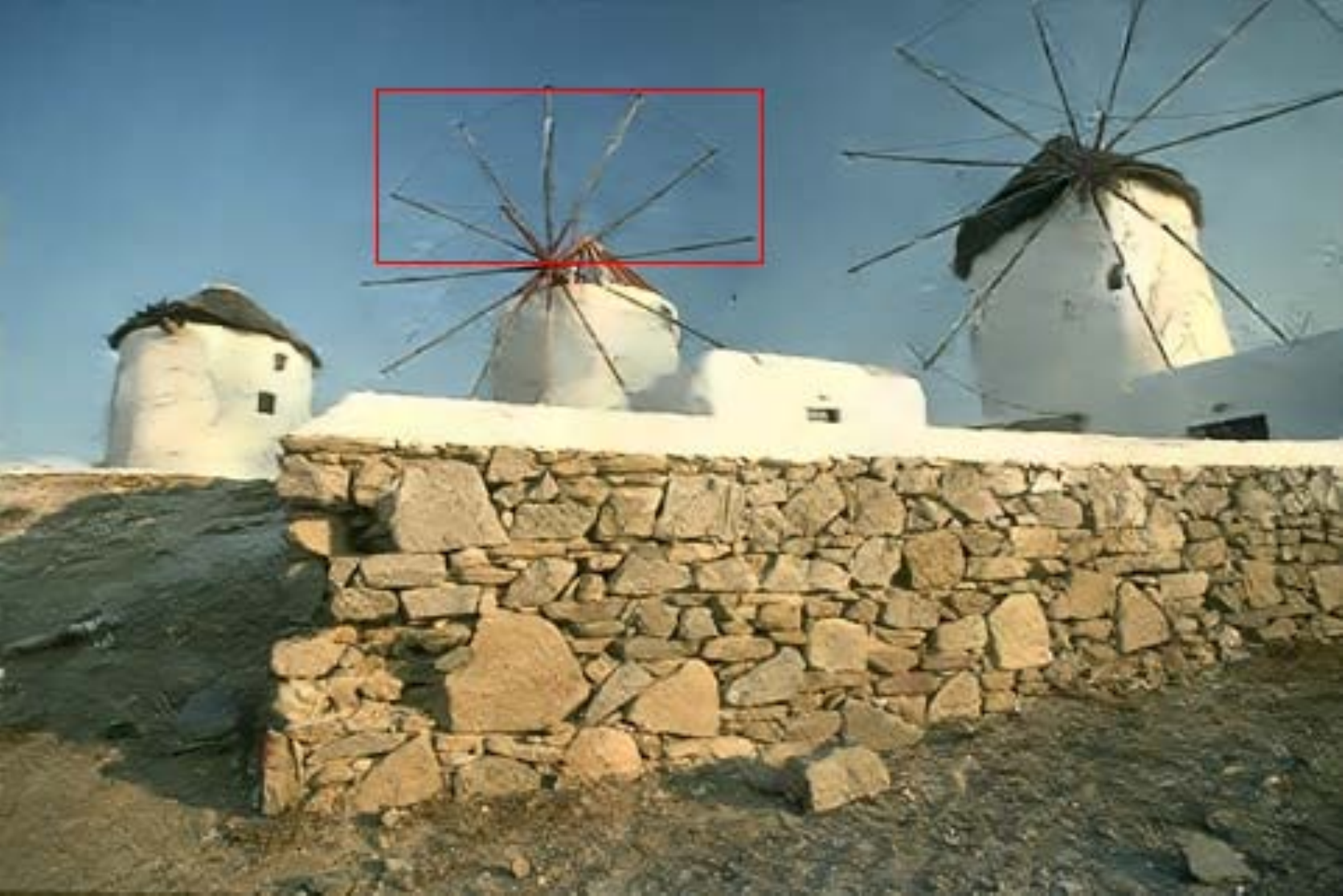} &\hspace{-4mm}
\includegraphics[width = 0.11\linewidth]{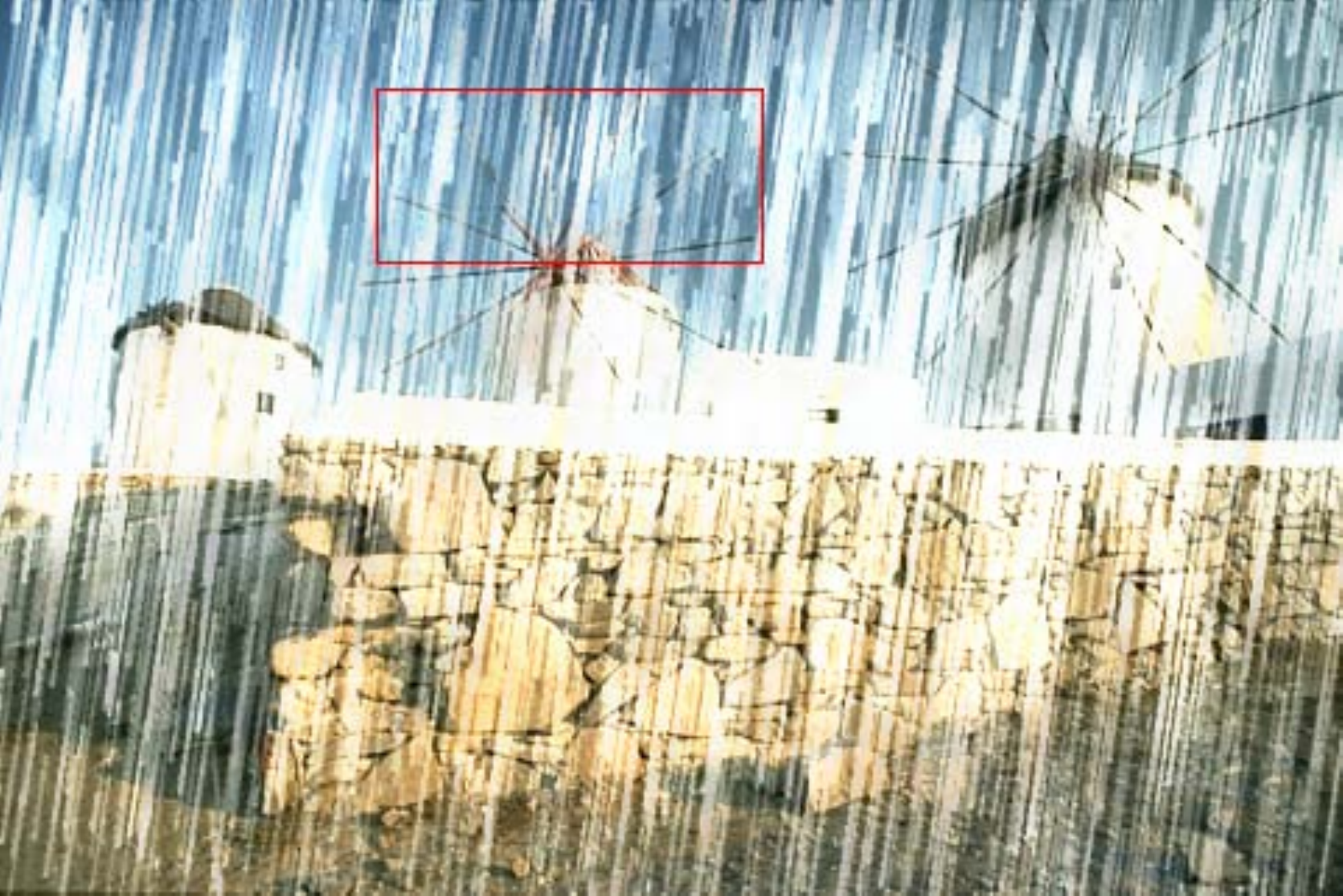} &\hspace{-4mm}
\includegraphics[width = 0.11\linewidth]{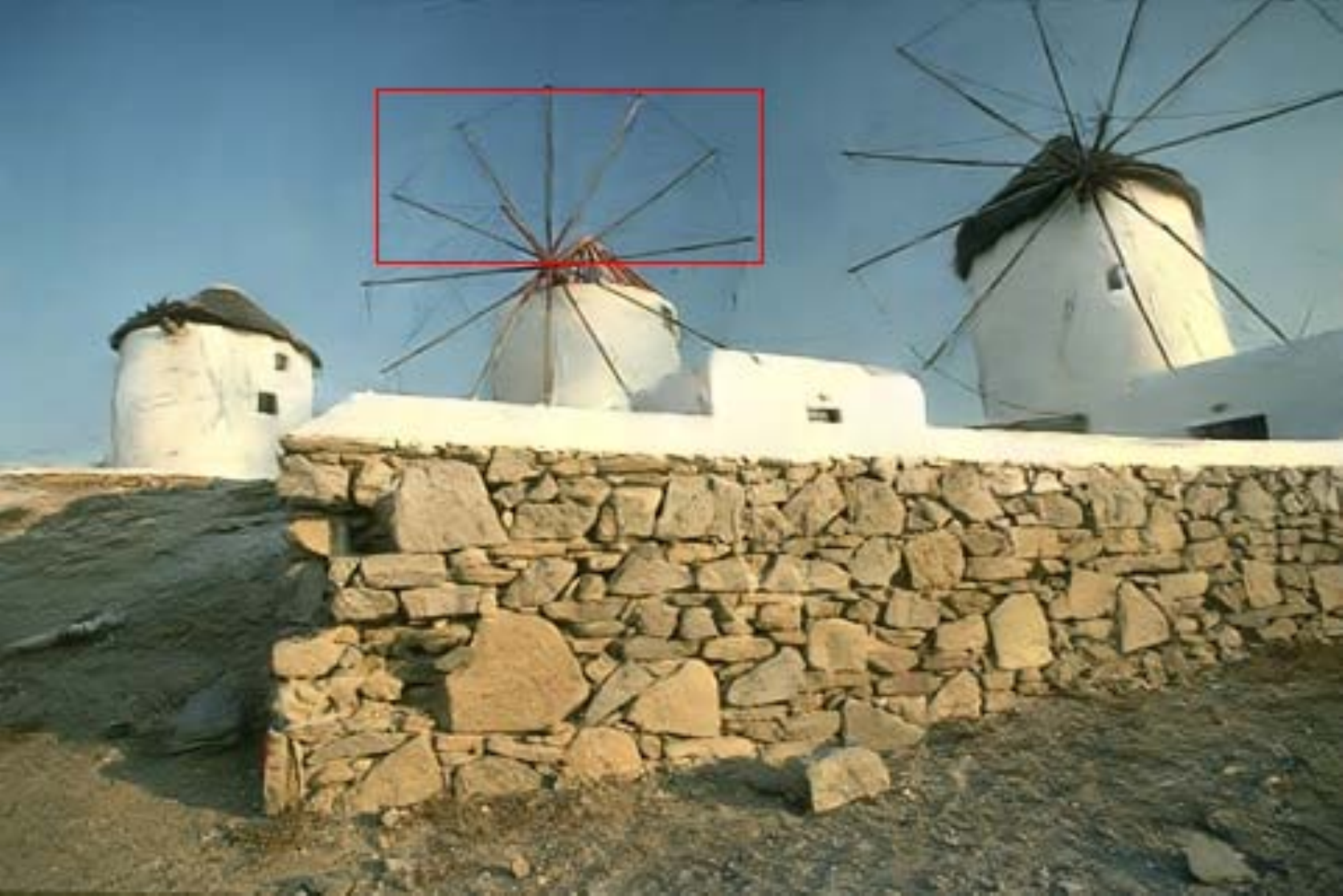} &\hspace{-4mm}
\includegraphics[width = 0.11\linewidth]{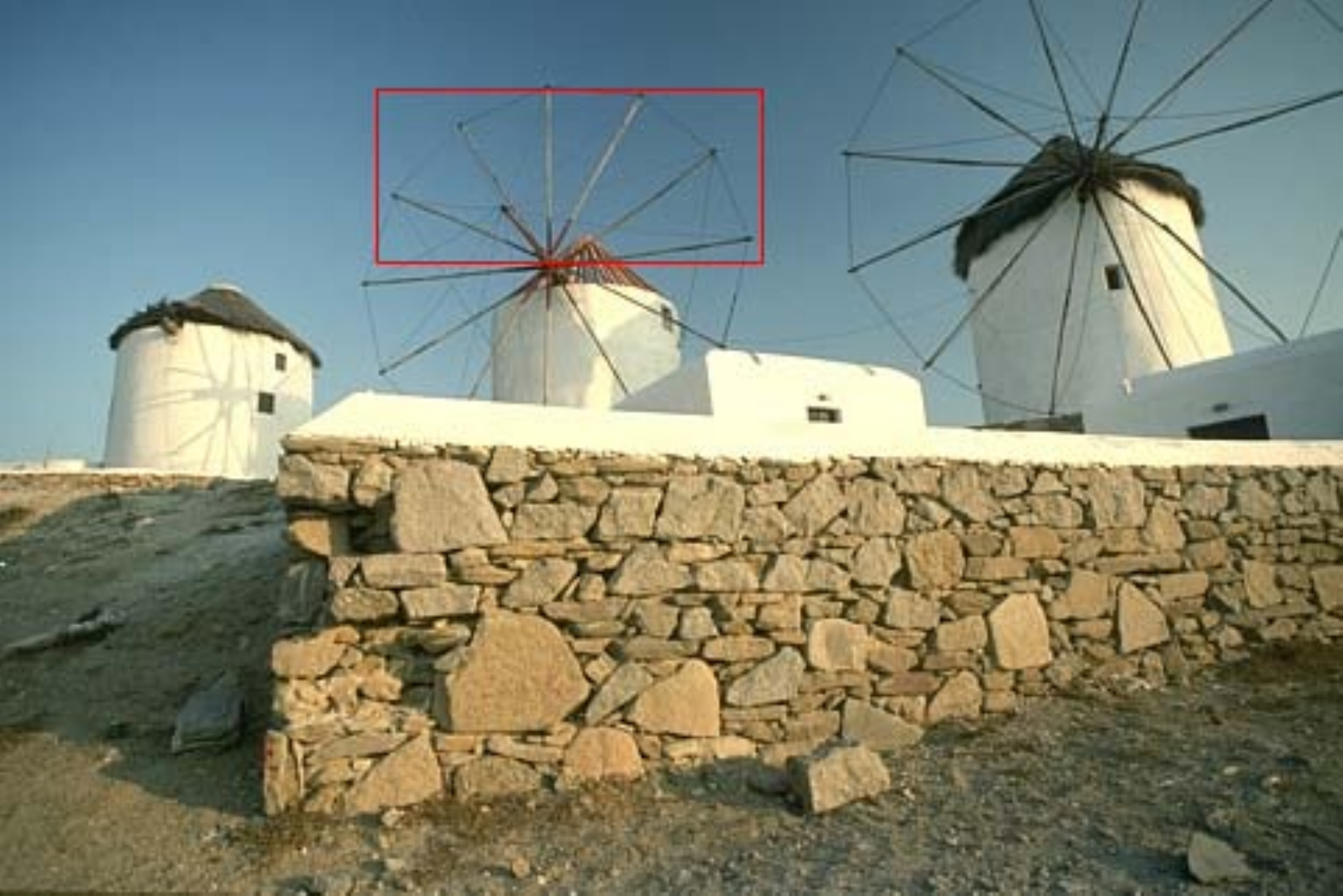}
\\
\includegraphics[width = 0.11\linewidth]{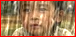} &\hspace{-4mm}
\includegraphics[width = 0.11\linewidth]{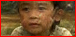} &\hspace{-4mm}
\includegraphics[width = 0.11\linewidth]{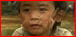} &\hspace{-4mm}
\includegraphics[width = 0.11\linewidth]{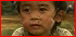} &\hspace{-4mm}
\includegraphics[width = 0.11\linewidth]{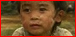} &\hspace{-4mm}
\includegraphics[width = 0.11\linewidth]{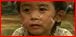} &\hspace{-4mm}
\includegraphics[width = 0.11\linewidth]{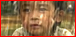} &\hspace{-4mm}
\includegraphics[width = 0.11\linewidth]{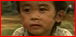} &\hspace{-4mm}
\includegraphics[width = 0.11\linewidth]{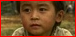}
\\
\includegraphics[width = 0.11\linewidth]{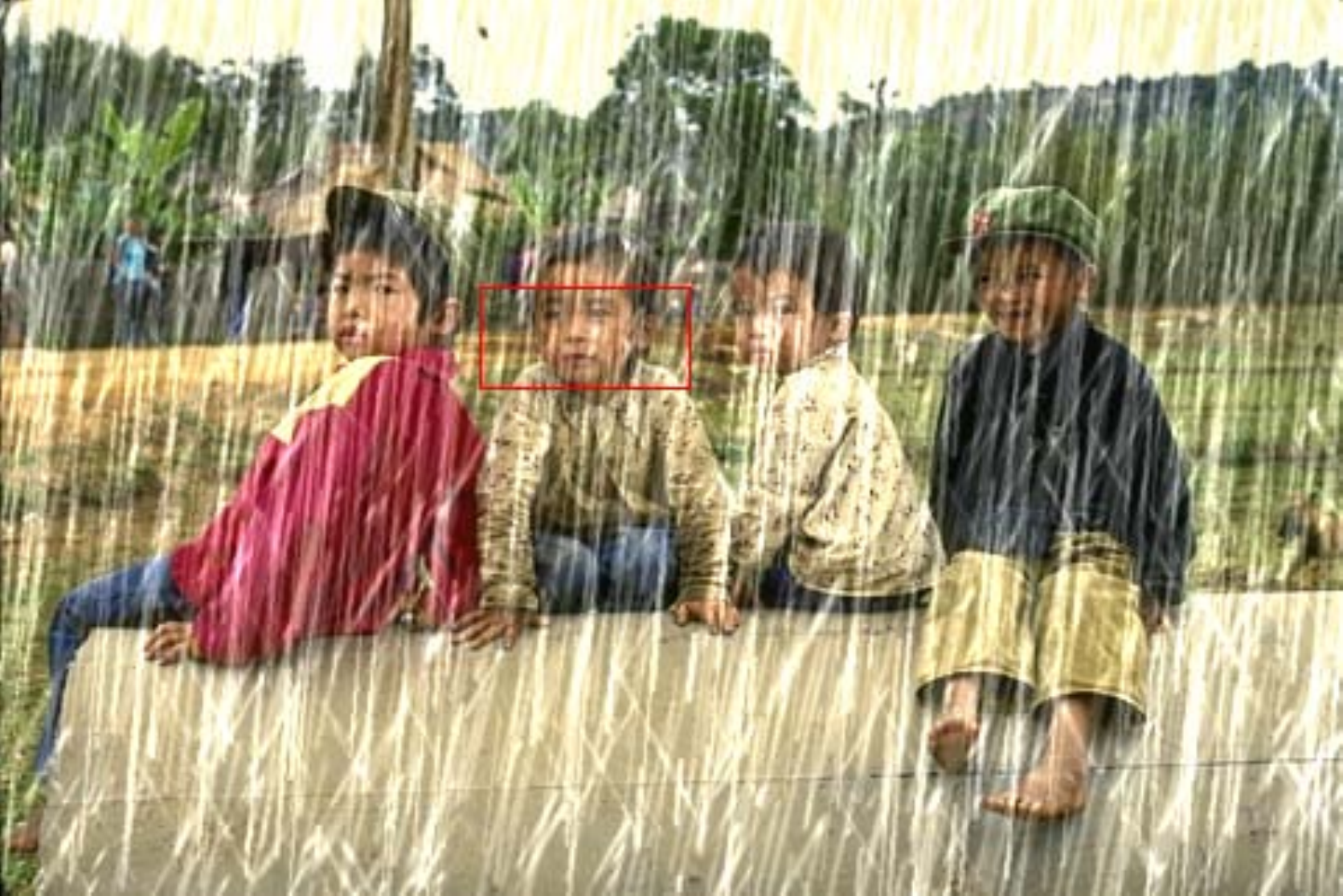} &\hspace{-4mm}
\includegraphics[width = 0.11\linewidth]{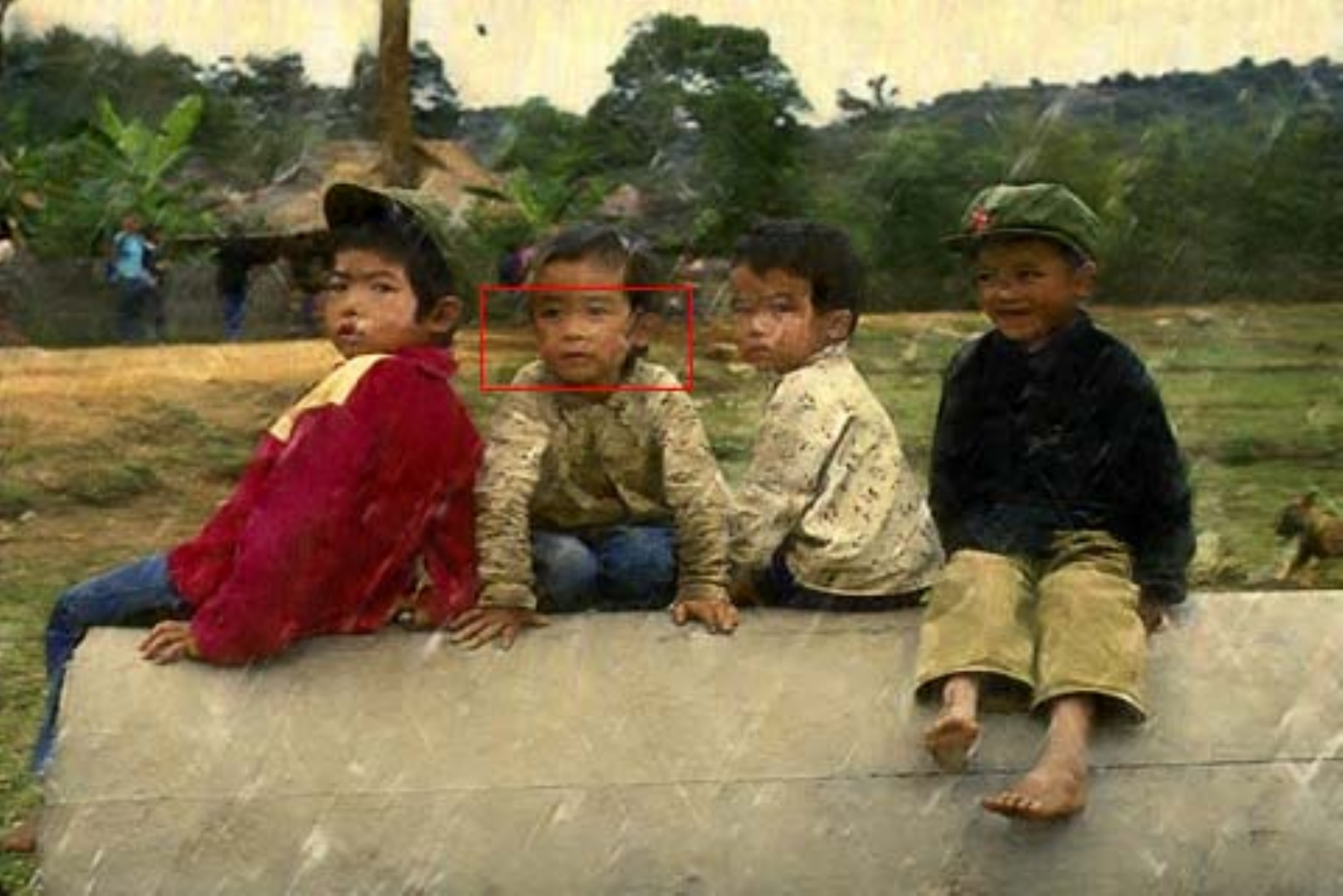} &\hspace{-4mm}
\includegraphics[width = 0.11\linewidth]{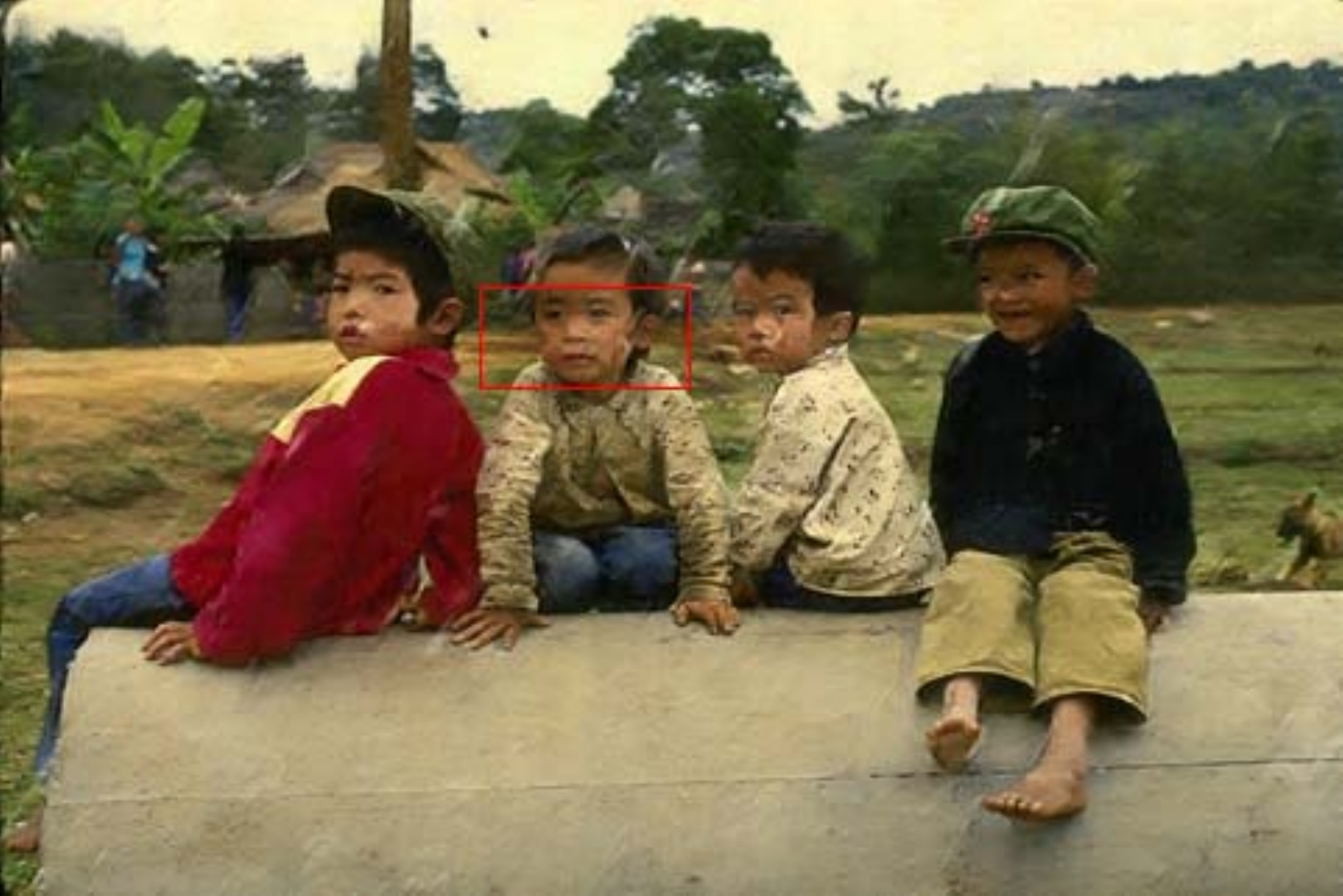} &\hspace{-4mm}
\includegraphics[width = 0.11\linewidth]{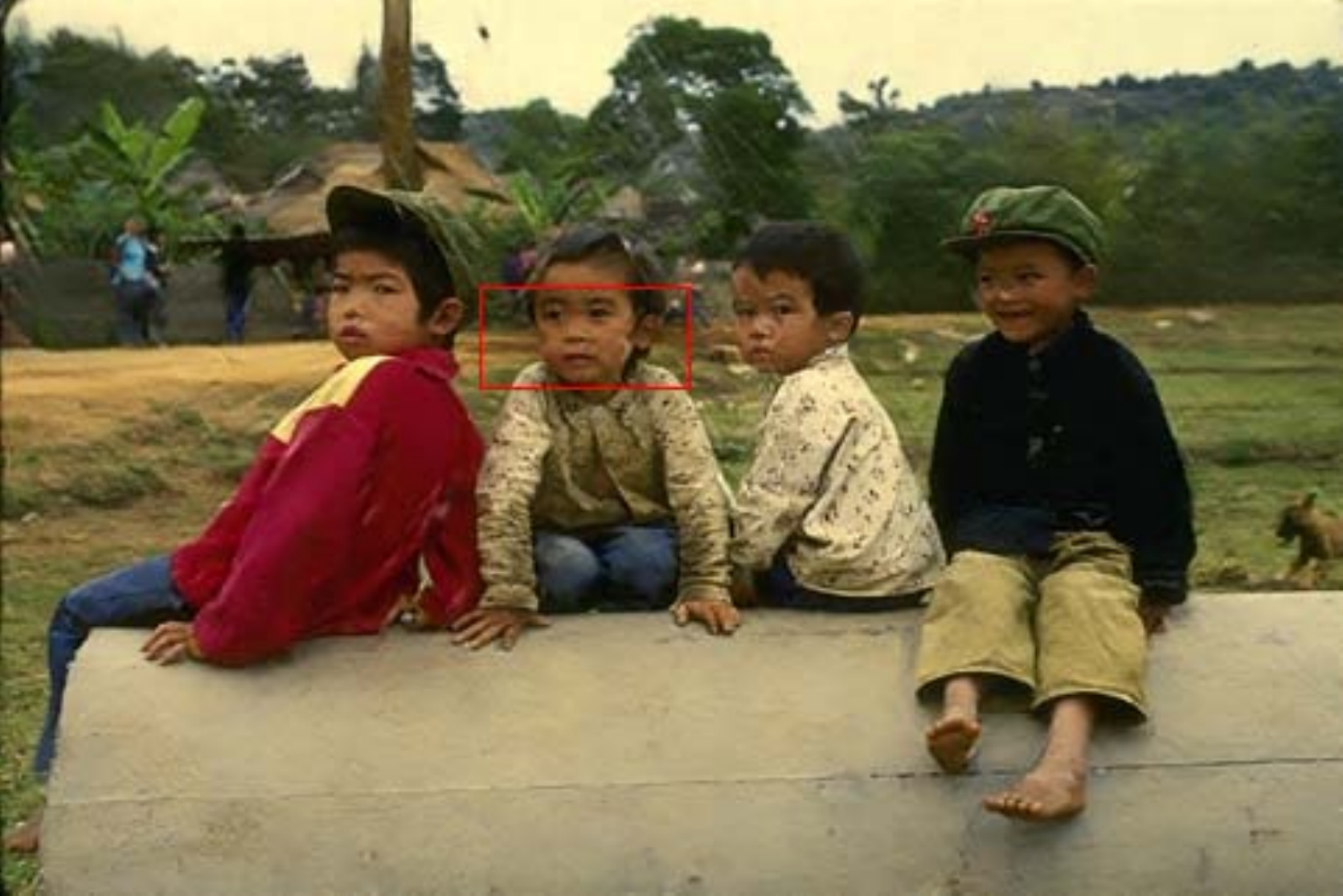} &\hspace{-4mm}
\includegraphics[width = 0.11\linewidth]{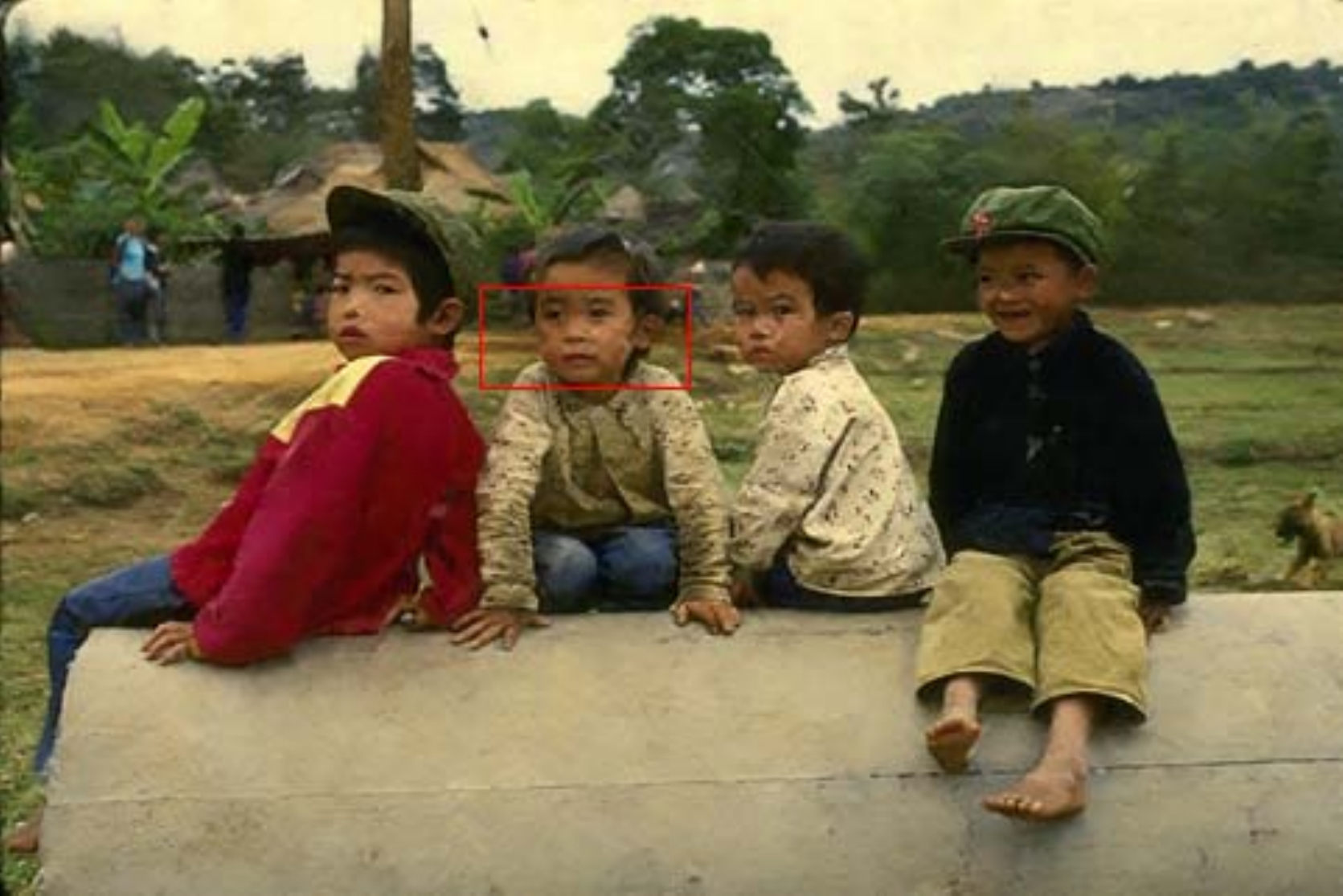} &\hspace{-4mm}
\includegraphics[width = 0.11\linewidth]{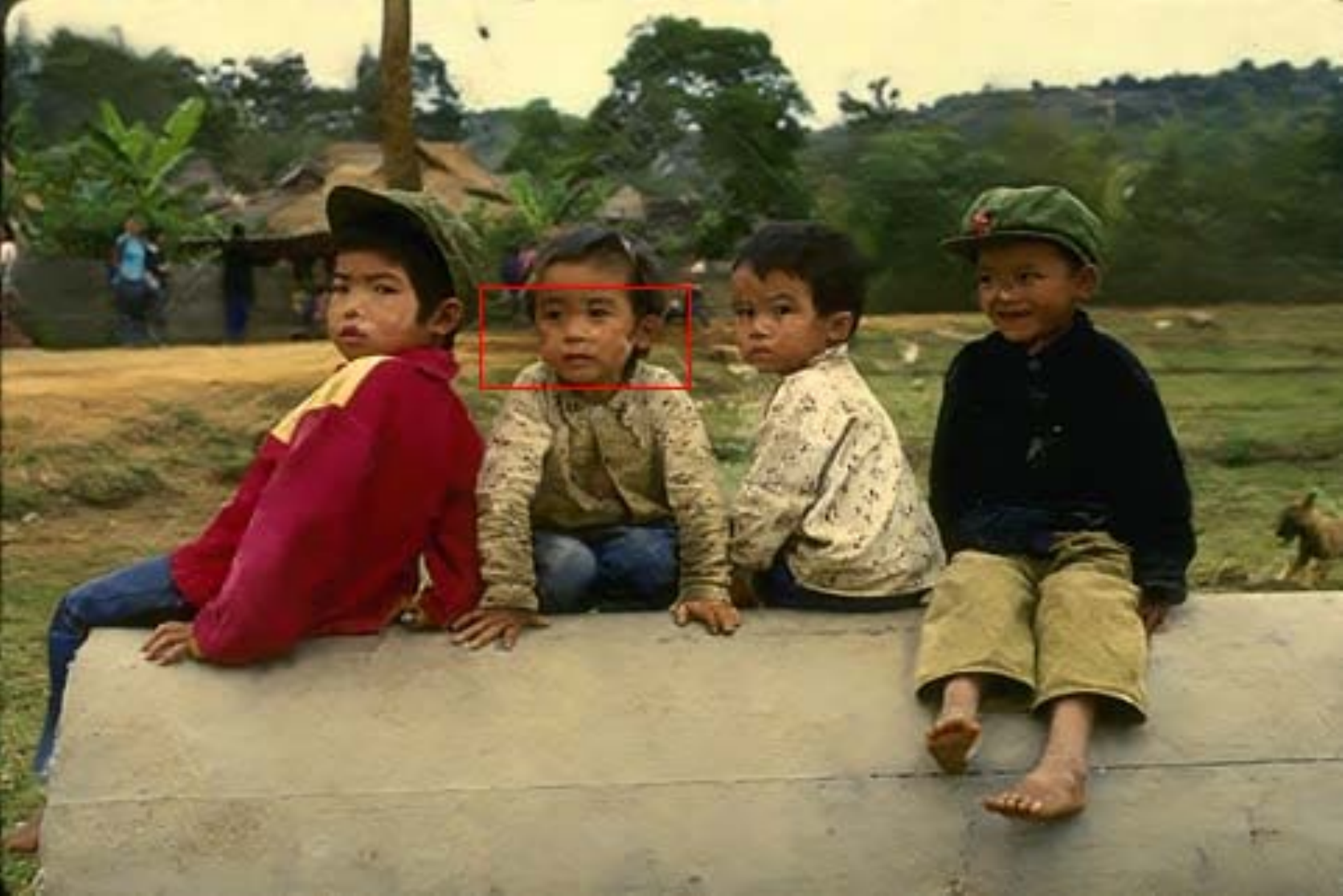} &\hspace{-4mm}
\includegraphics[width = 0.11\linewidth]{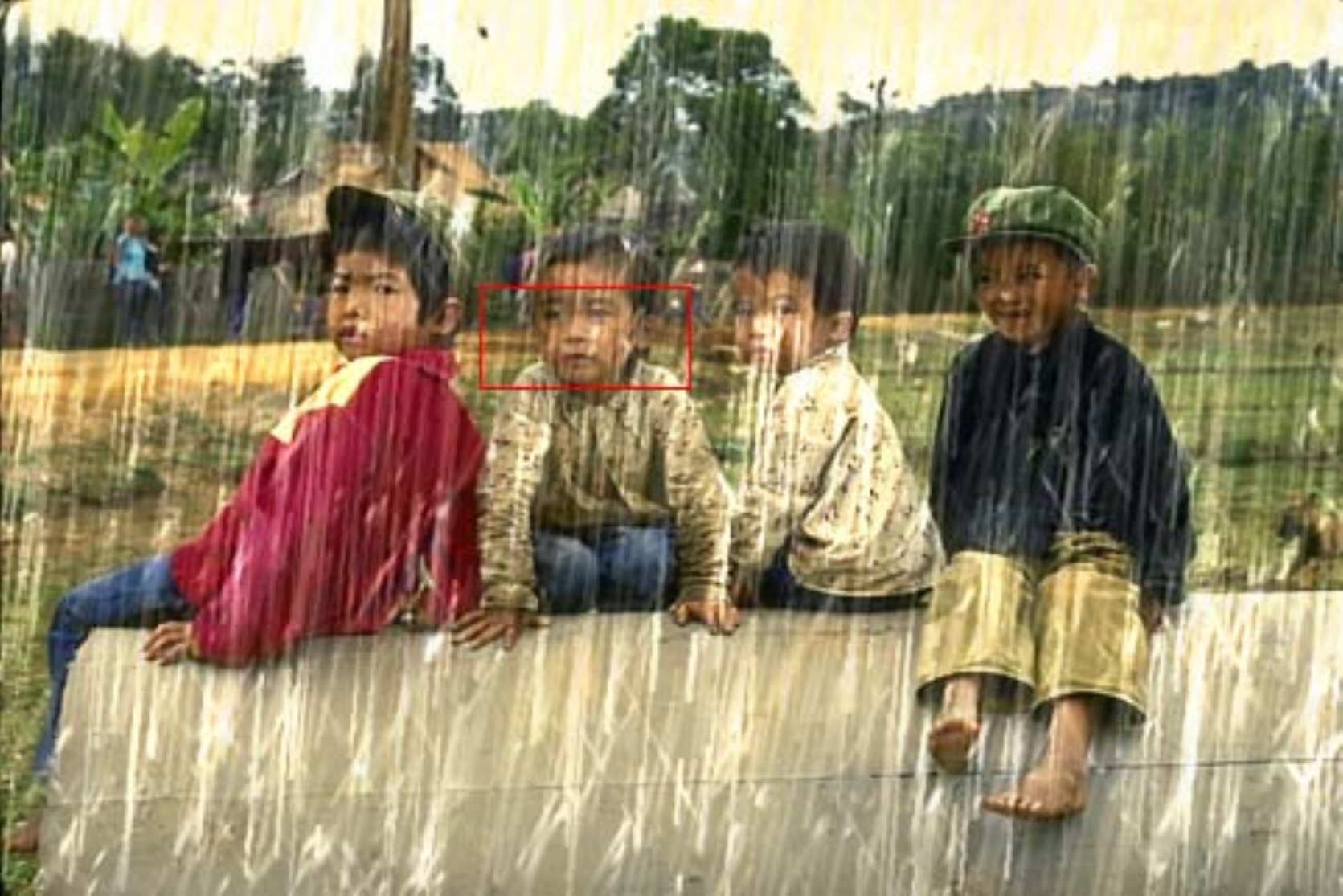} &\hspace{-4mm}
\includegraphics[width = 0.11\linewidth]{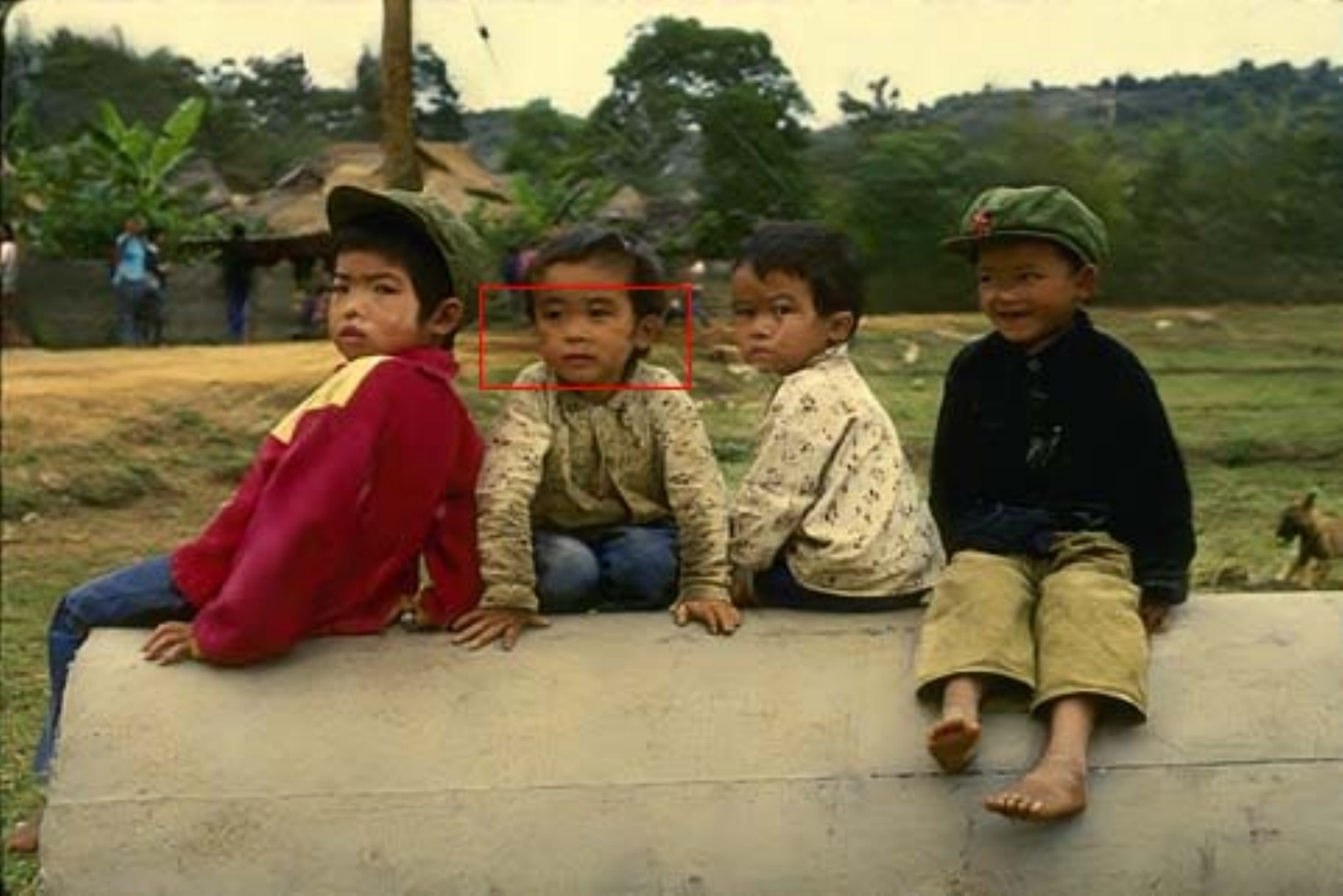} &\hspace{-4mm}
\includegraphics[width = 0.11\linewidth]{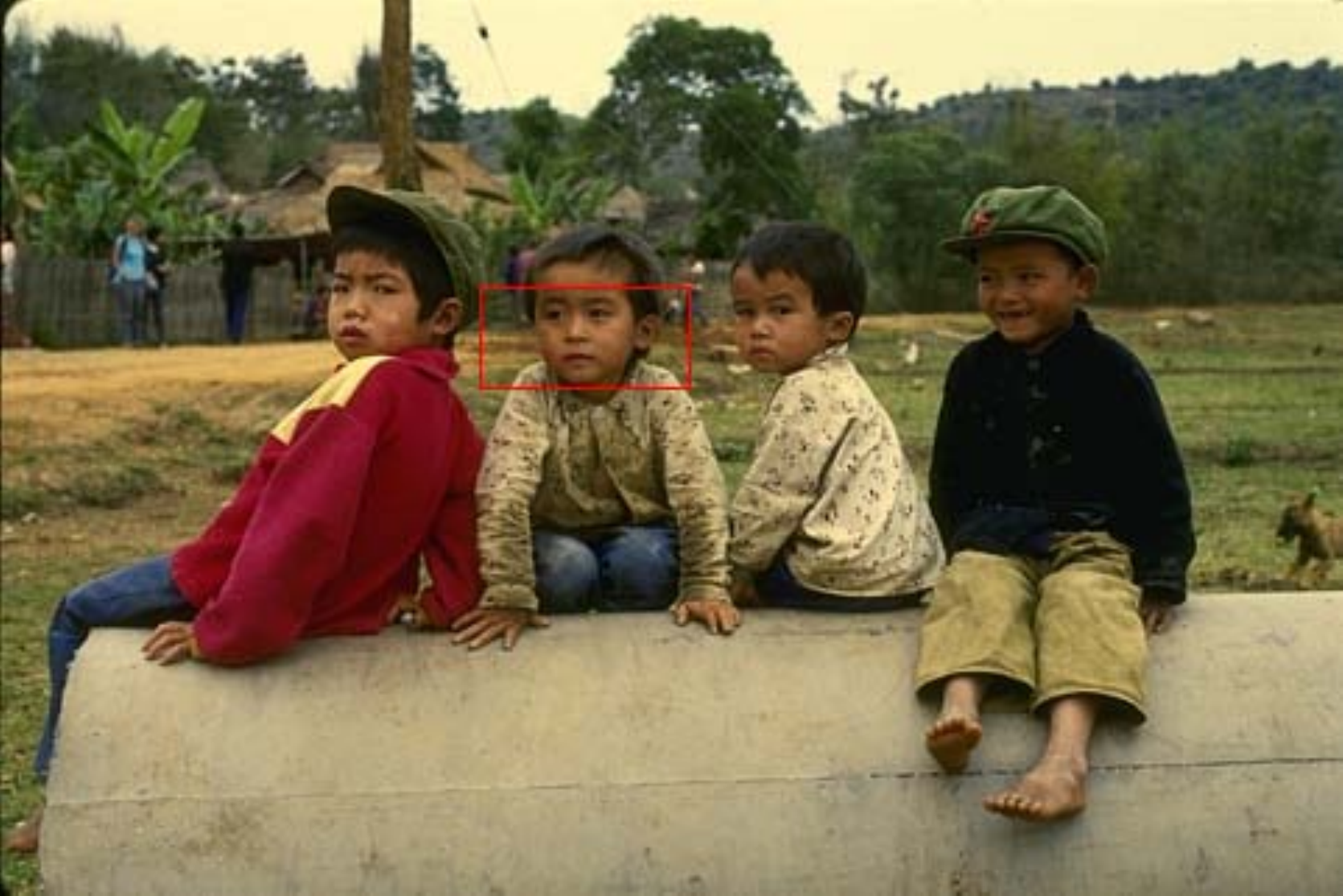}
\\
\includegraphics[width = 0.11\linewidth]{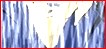} &\hspace{-4mm}
\includegraphics[width = 0.11\linewidth]{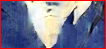} &\hspace{-4mm}
\includegraphics[width = 0.11\linewidth]{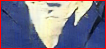} &\hspace{-4mm}
\includegraphics[width = 0.11\linewidth]{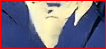} &\hspace{-4mm}
\includegraphics[width = 0.11\linewidth]{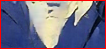} &\hspace{-4mm}
\includegraphics[width = 0.11\linewidth]{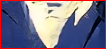}&\hspace{-4mm}
\includegraphics[width = 0.11\linewidth]{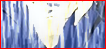} &\hspace{-4mm}
\includegraphics[width = 0.11\linewidth]{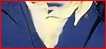} &\hspace{-4mm}
\includegraphics[width = 0.11\linewidth]{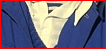}
\\
\includegraphics[width = 0.11\linewidth]{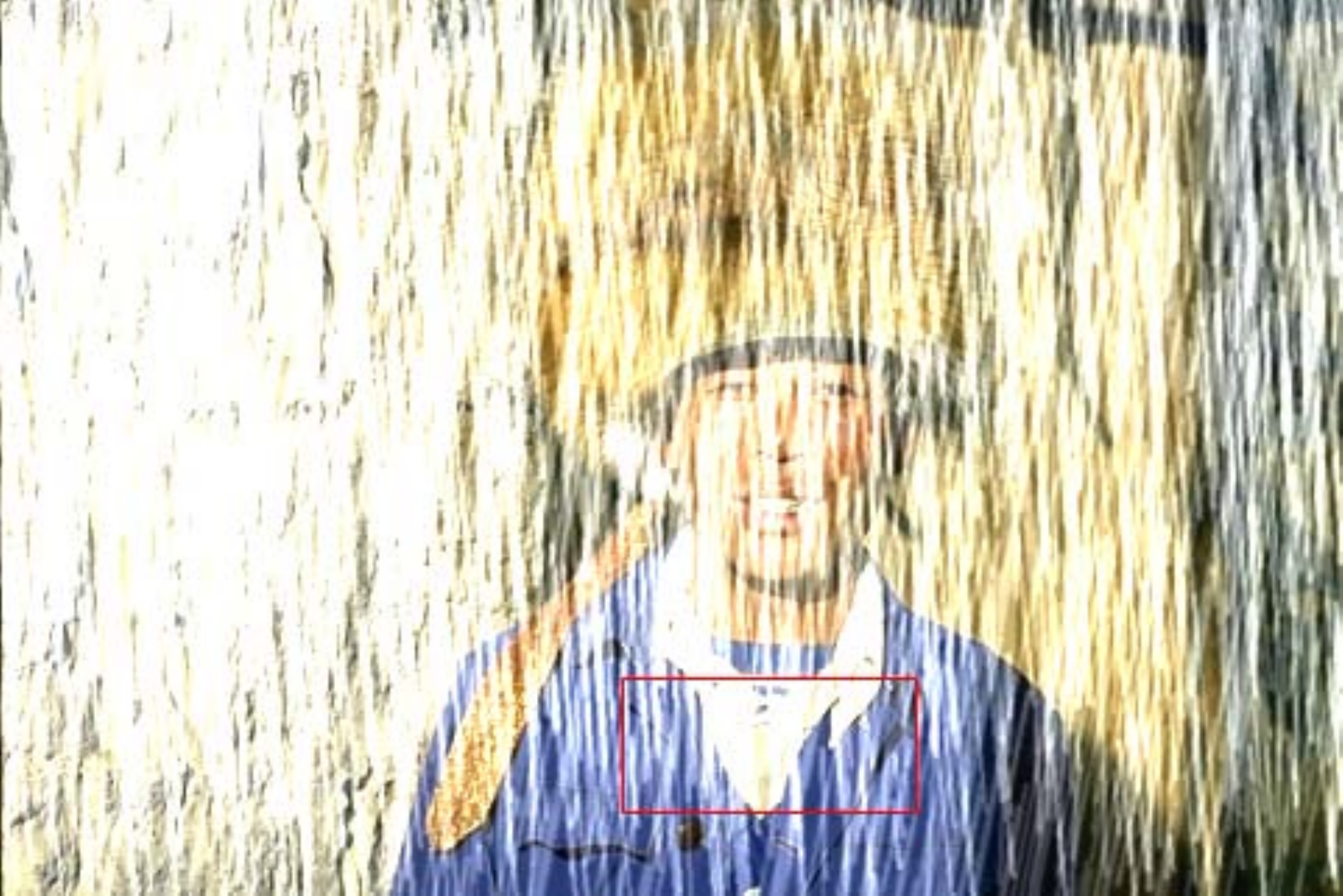} &\hspace{-4mm}
\includegraphics[width = 0.11\linewidth]{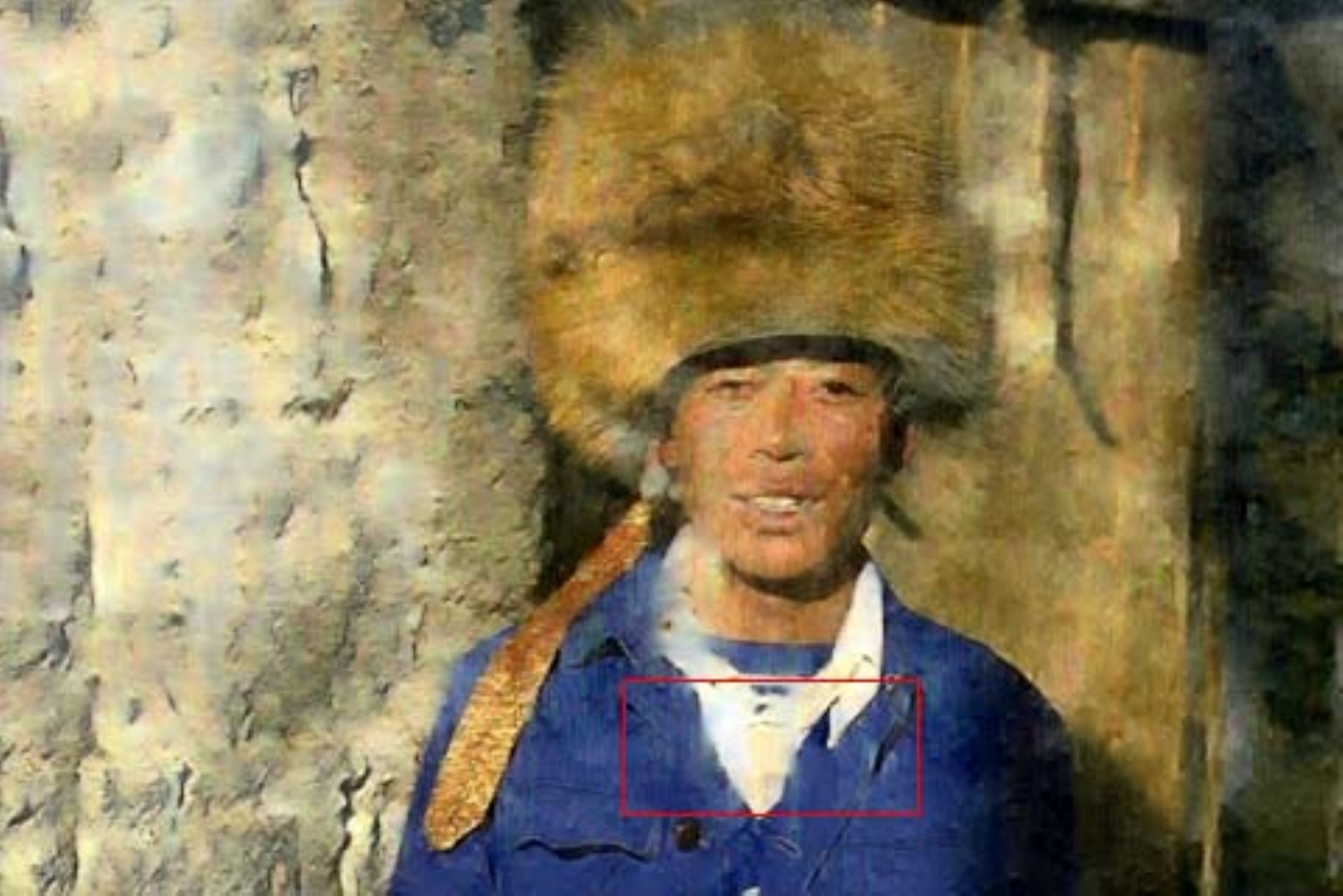} &\hspace{-4mm}
\includegraphics[width = 0.11\linewidth]{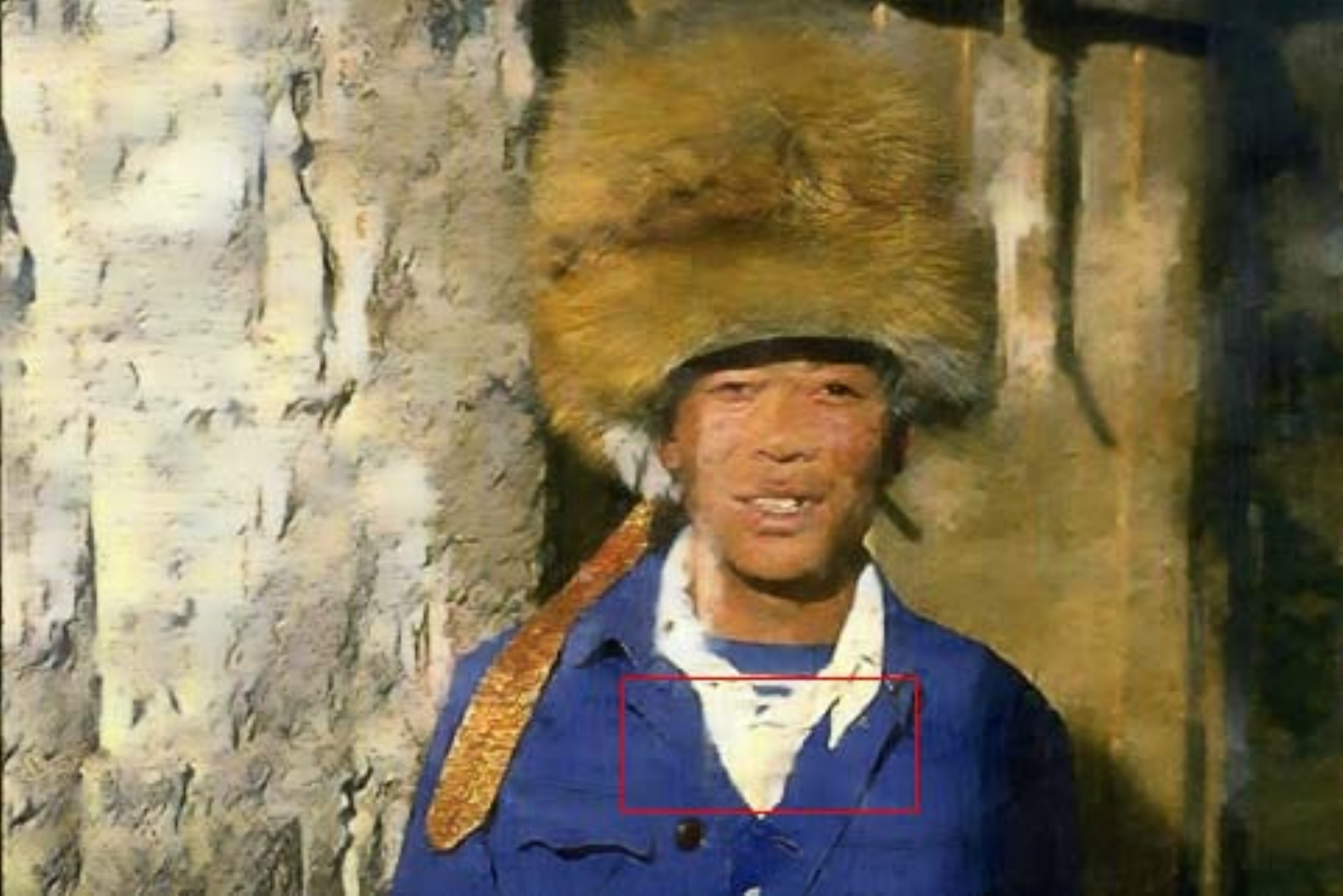} &\hspace{-4mm}
\includegraphics[width = 0.11\linewidth]{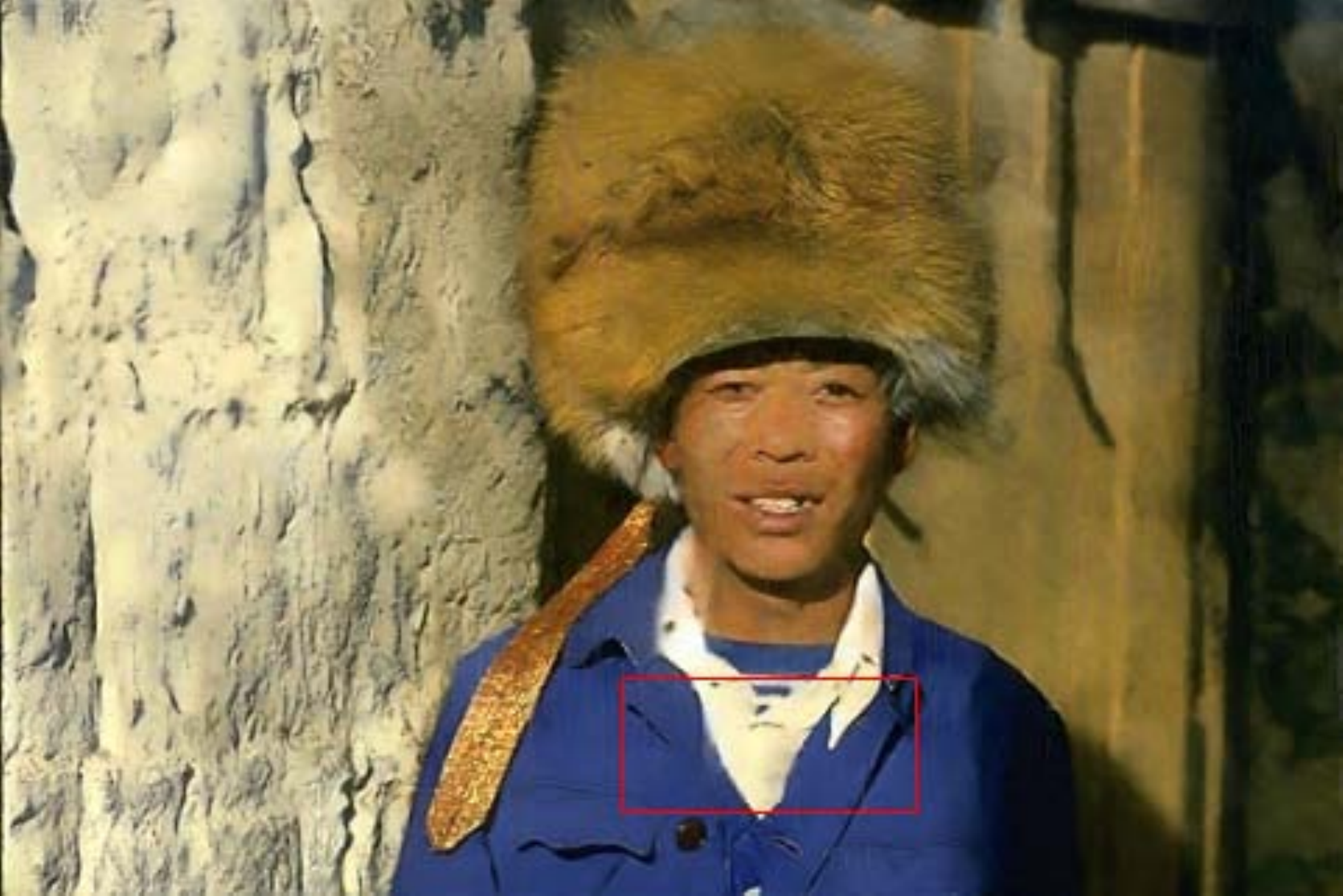} &\hspace{-4mm}
\includegraphics[width = 0.11\linewidth]{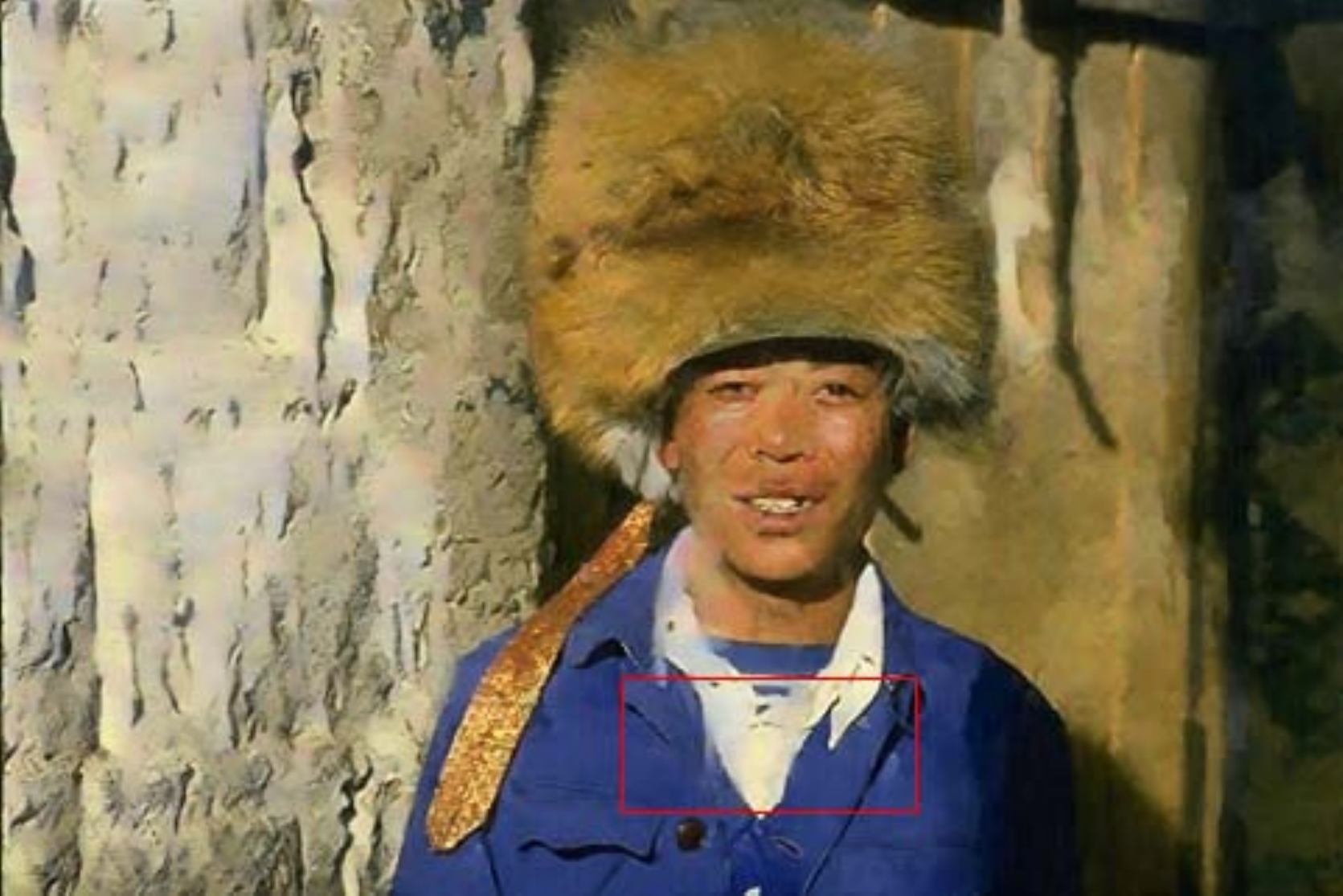} &\hspace{-4mm}
\includegraphics[width = 0.11\linewidth]{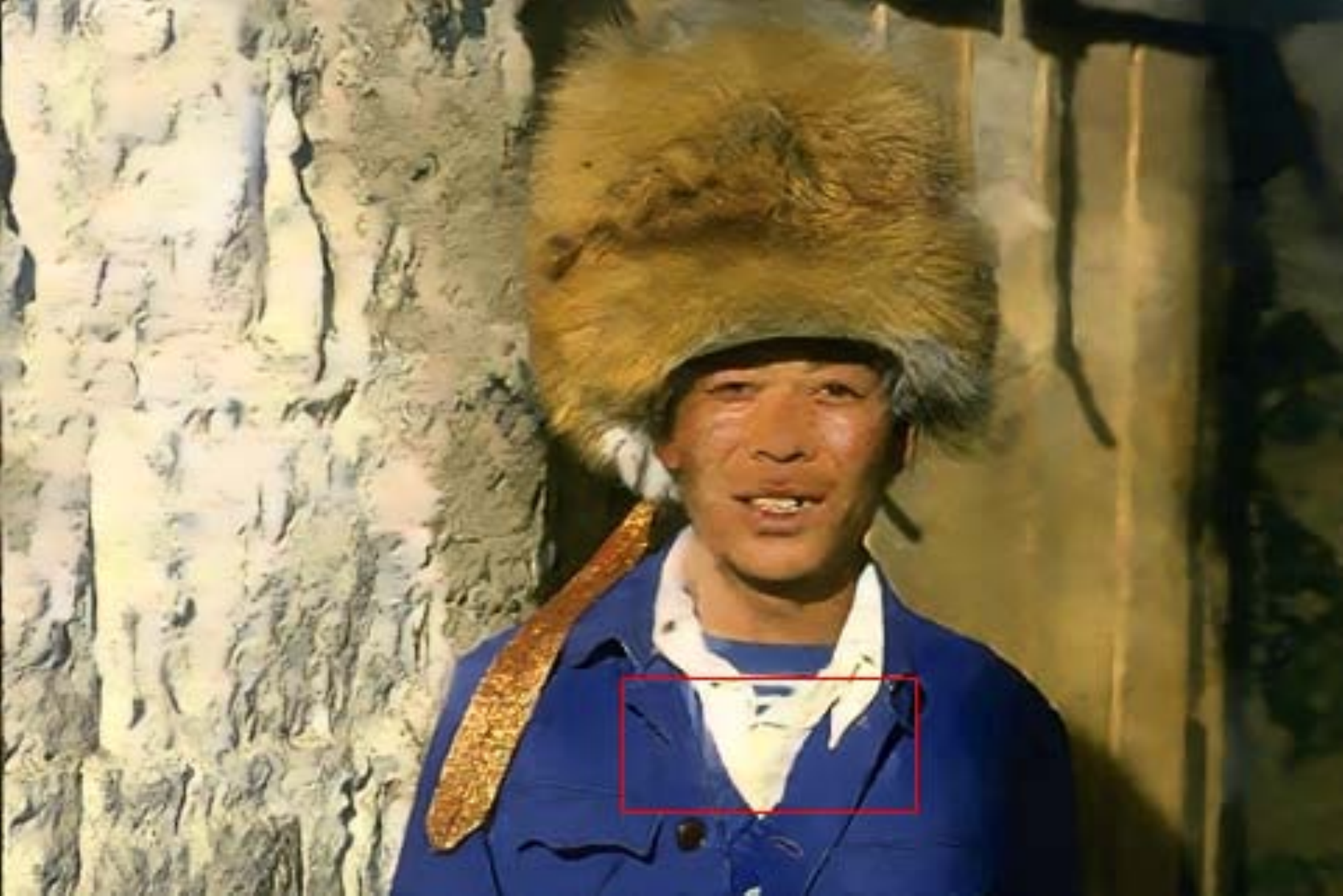}&\hspace{-4mm}
\includegraphics[width = 0.11\linewidth]{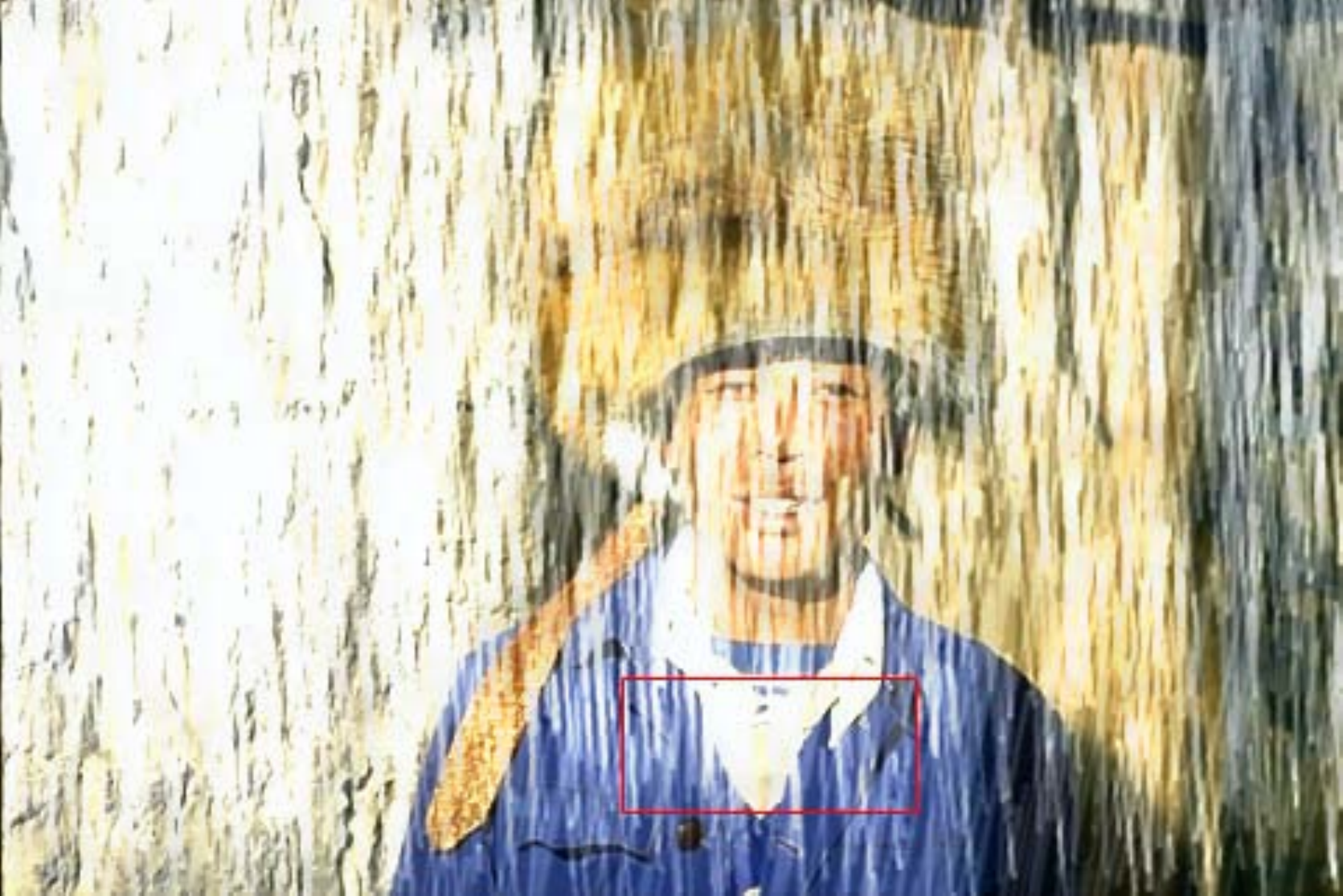} &\hspace{-4mm}
\includegraphics[width = 0.11\linewidth]{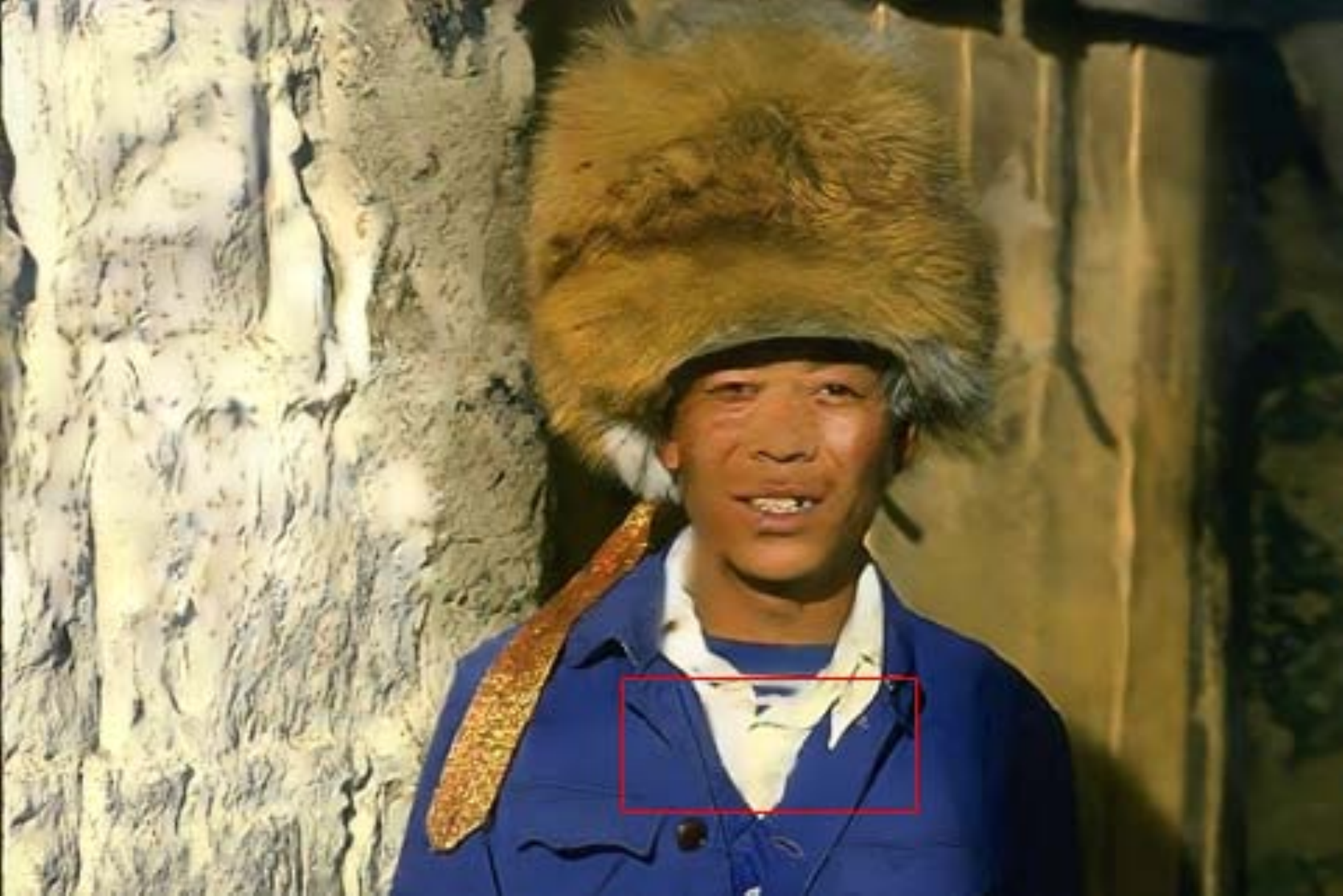} &\hspace{-4mm}
\includegraphics[width = 0.11\linewidth]{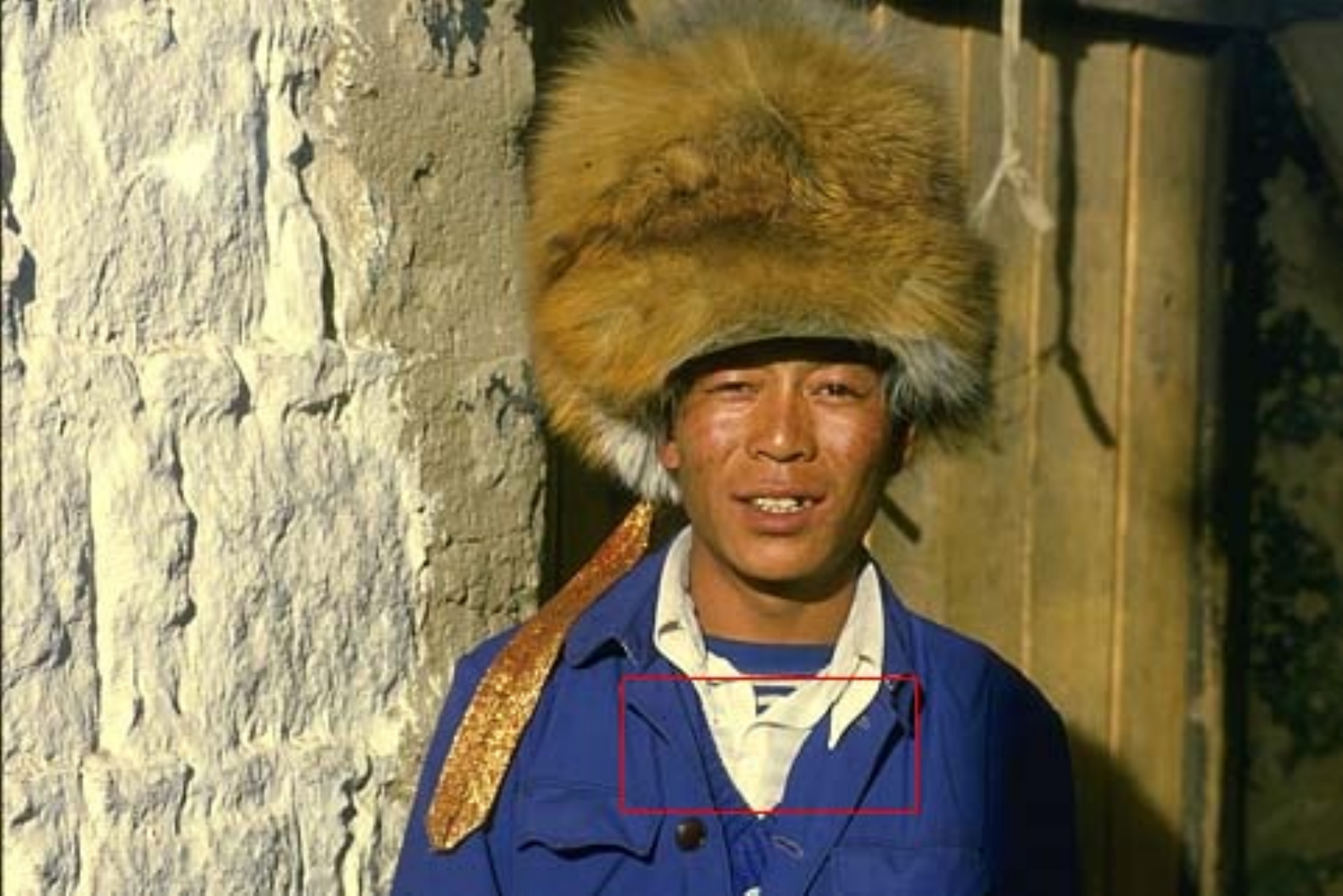}
\\
\includegraphics[width = 0.11\linewidth]{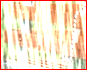} &\hspace{-4mm}
\includegraphics[width = 0.11\linewidth]{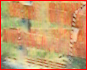} &\hspace{-4mm}
\includegraphics[width = 0.11\linewidth]{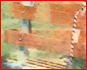} &\hspace{-4mm}
\includegraphics[width = 0.11\linewidth]{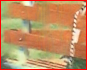} &\hspace{-4mm}
\includegraphics[width = 0.11\linewidth]{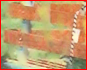} &\hspace{-4mm}
\includegraphics[width = 0.11\linewidth]{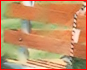} &\hspace{-4mm}
\includegraphics[width = 0.11\linewidth]{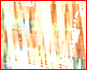} &\hspace{-4mm}
\includegraphics[width = 0.11\linewidth]{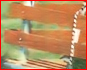} &\hspace{-4mm}
\includegraphics[width = 0.11\linewidth]{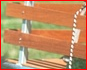}
\\
\includegraphics[width = 0.11\linewidth]{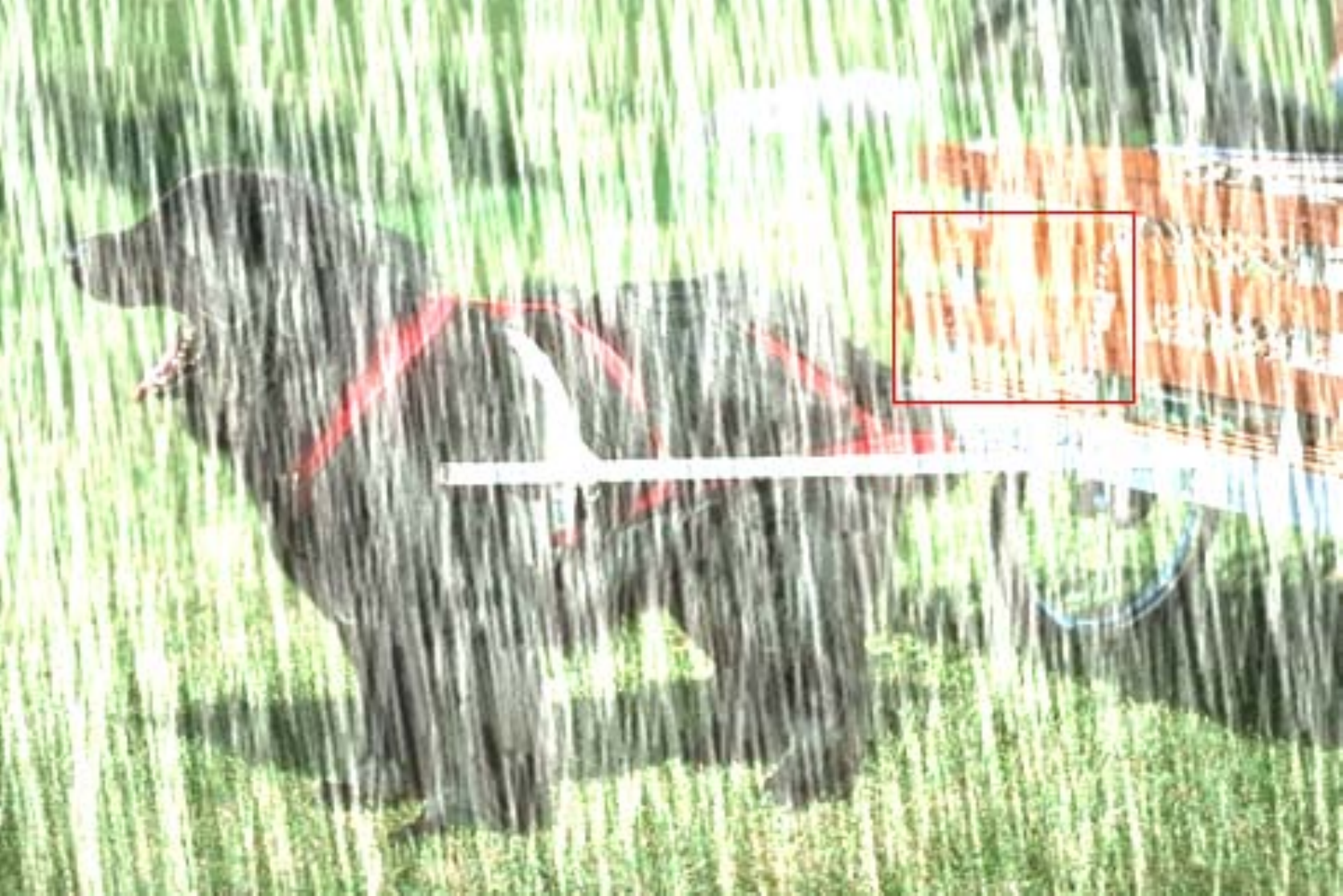} &\hspace{-4mm}
\includegraphics[width = 0.11\linewidth]{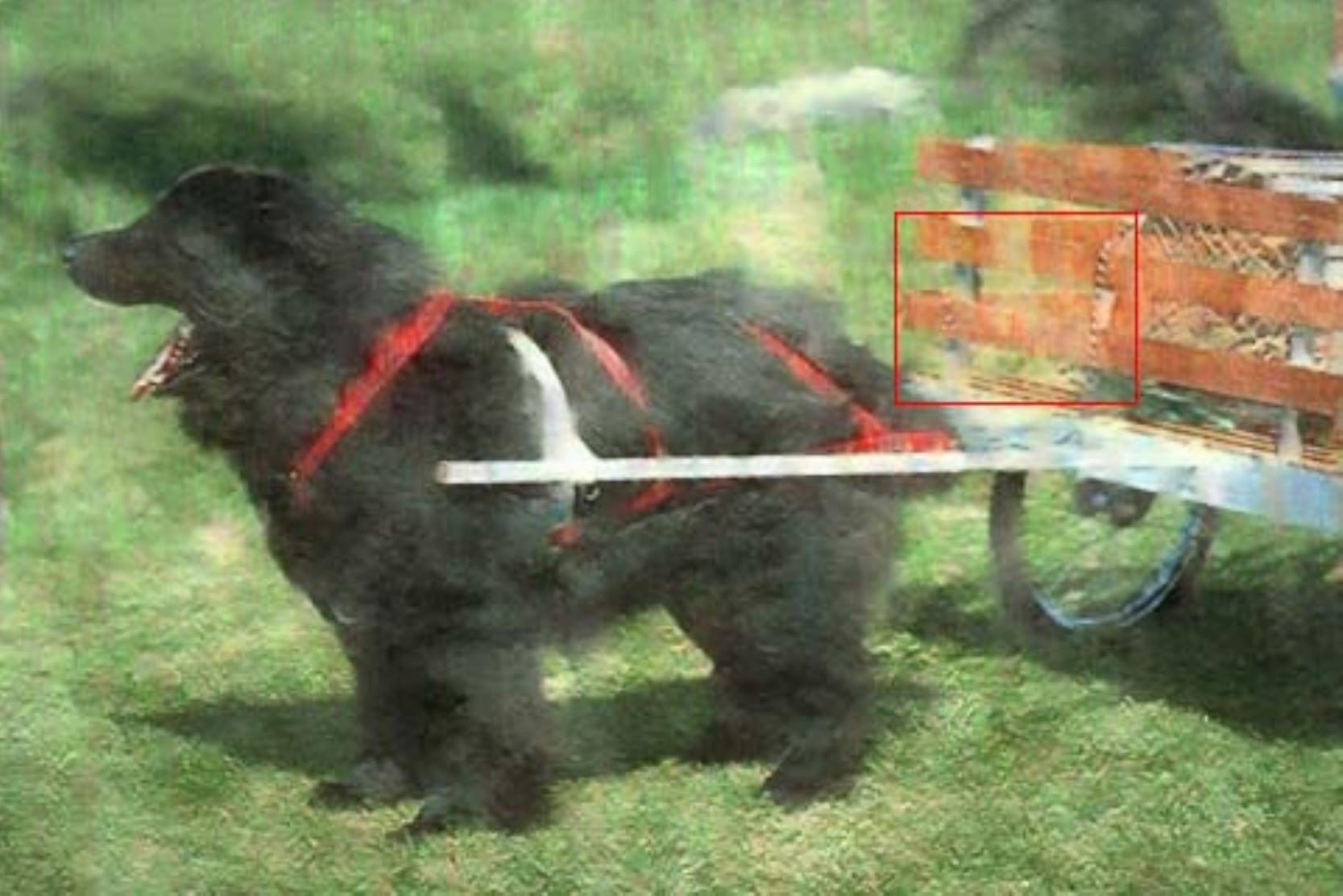} &\hspace{-4mm}
\includegraphics[width = 0.11\linewidth]{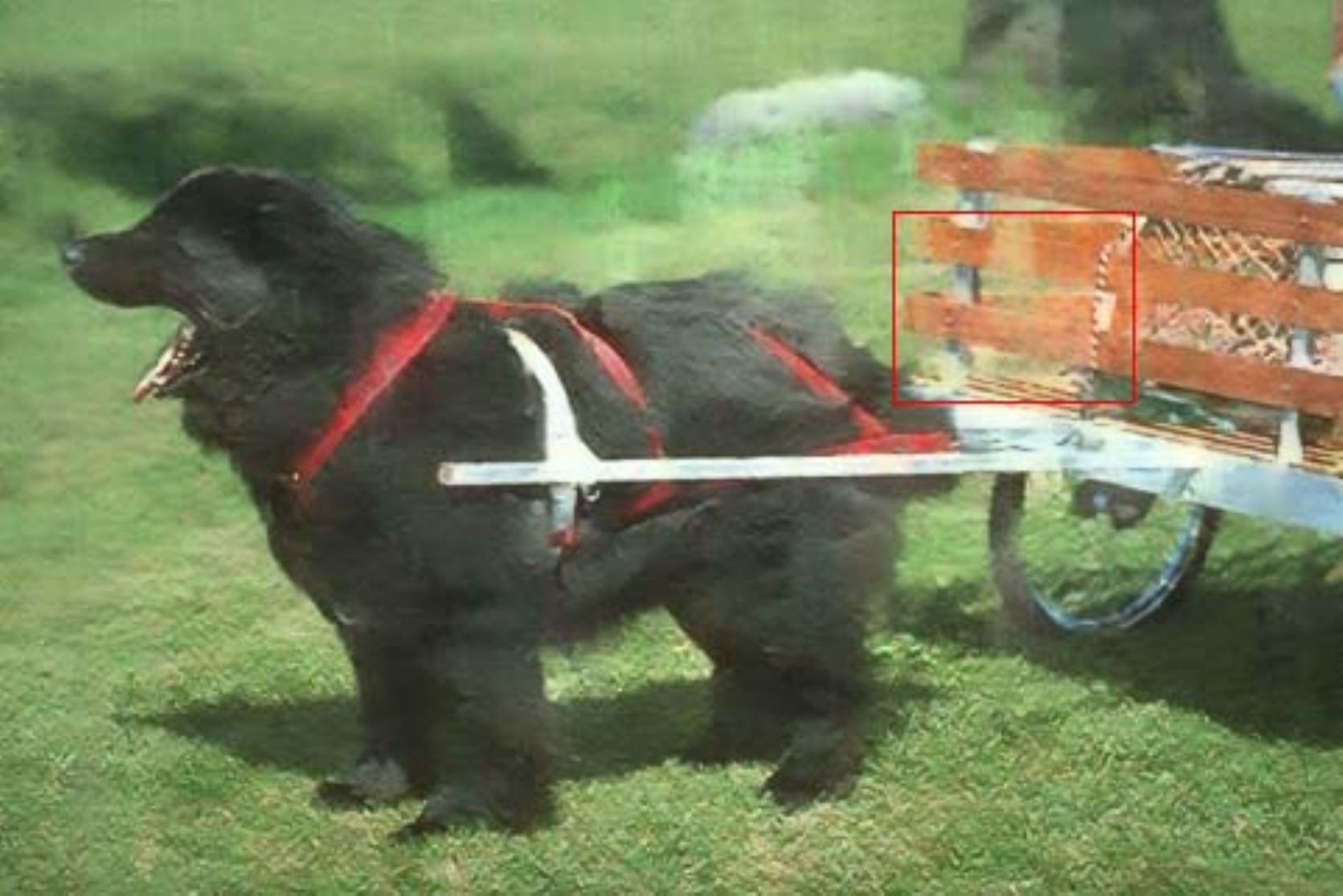} &\hspace{-4mm}
\includegraphics[width = 0.11\linewidth]{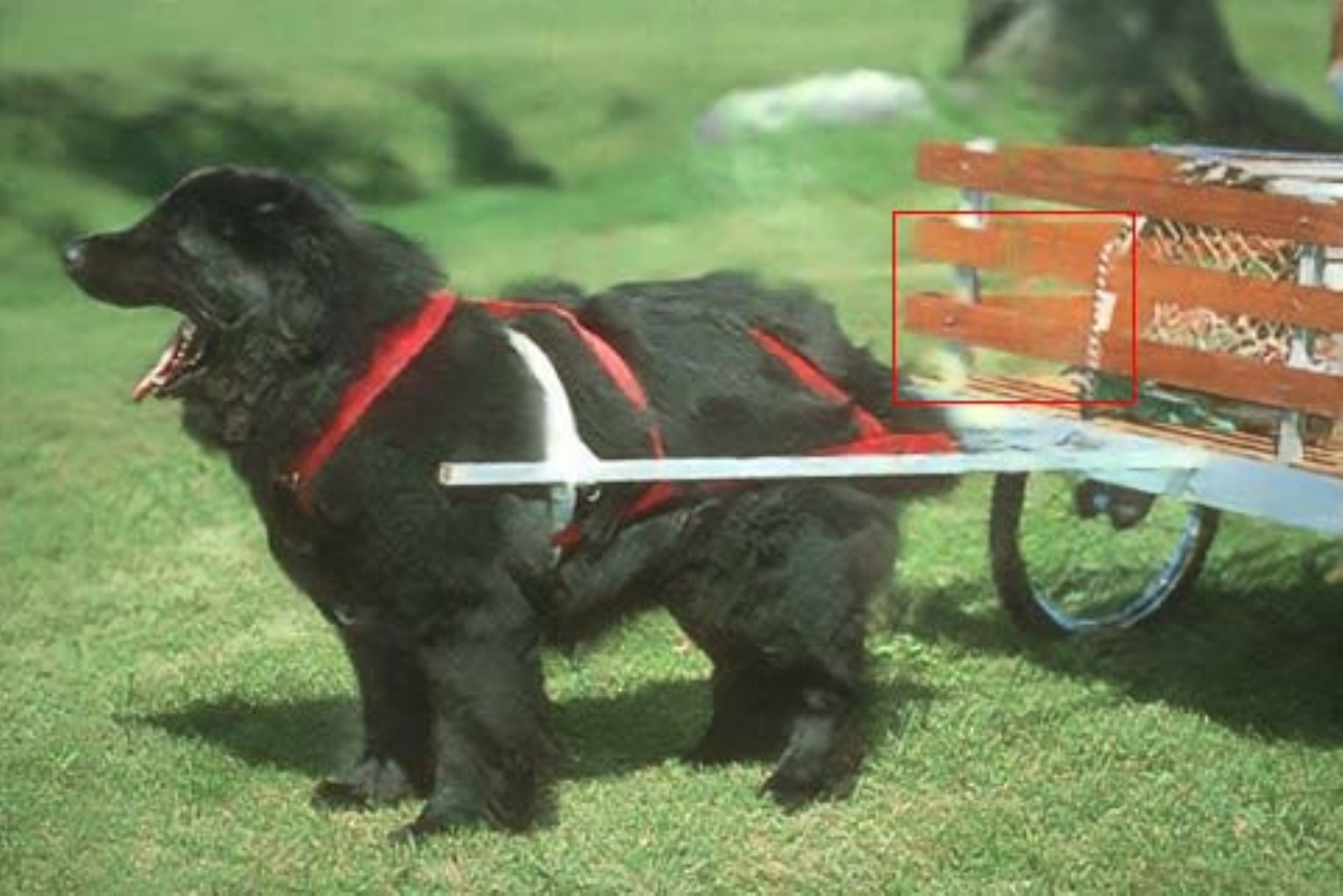} &\hspace{-4mm}
\includegraphics[width = 0.11\linewidth]{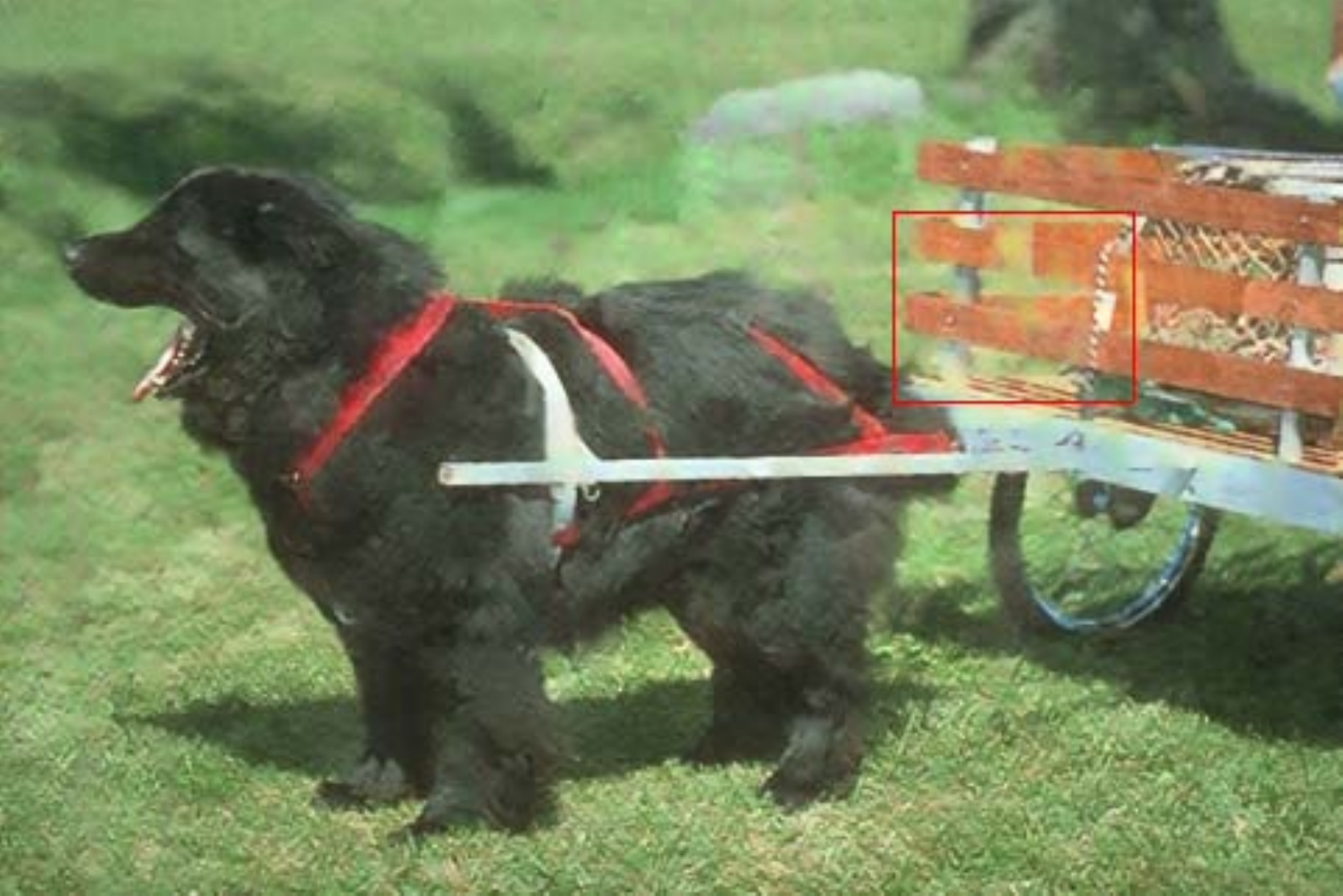} &\hspace{-4mm}
\includegraphics[width = 0.11\linewidth]{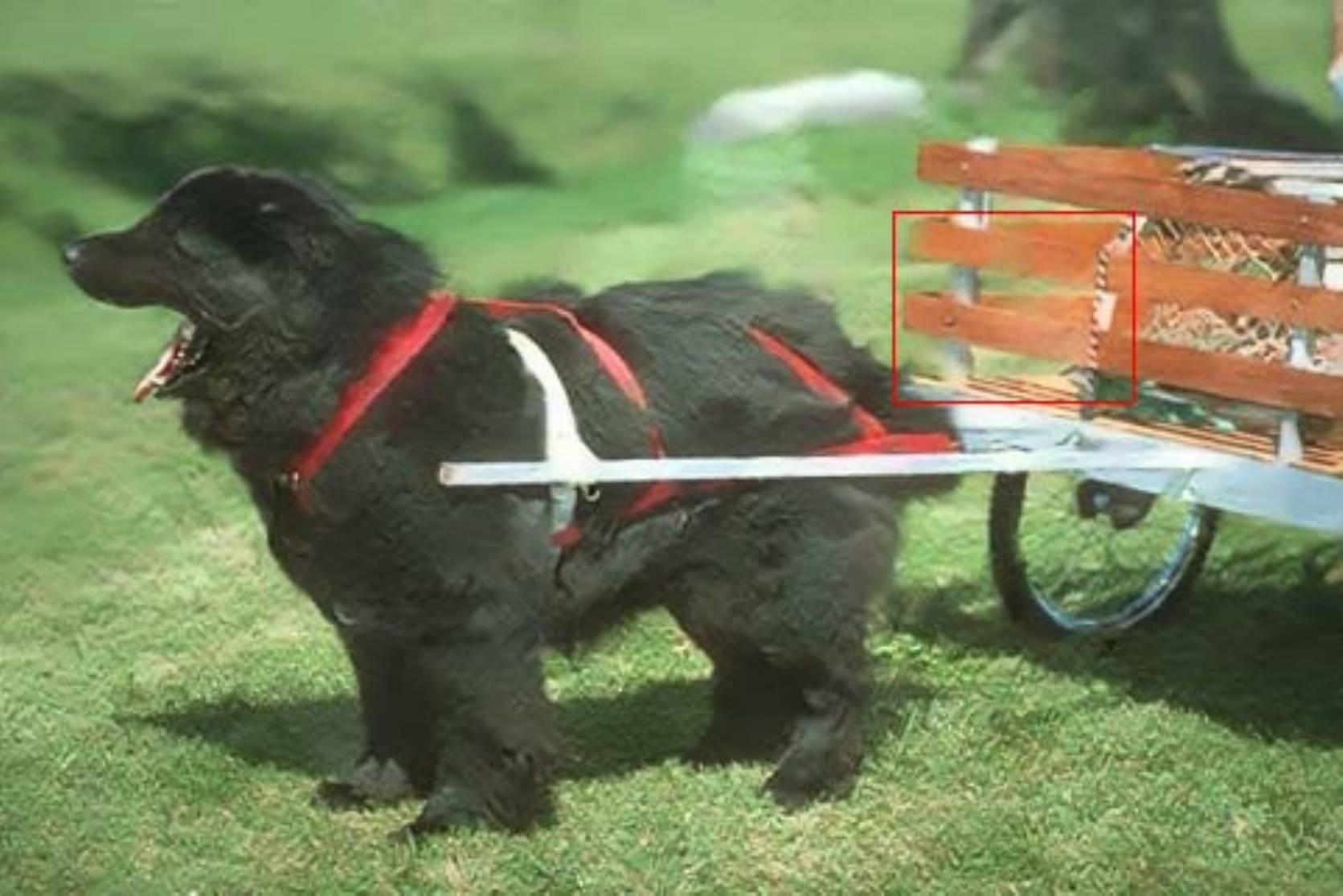} &\hspace{-4mm}
\includegraphics[width = 0.11\linewidth]{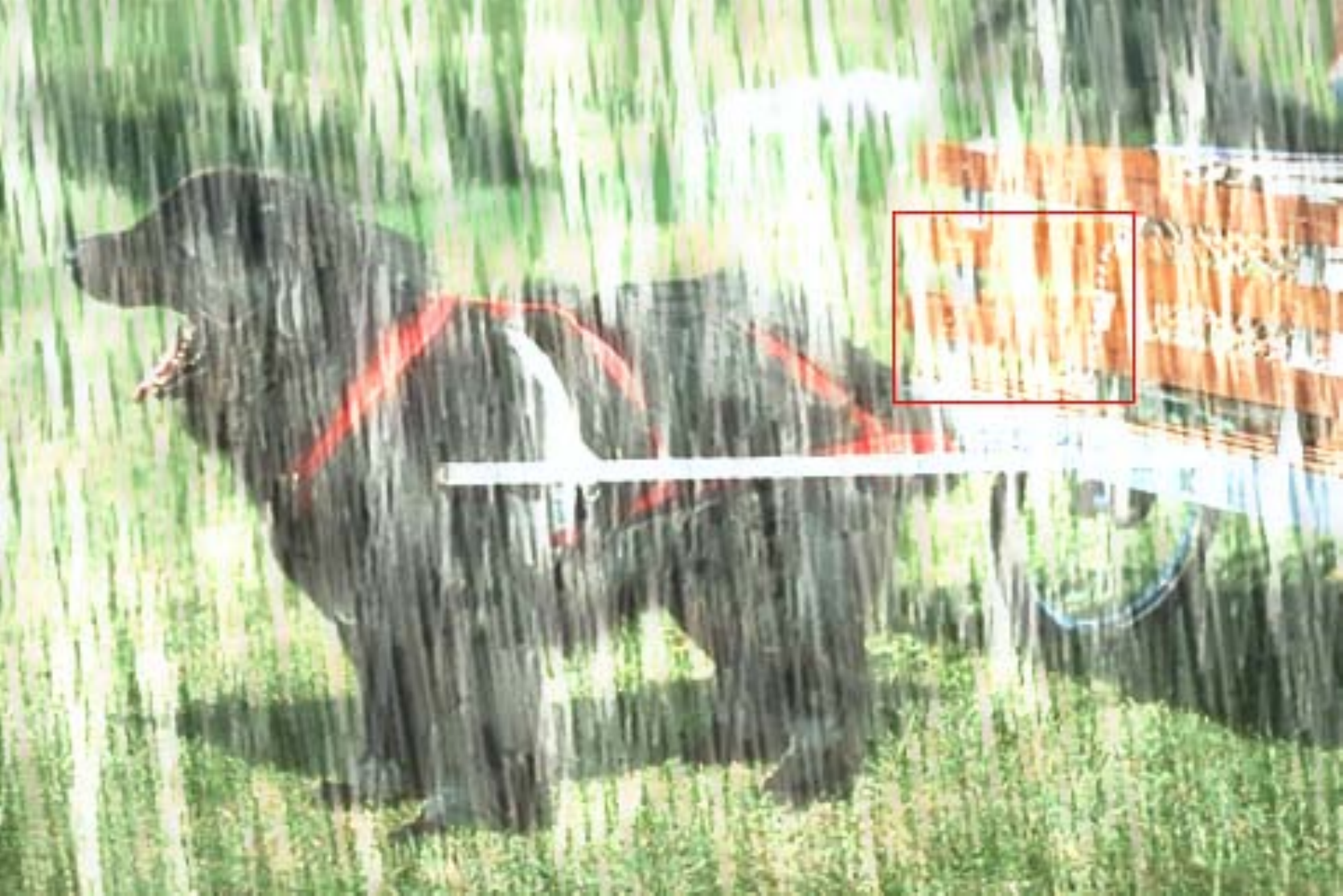} &\hspace{-4mm}
\includegraphics[width = 0.11\linewidth]{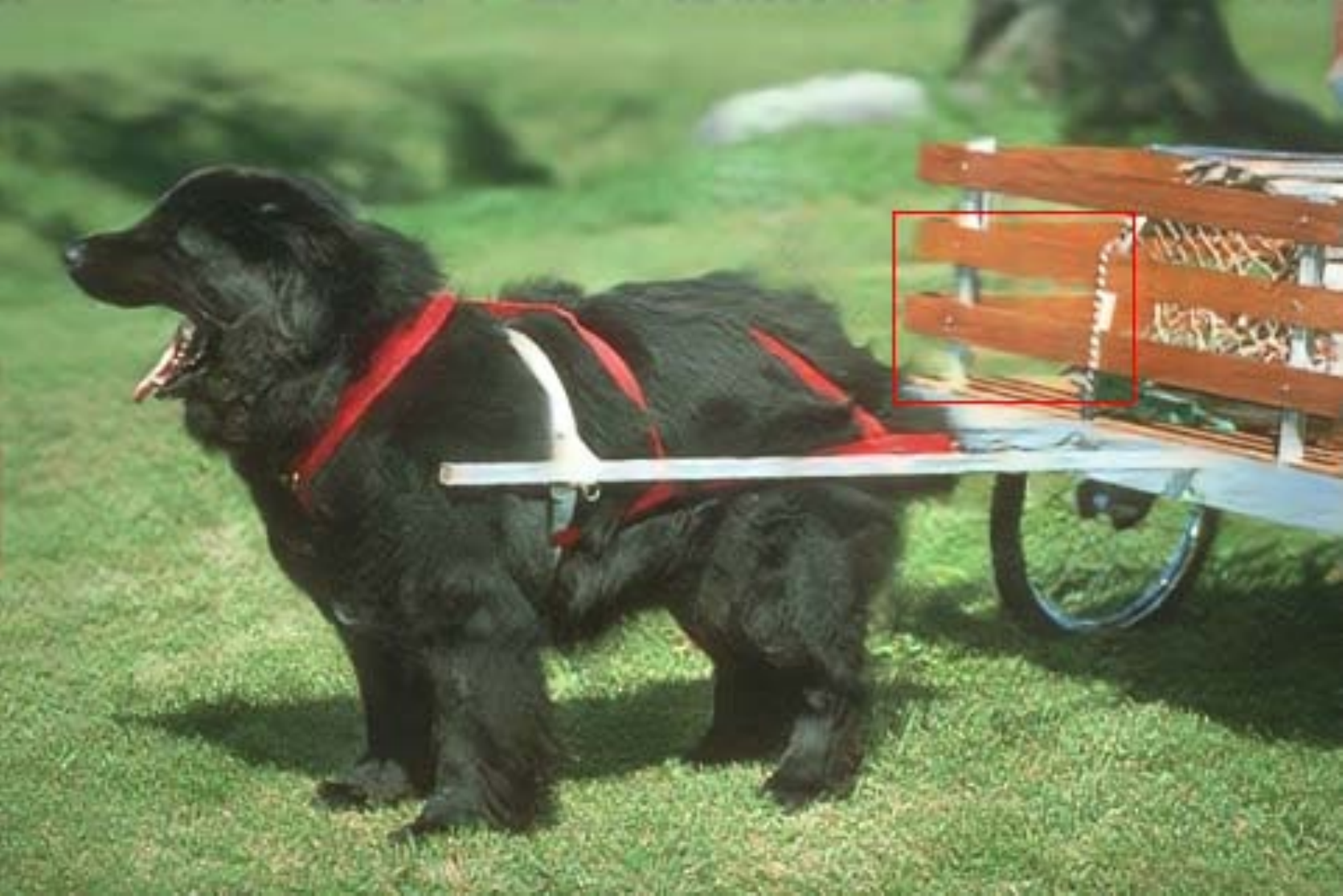} &\hspace{-4mm}
\includegraphics[width = 0.11\linewidth]{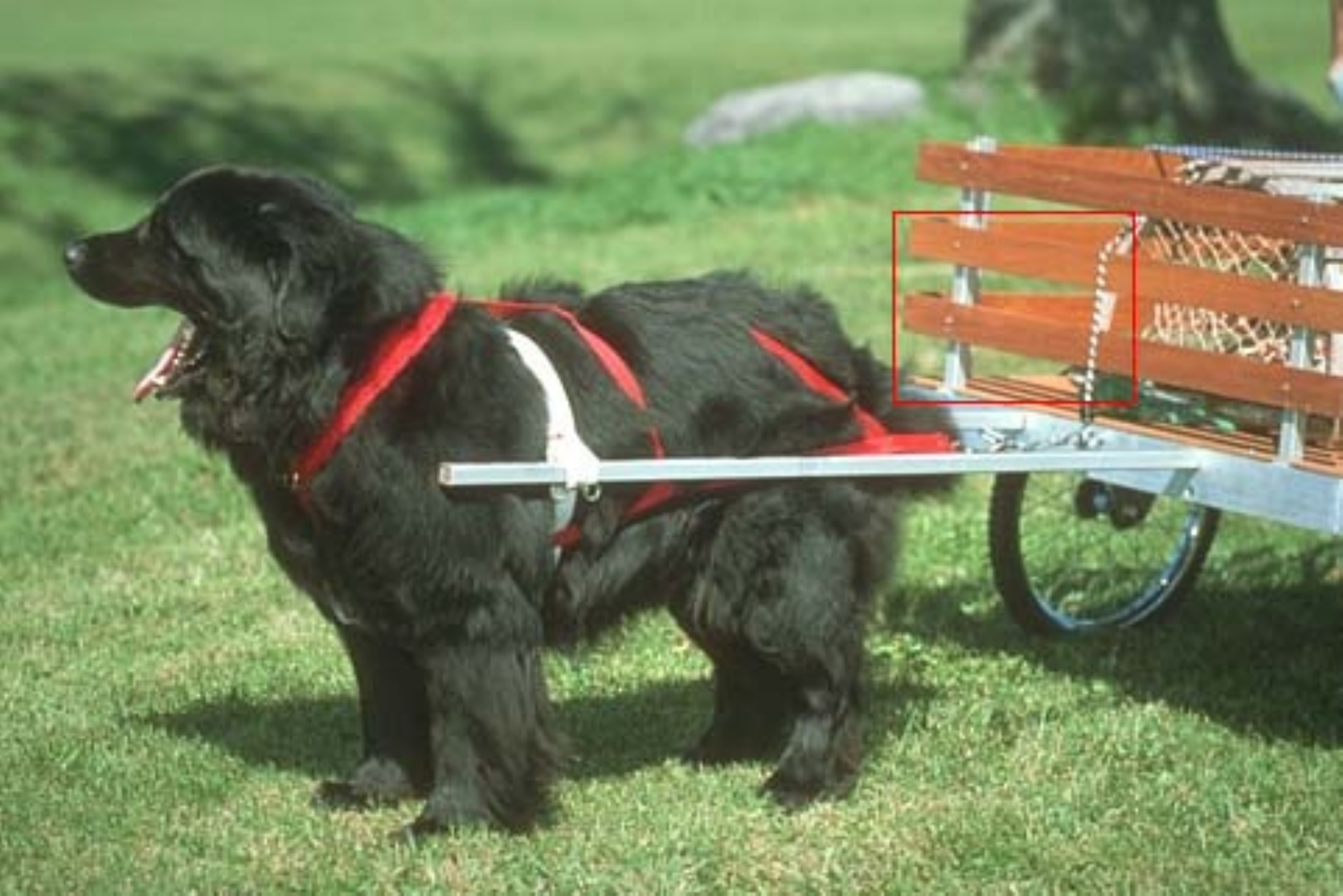}
\\
(a) Input&\hspace{-4mm} (b) DDN&\hspace{-4mm} (c) RESCAN&\hspace{-4mm} (d) NLEDN  &\hspace{-4mm} (e) REHEN&\hspace{-4mm} (f) PreNet&\hspace{-4mm} (g) SSIR&\hspace{-4mm} (h) DCSFN &\hspace{-4mm} (i) GT
\\
\end{tabular}
\end{center}
\caption{The results from synthetic datasets.
Our method restores better deraining results, while other method always hand down some artifacts or rain.
Especially, SSIR~\cite{Derain-cvpr19-semi} fails to restore rain-free images and its results are unacceptable.
}
\label{fig:deraining-syn-example}
\end{figure*}
In this section, we conduct extensive experiments to demonstrate the effectiveness of the proposed method on four synthetic widely used datasets and a lot of real-world images.
Six state-of-the-art methods are compared in this paper, DDN~\cite{derain_ddn_fu} (CVPR17), RESCAN~\cite{derain_rescan_li} (ECCV18), NLEDN~\cite{derain_nledn_li}~(ACM MM18), REHEN~\cite{derain-acmmm19-rehen}~(ACM MM19), \\
PreNet~\cite{derain_prenet_Ren_2019_CVPR}~(CVPR19), SSIR~\cite{Derain-cvpr19-semi}~(CVPR19).
Next, we will introduce detailedly the datasets and measurements in Sec.~\ref{sec:Datasets and Measurements}, implementing details in Sec.~\ref{sec:Implementing Details}, results on synthetic datasets in Sec.~\ref{sec:Results on Synthetic Datasets}, results on real-world datasets in Sec.~\ref{sec:Results on Real-world Datasets} and ablation study in Sec.~\ref{sec:Ablation Study}.
\subsection{Datasets and Measurements}\label{sec:Datasets and Measurements}
\begin{figure*}[!t]
\begin{center}
\begin{tabular}{cccccccc}
\includegraphics[width = 0.123\linewidth]{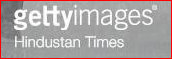} &\hspace{-4mm}
\includegraphics[width = 0.123\linewidth]{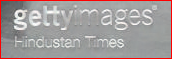} &\hspace{-4mm}
\includegraphics[width = 0.123\linewidth]{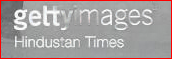} &\hspace{-4mm}
\includegraphics[width = 0.123\linewidth]{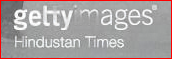} &\hspace{-4mm}
\includegraphics[width = 0.123\linewidth]{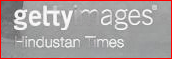} &\hspace{-4mm}
\includegraphics[width = 0.123\linewidth]{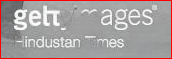} &\hspace{-4mm}
\includegraphics[width = 0.123\linewidth]{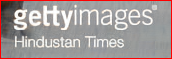} &\hspace{-4mm}
\includegraphics[width = 0.123\linewidth]{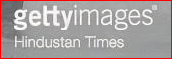}
\\
\includegraphics[width = 0.123\linewidth]{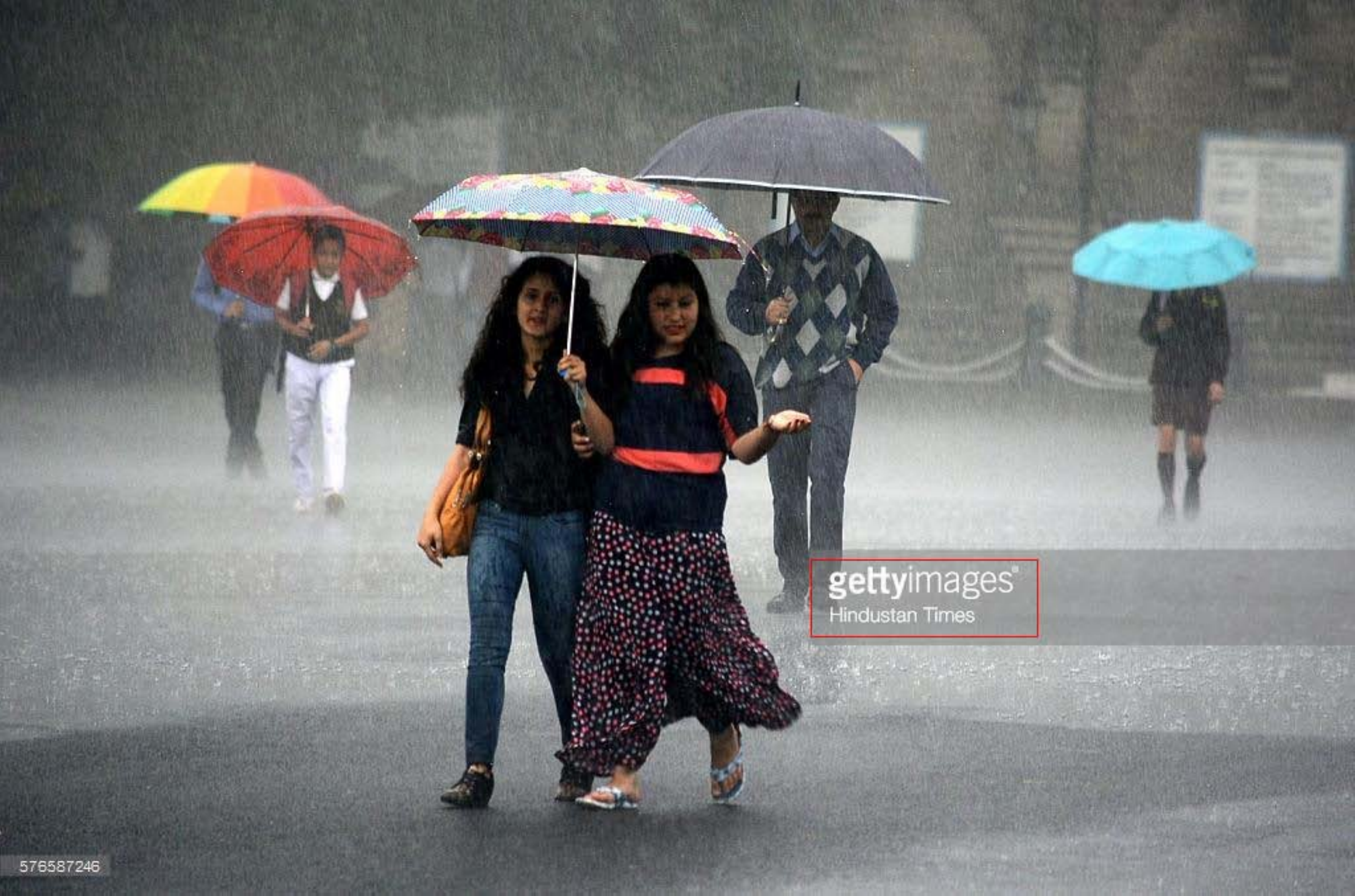} &\hspace{-4mm}
\includegraphics[width = 0.123\linewidth]{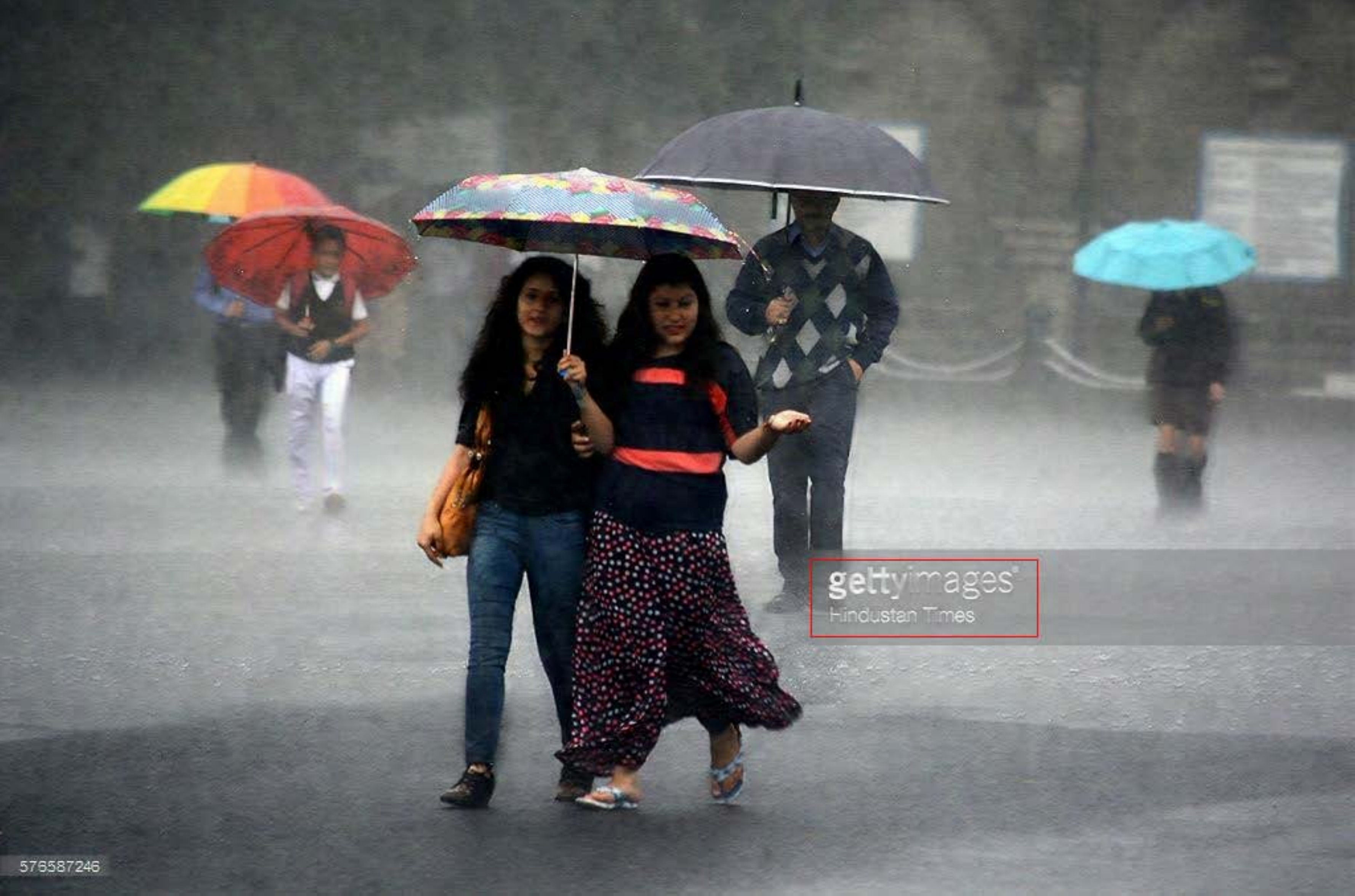} &\hspace{-4mm}
\includegraphics[width = 0.123\linewidth]{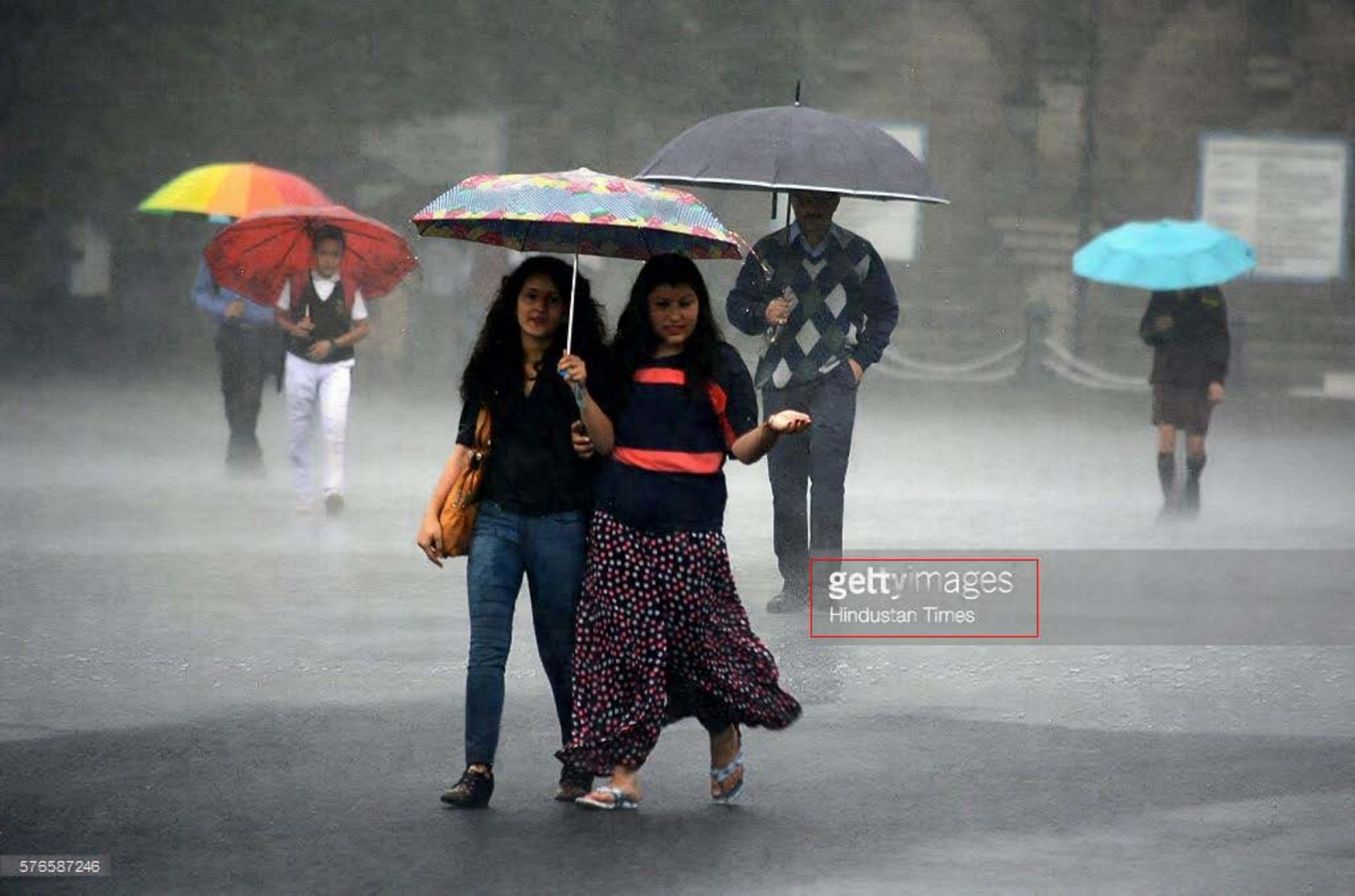} &\hspace{-4mm}
\includegraphics[width = 0.123\linewidth]{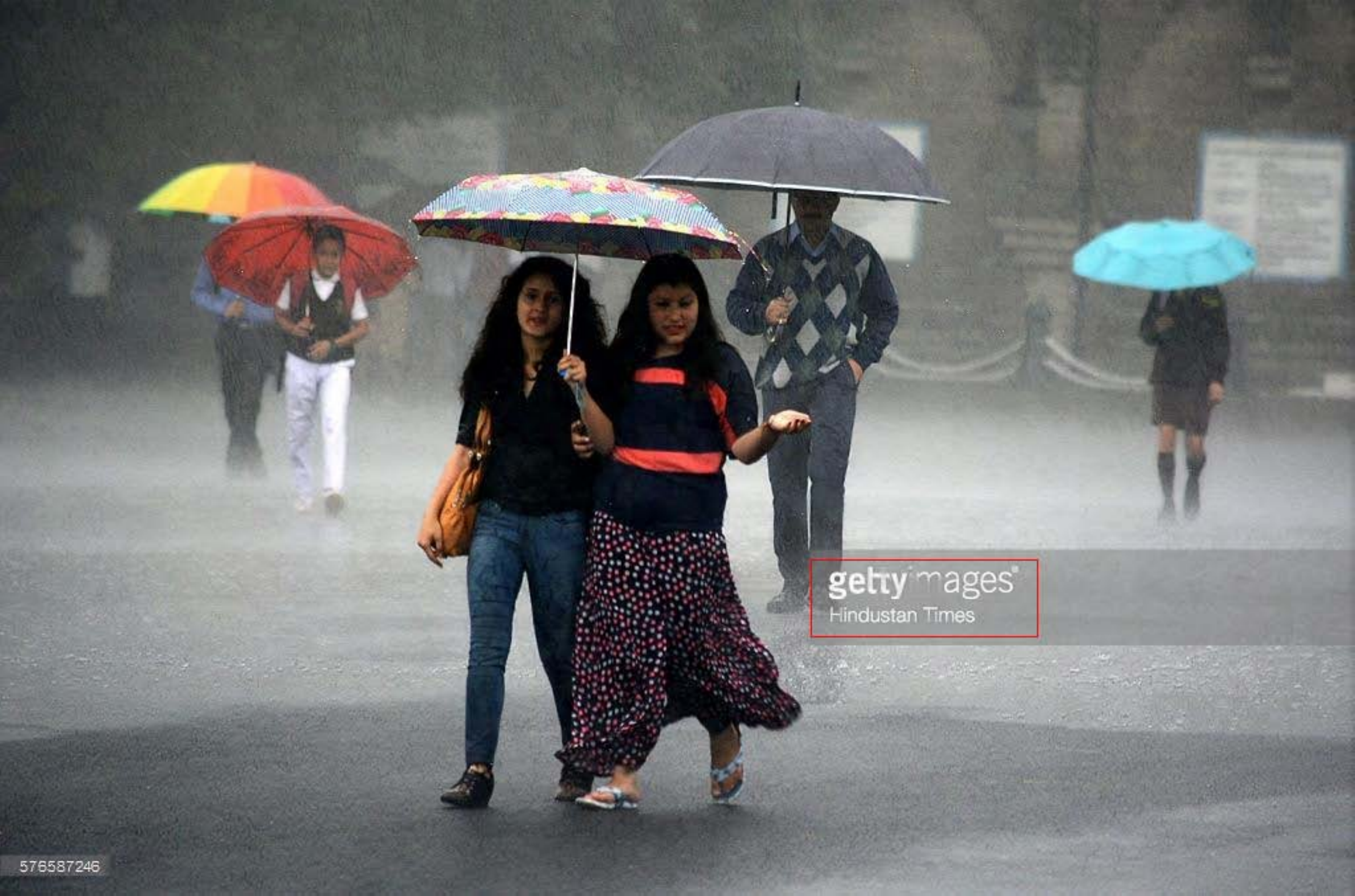} &\hspace{-4mm}
\includegraphics[width = 0.123\linewidth]{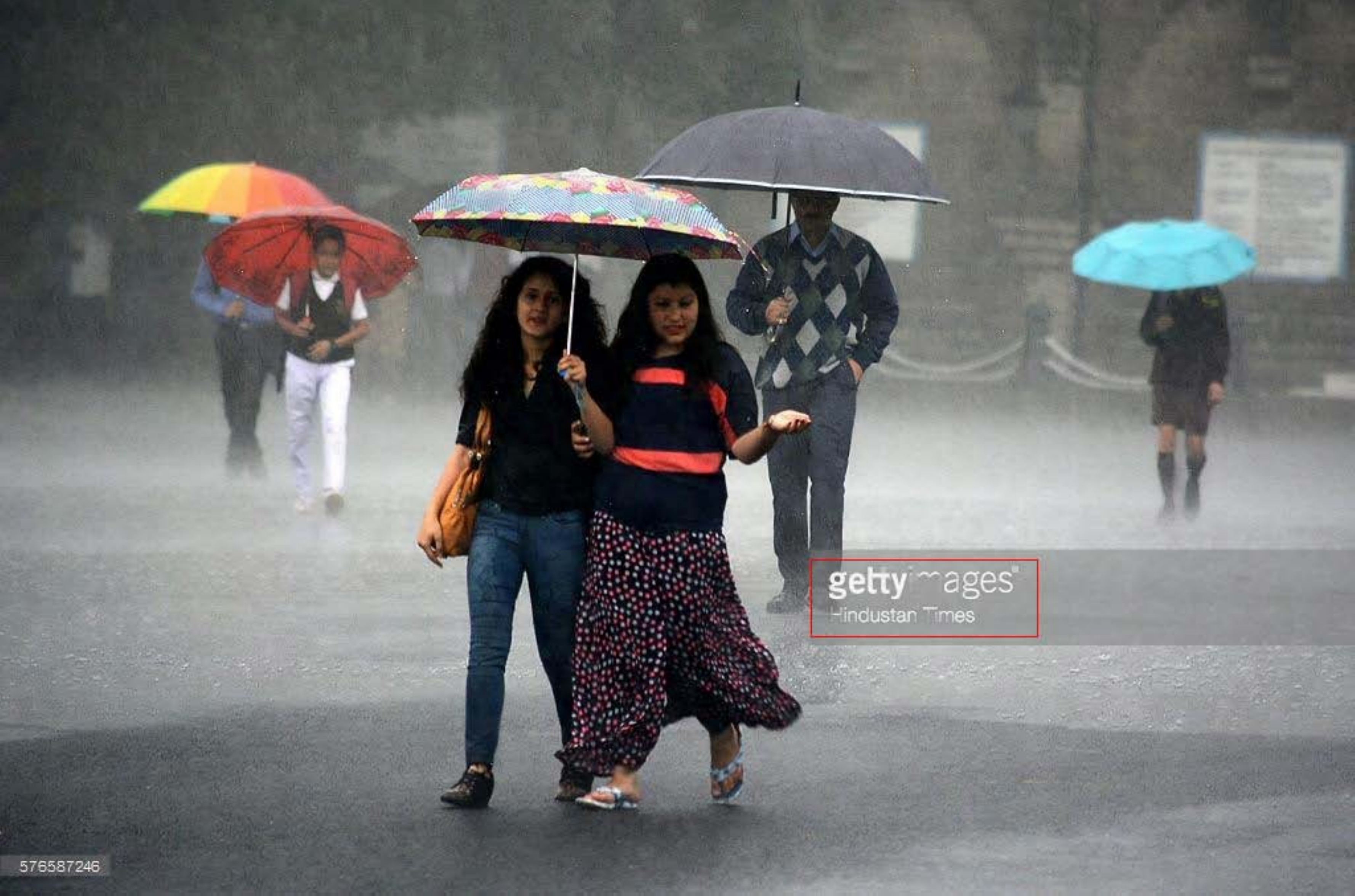} &\hspace{-4mm}
\includegraphics[width = 0.123\linewidth]{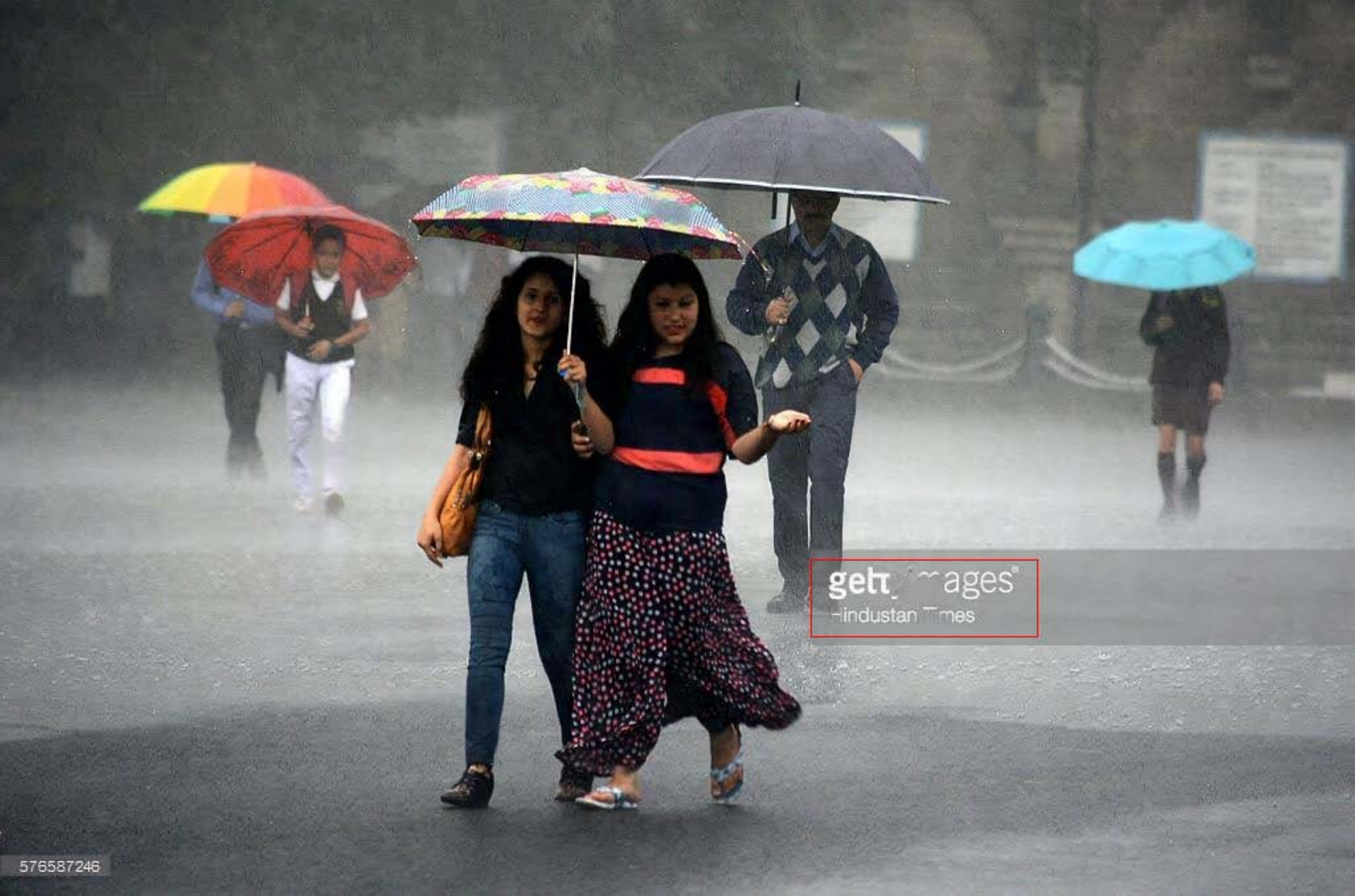} &\hspace{-4mm}
\includegraphics[width = 0.123\linewidth]{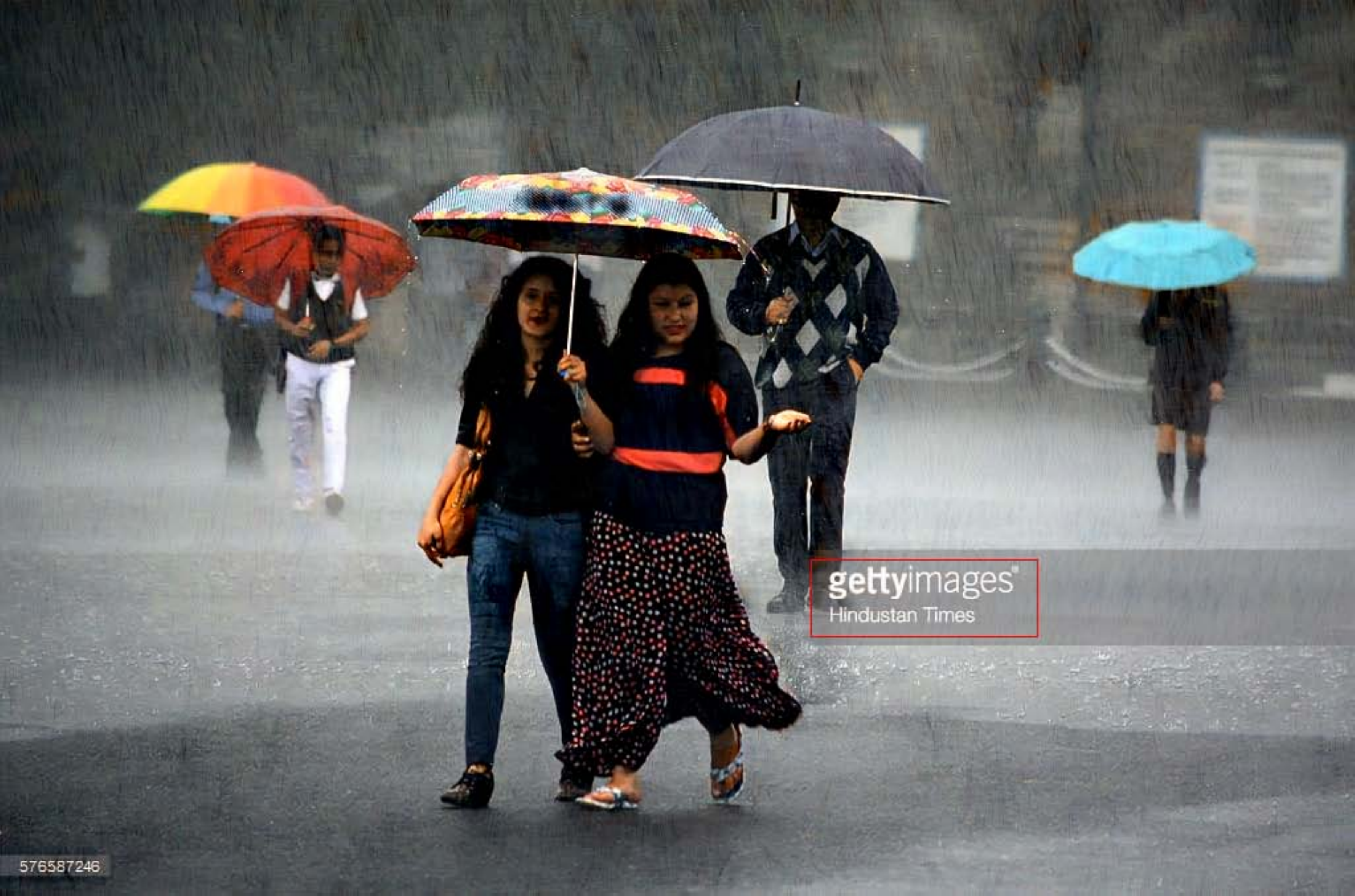} &\hspace{-4mm}
\includegraphics[width = 0.123\linewidth]{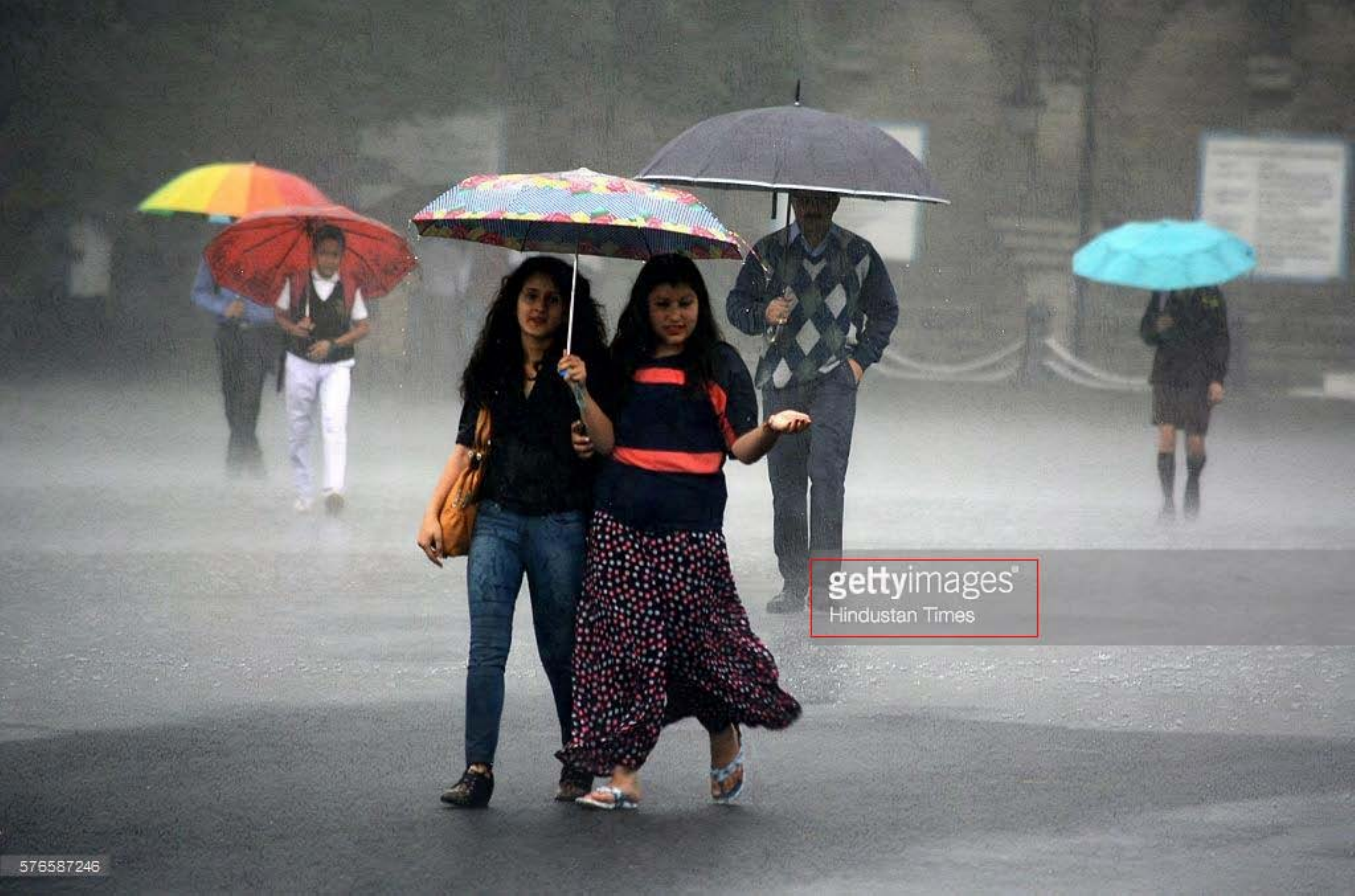}
\\
\includegraphics[width = 0.123\linewidth]{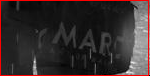} &\hspace{-4mm}
\includegraphics[width = 0.123\linewidth]{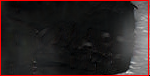} &\hspace{-4mm}
\includegraphics[width = 0.123\linewidth]{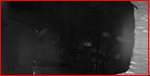} &\hspace{-4mm}
\includegraphics[width = 0.123\linewidth]{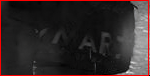} &\hspace{-4mm}
\includegraphics[width = 0.123\linewidth]{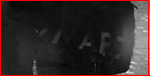} &\hspace{-4mm}
\includegraphics[width = 0.123\linewidth]{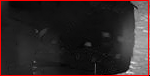} &\hspace{-4mm}
\includegraphics[width = 0.123\linewidth]{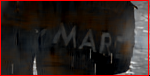} &\hspace{-4mm}
\includegraphics[width = 0.123\linewidth]{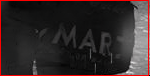}
\\
\includegraphics[width = 0.123\linewidth]{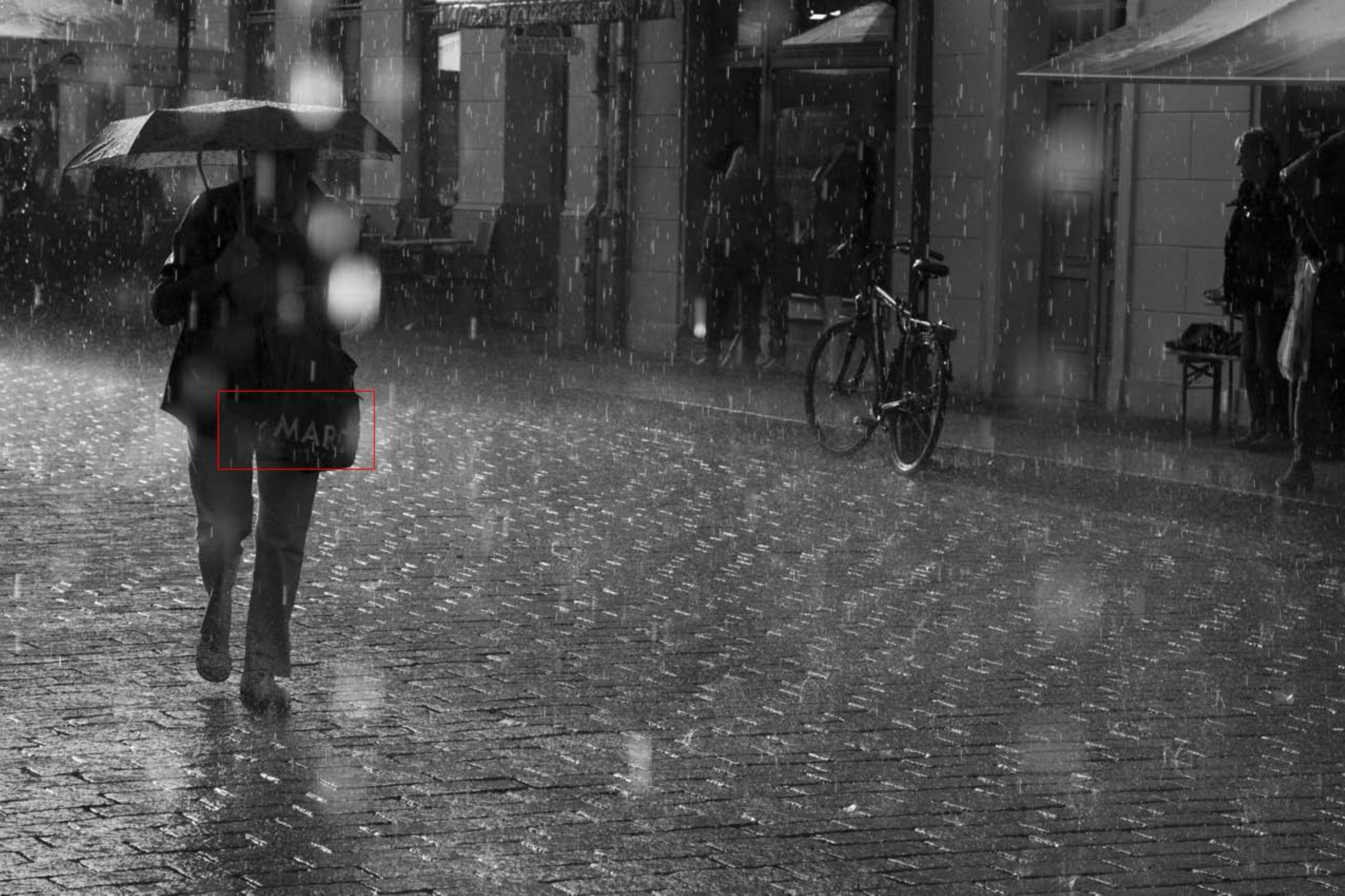} &\hspace{-4mm}
\includegraphics[width = 0.123\linewidth]{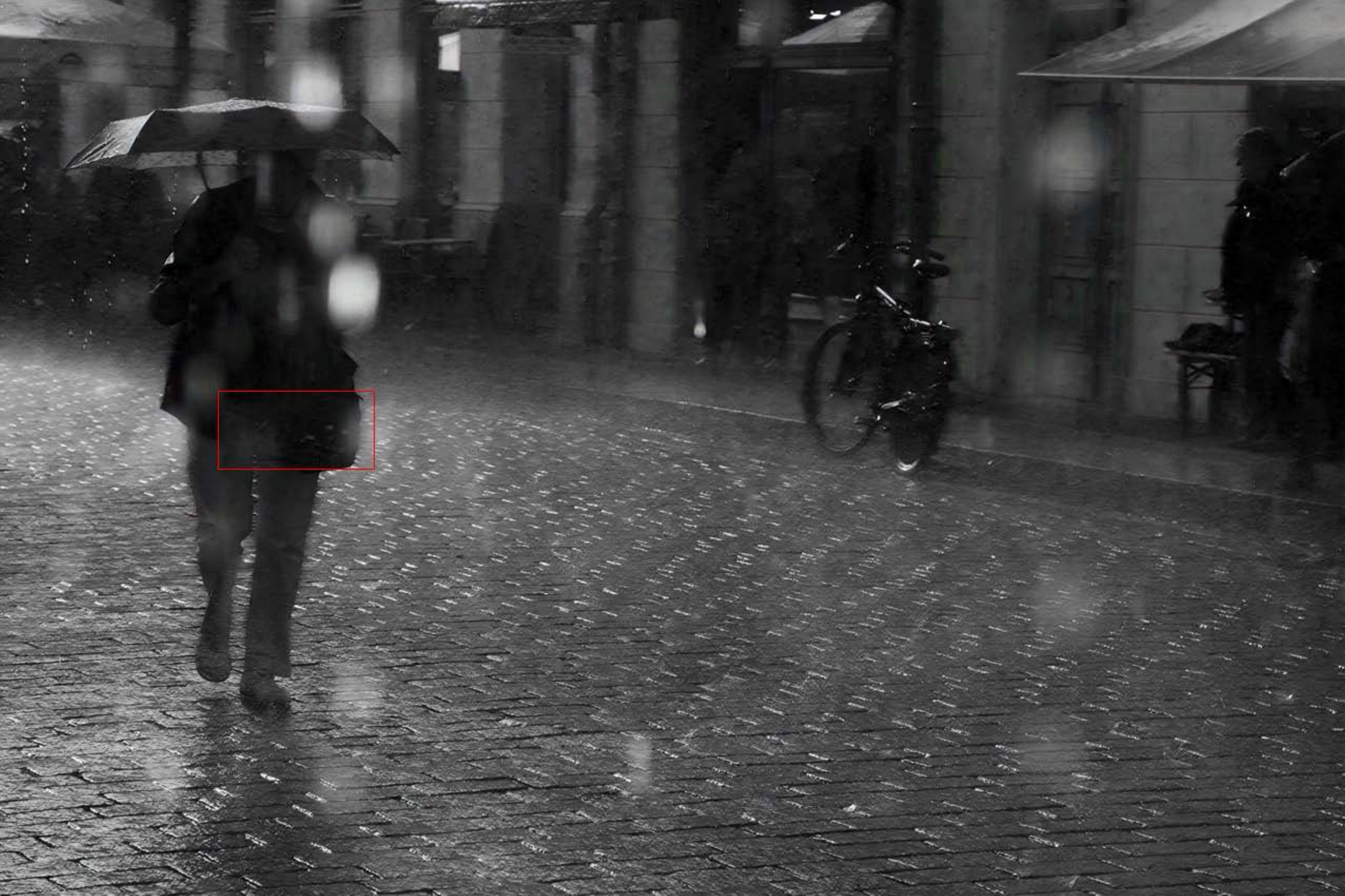} &\hspace{-4mm}
\includegraphics[width = 0.123\linewidth]{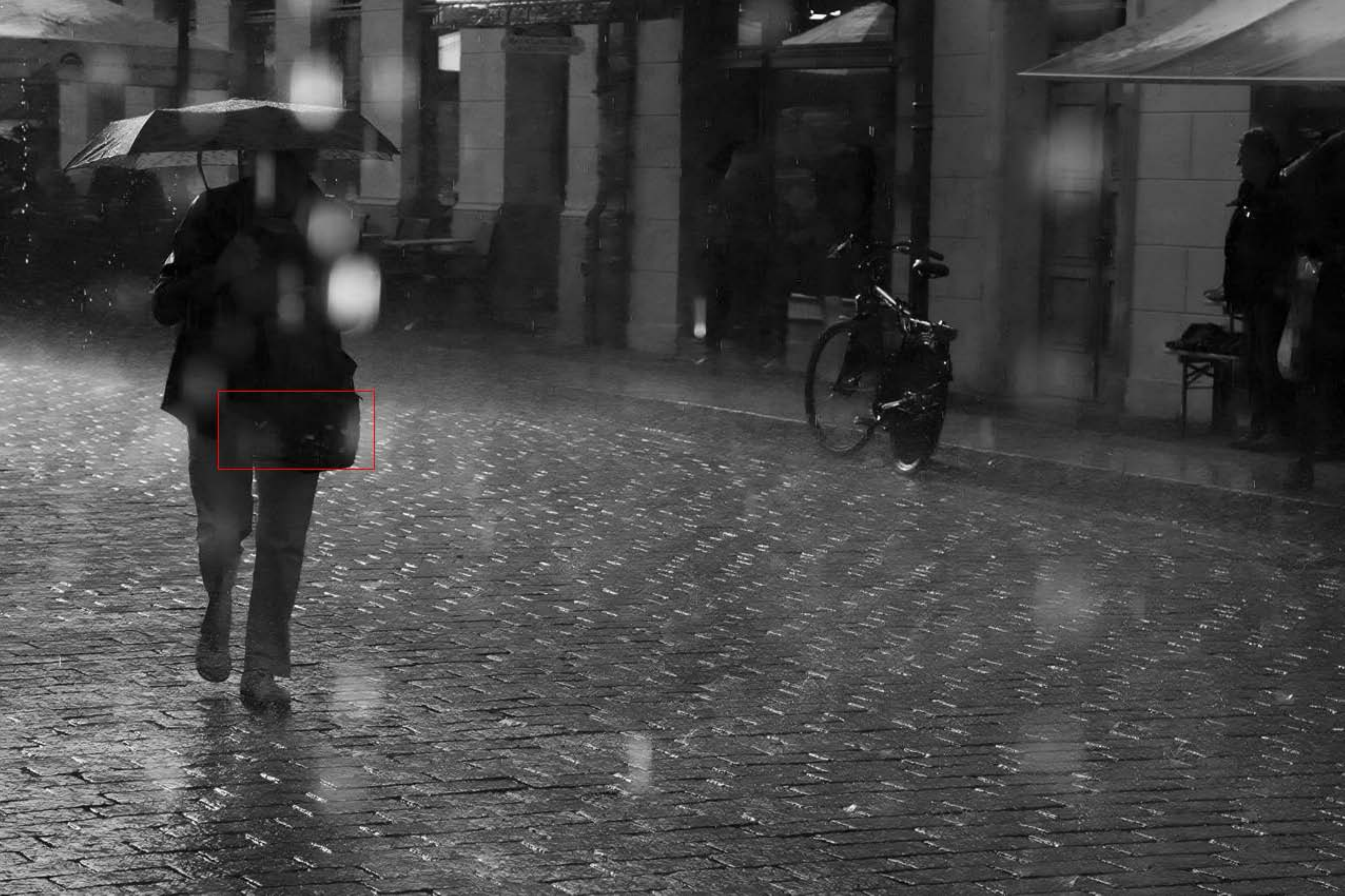} &\hspace{-4mm}
\includegraphics[width = 0.123\linewidth]{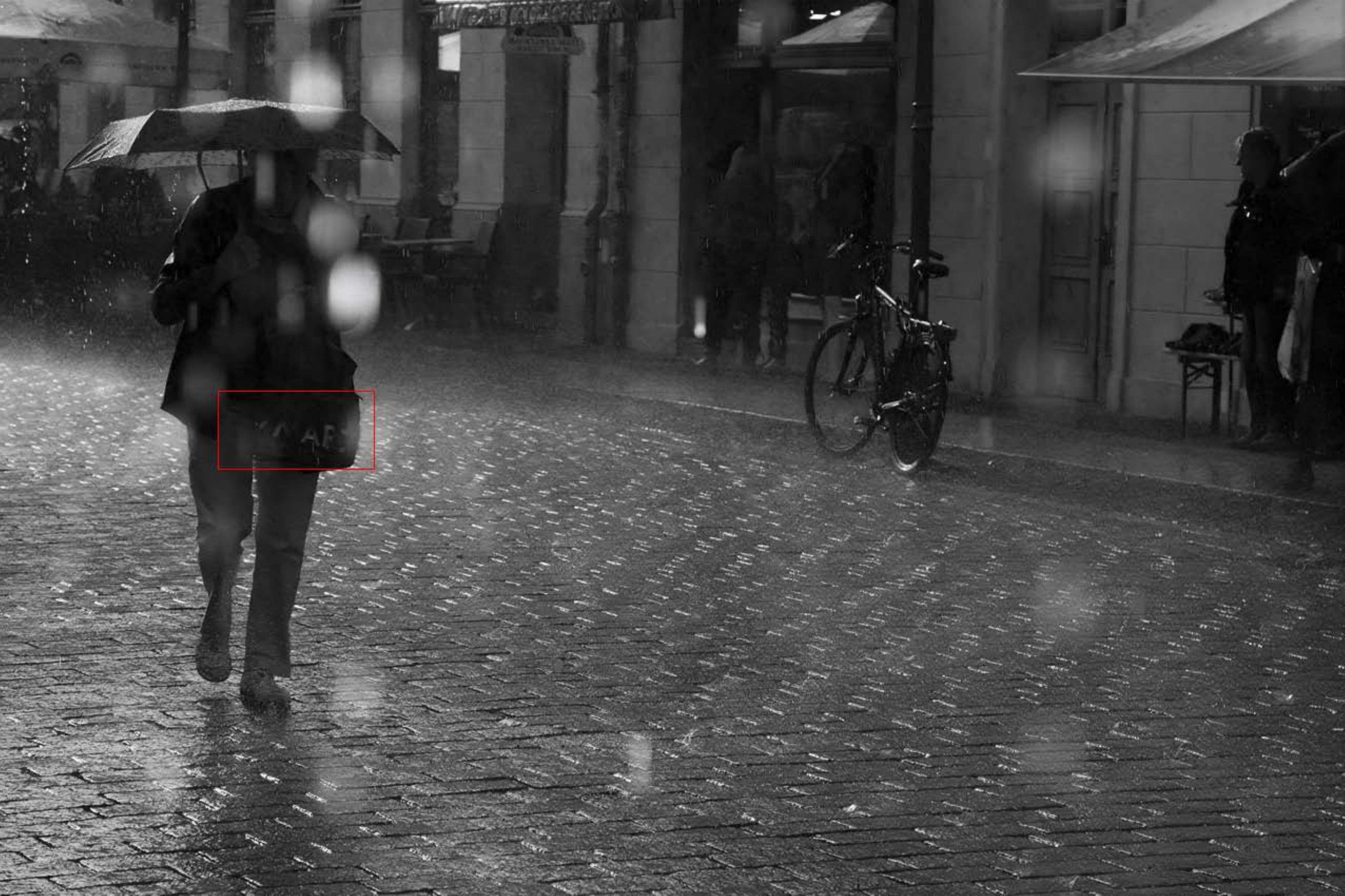} &\hspace{-4mm}
\includegraphics[width = 0.123\linewidth]{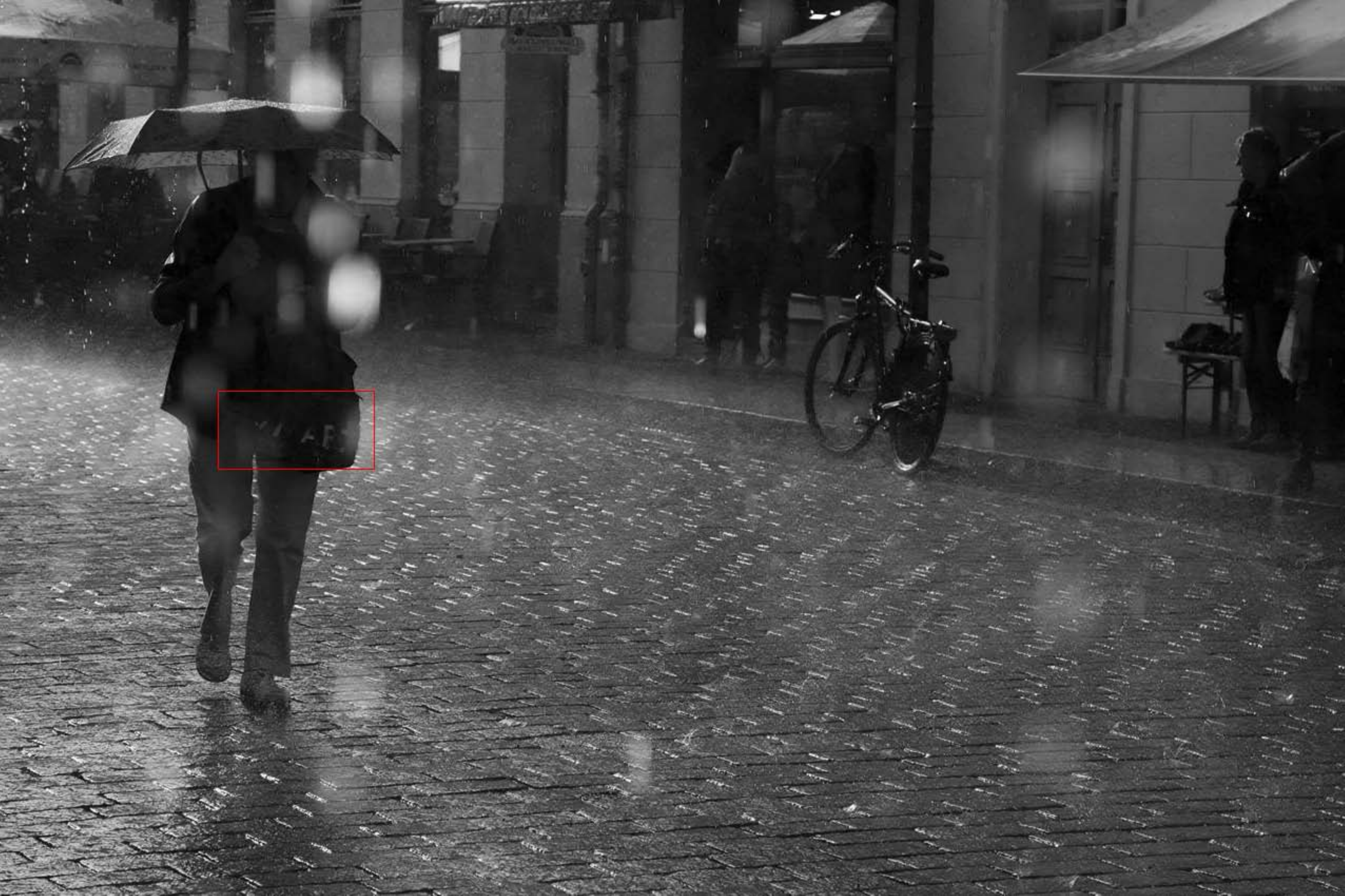} &\hspace{-4mm}
\includegraphics[width = 0.123\linewidth]{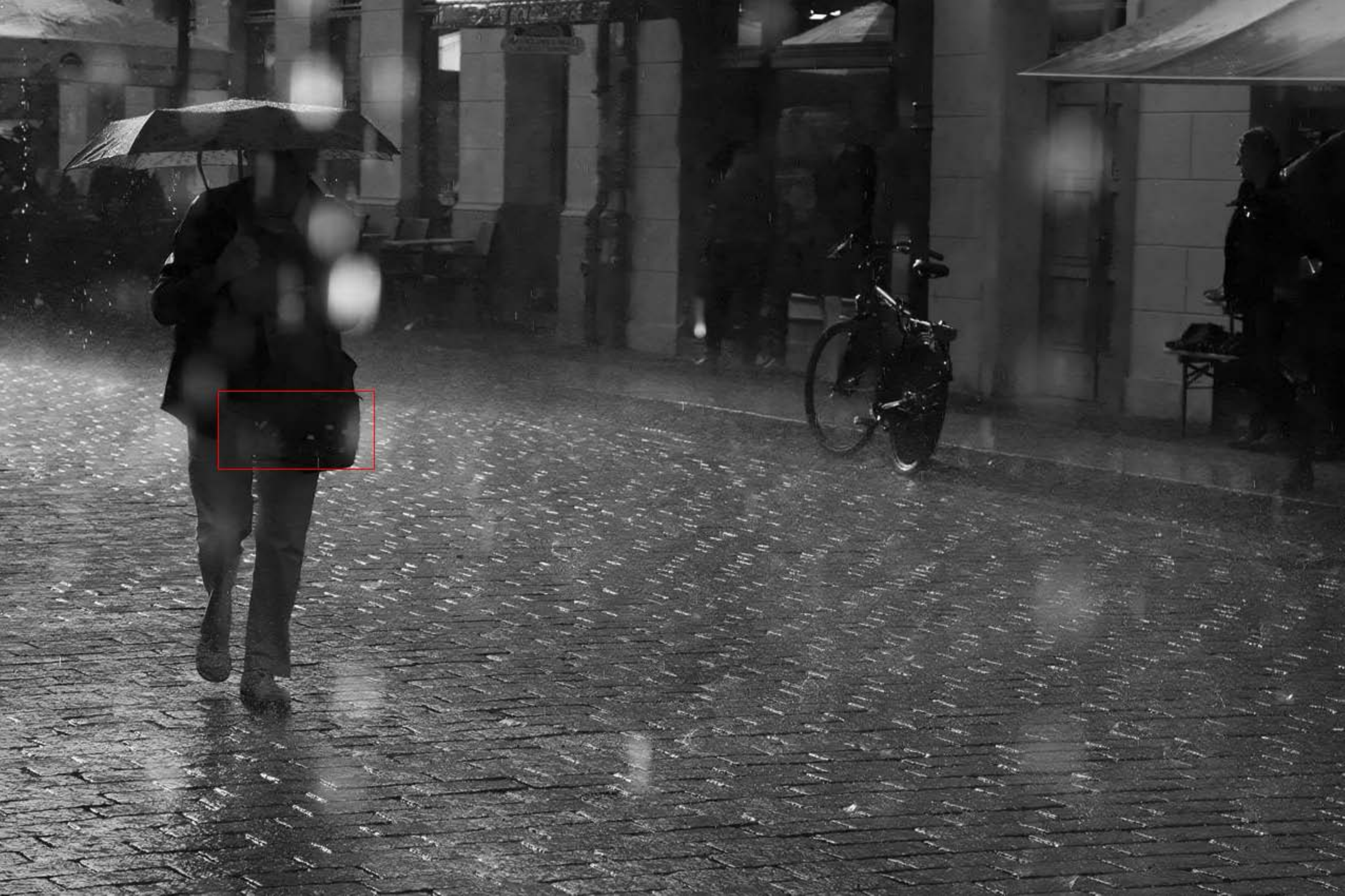} &\hspace{-4mm}
\includegraphics[width = 0.123\linewidth]{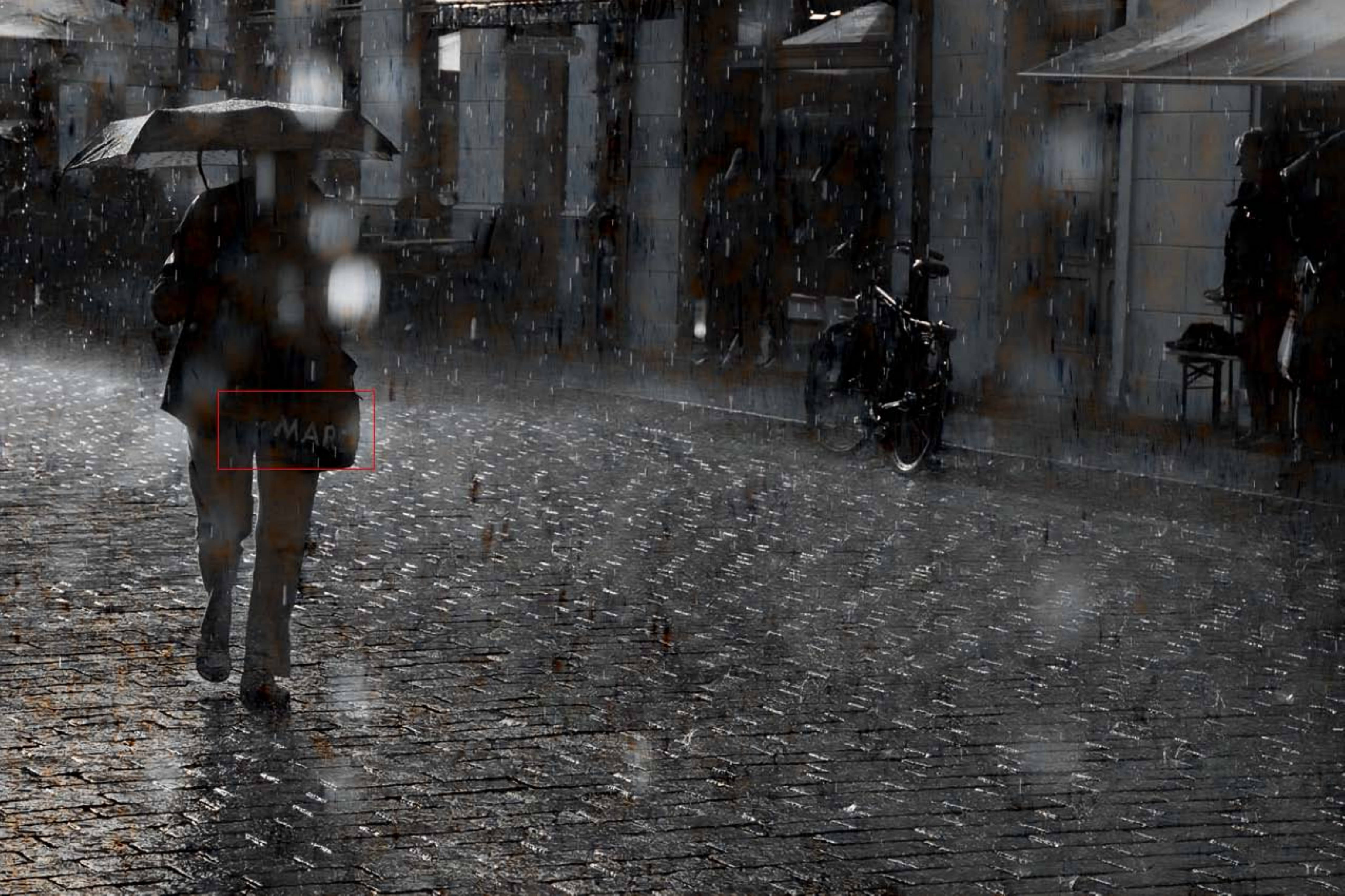} &\hspace{-4mm}
\includegraphics[width = 0.123\linewidth]{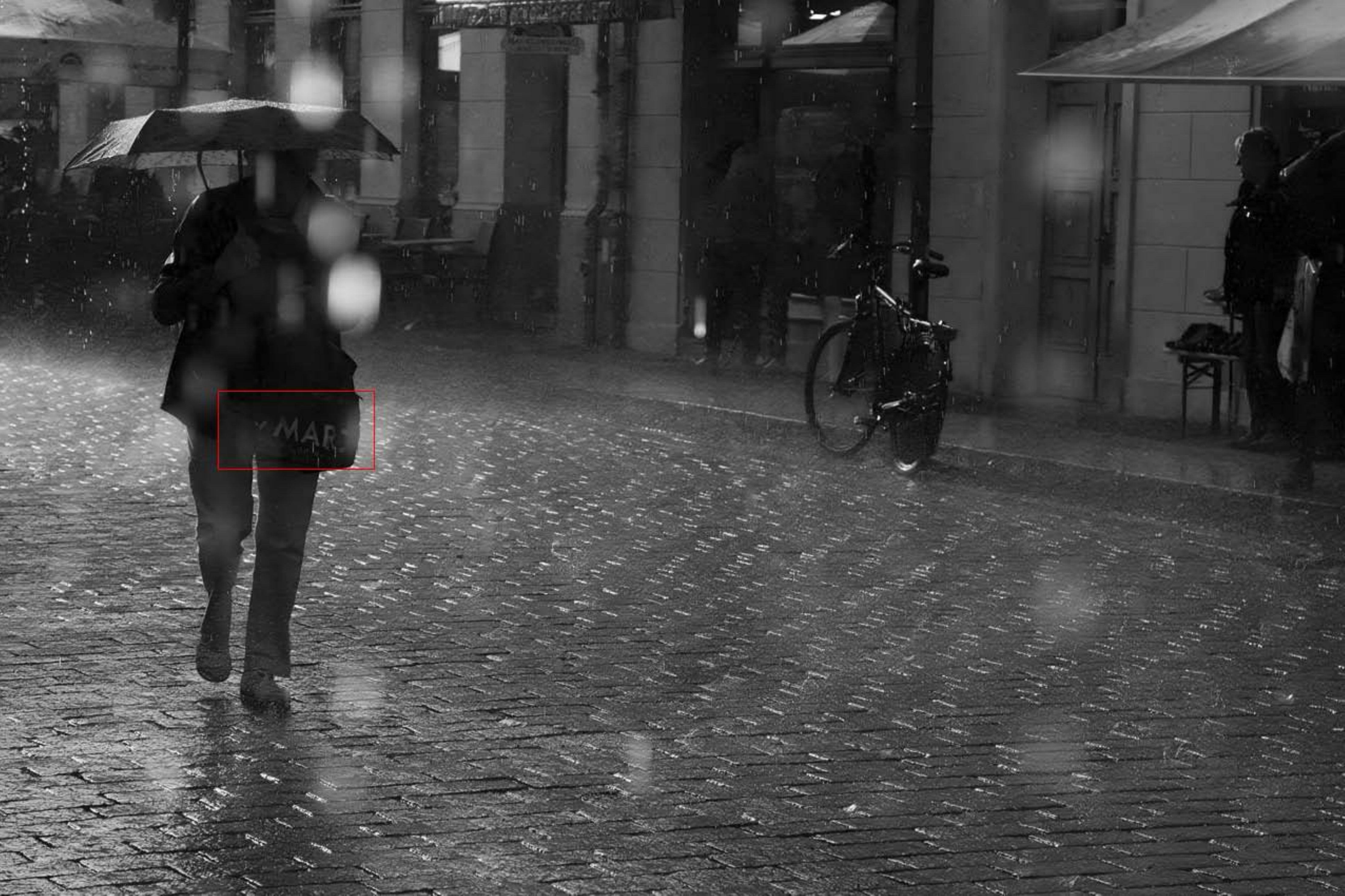}
\\
\includegraphics[width = 0.123\linewidth]{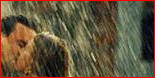} &\hspace{-4mm}
\includegraphics[width = 0.123\linewidth]{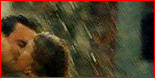} &\hspace{-4mm}
\includegraphics[width = 0.123\linewidth]{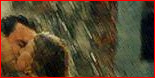} &\hspace{-4mm}
\includegraphics[width = 0.123\linewidth]{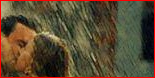} &\hspace{-4mm}
\includegraphics[width = 0.123\linewidth]{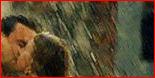} &\hspace{-4mm}
\includegraphics[width = 0.123\linewidth]{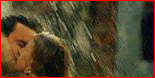} &\hspace{-4mm}
\includegraphics[width = 0.123\linewidth]{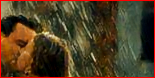} &\hspace{-4mm}
\includegraphics[width = 0.123\linewidth]{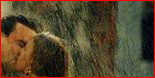}
\\
\includegraphics[width = 0.123\linewidth]{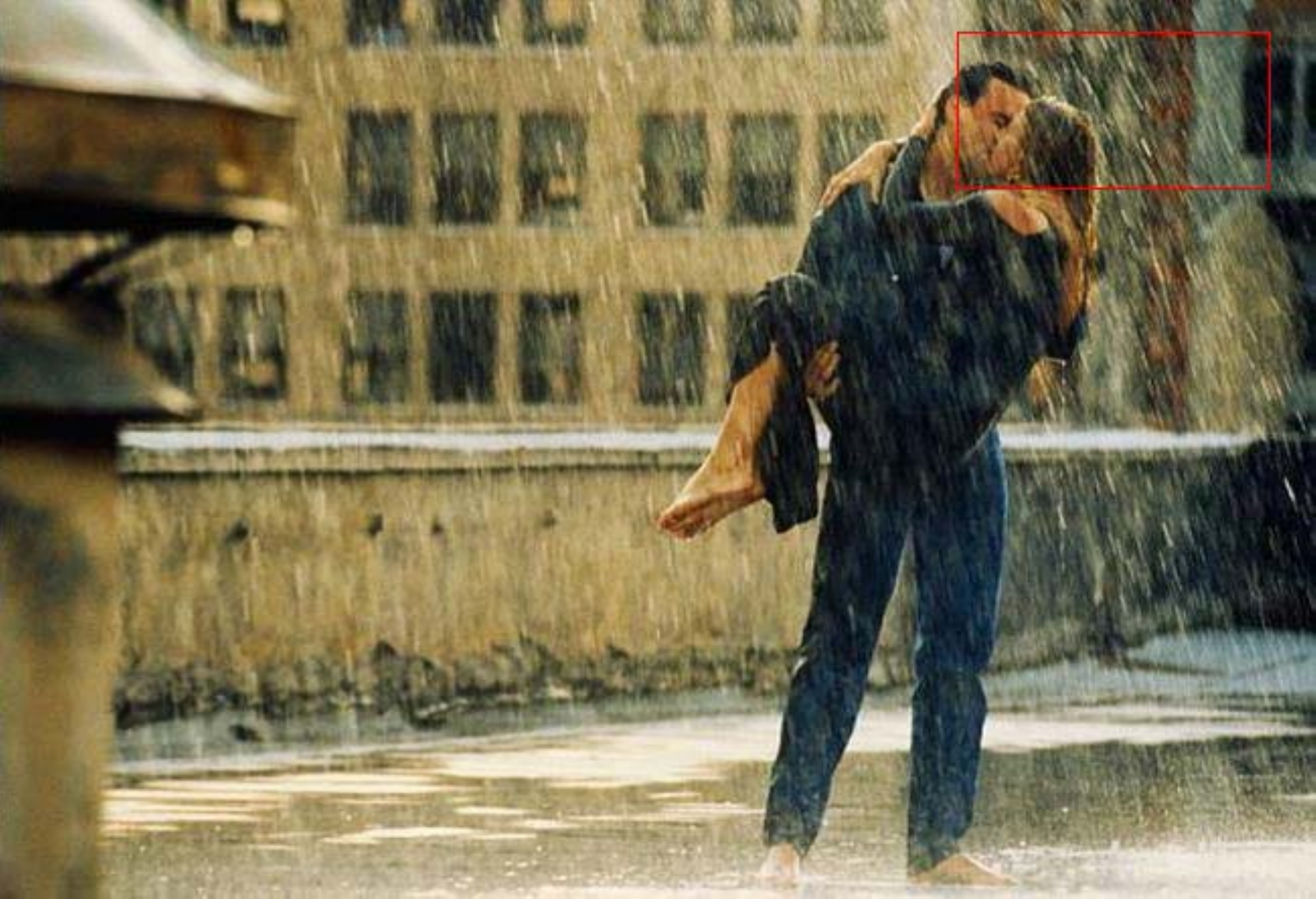} &\hspace{-4mm}
\includegraphics[width = 0.123\linewidth]{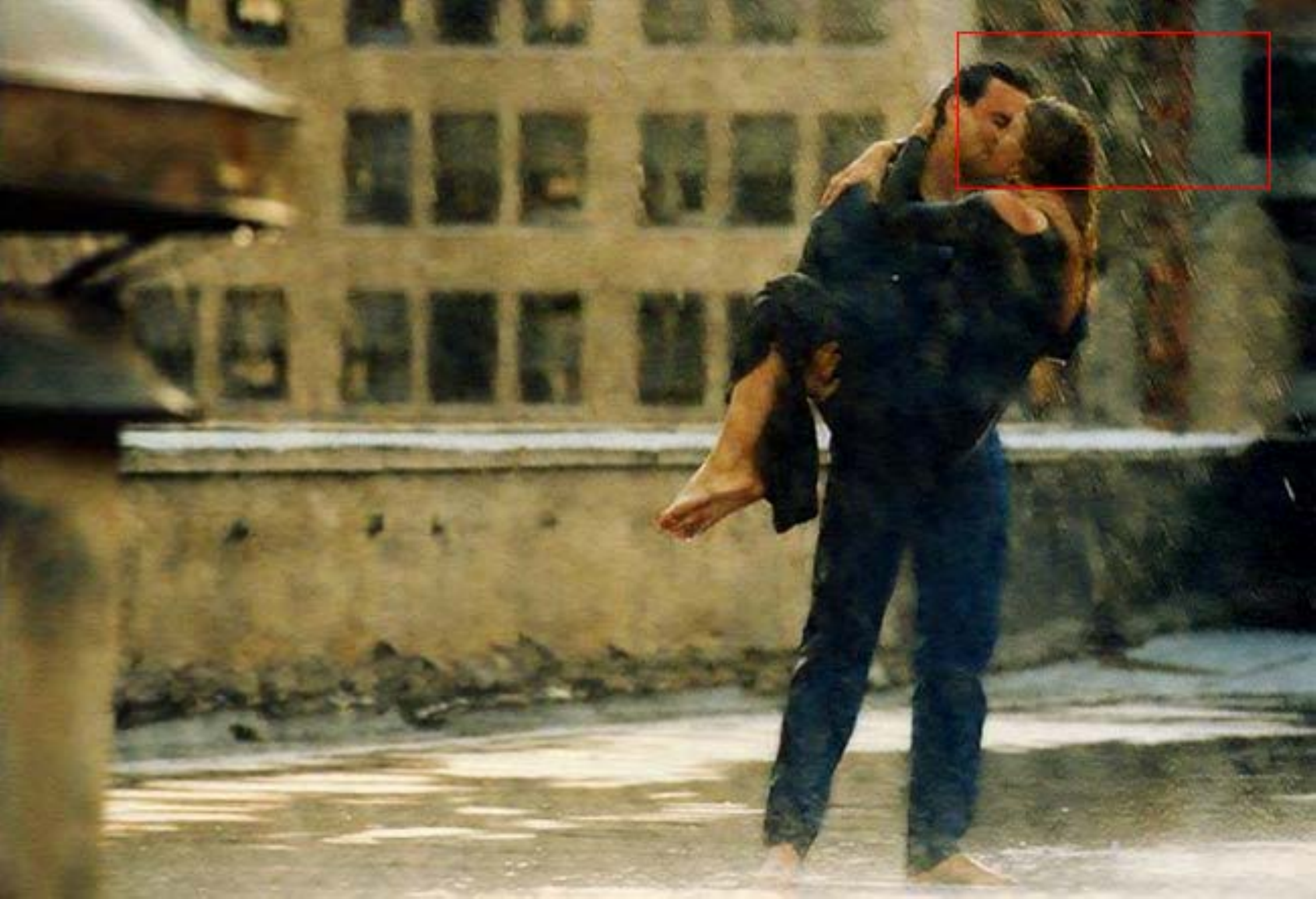} &\hspace{-4mm}
\includegraphics[width = 0.123\linewidth]{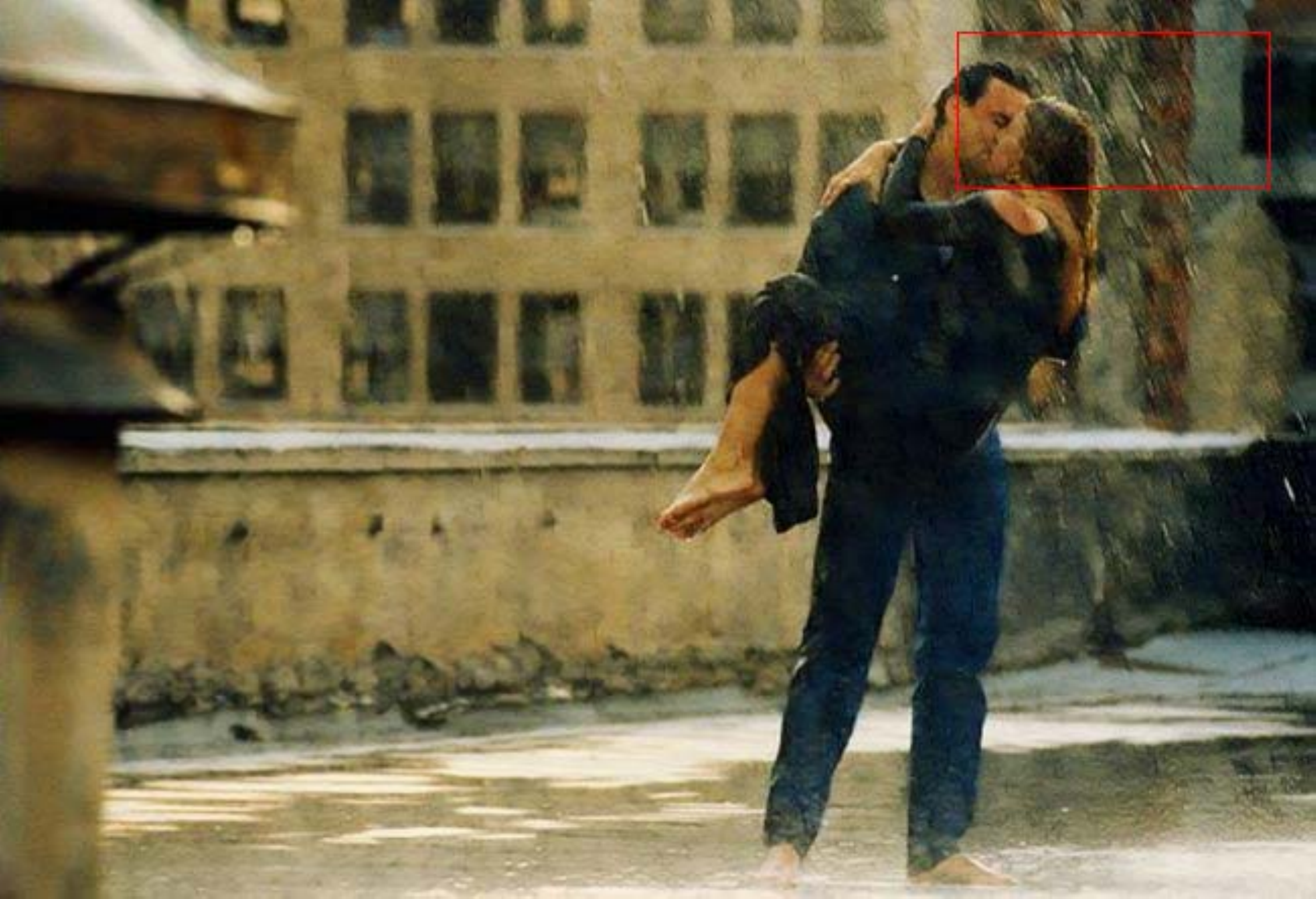} &\hspace{-4mm}
\includegraphics[width = 0.123\linewidth]{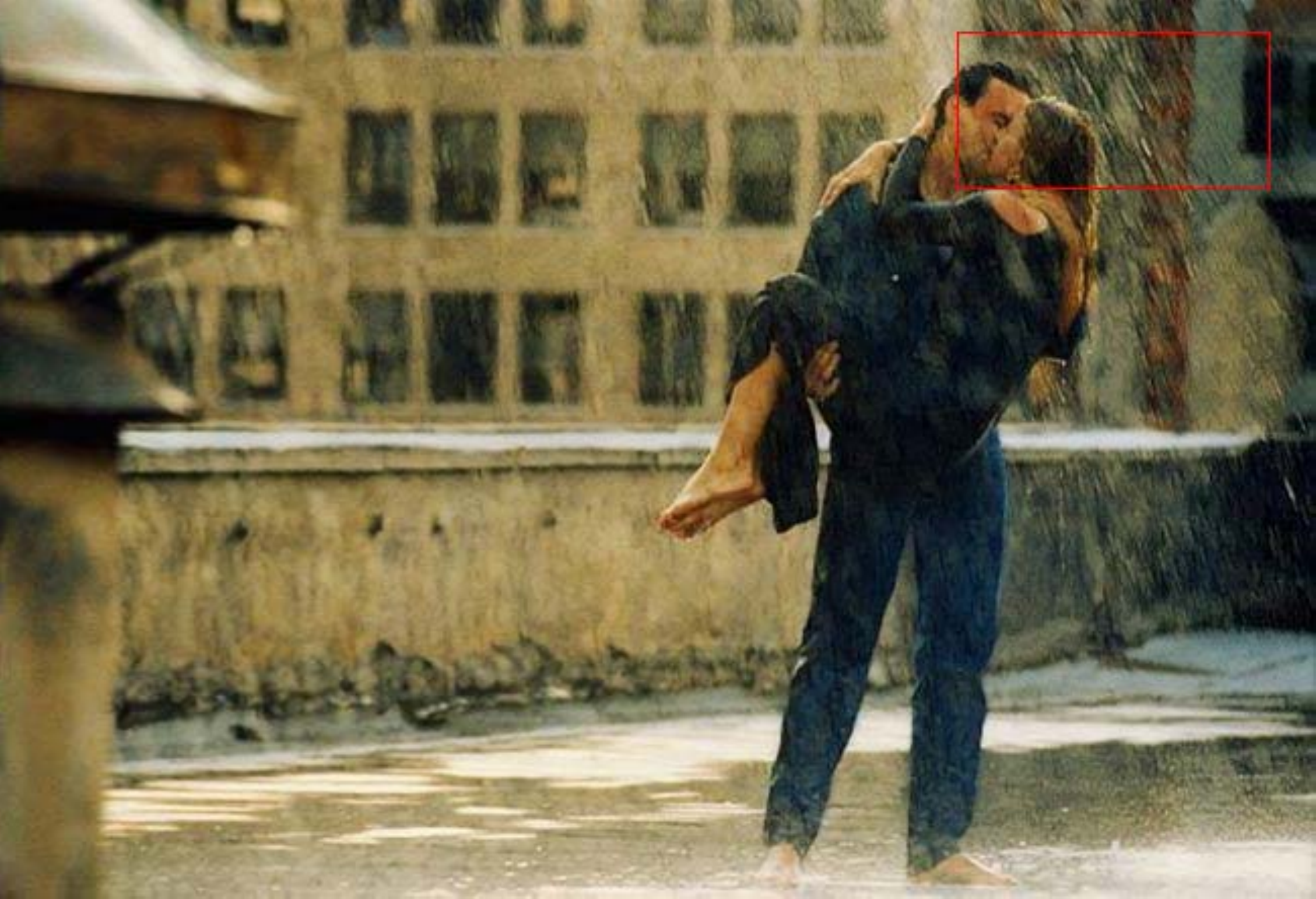} &\hspace{-4mm}
\includegraphics[width = 0.123\linewidth]{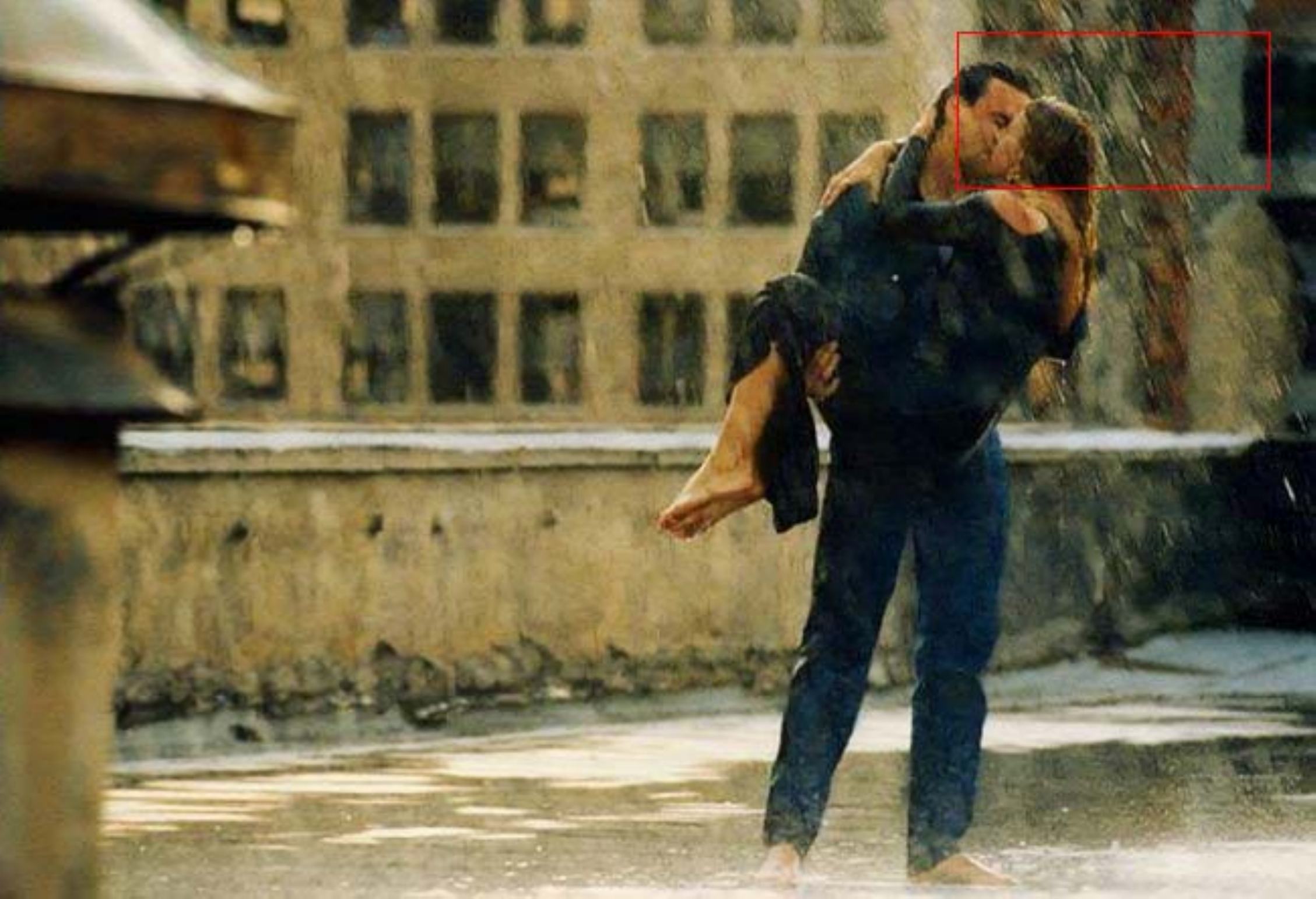} &\hspace{-4mm}
\includegraphics[width = 0.123\linewidth]{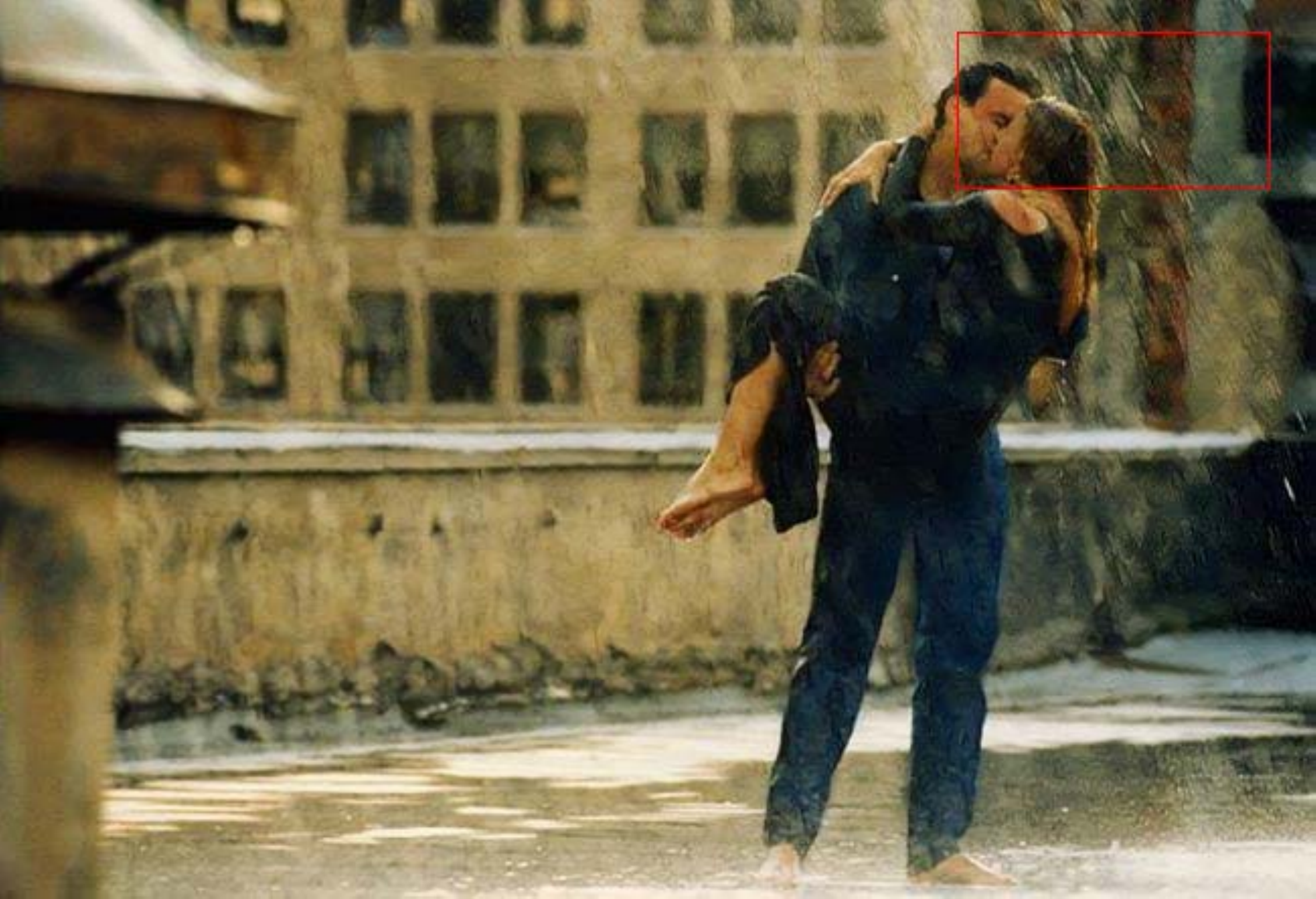} &\hspace{-4mm}
\includegraphics[width = 0.123\linewidth]{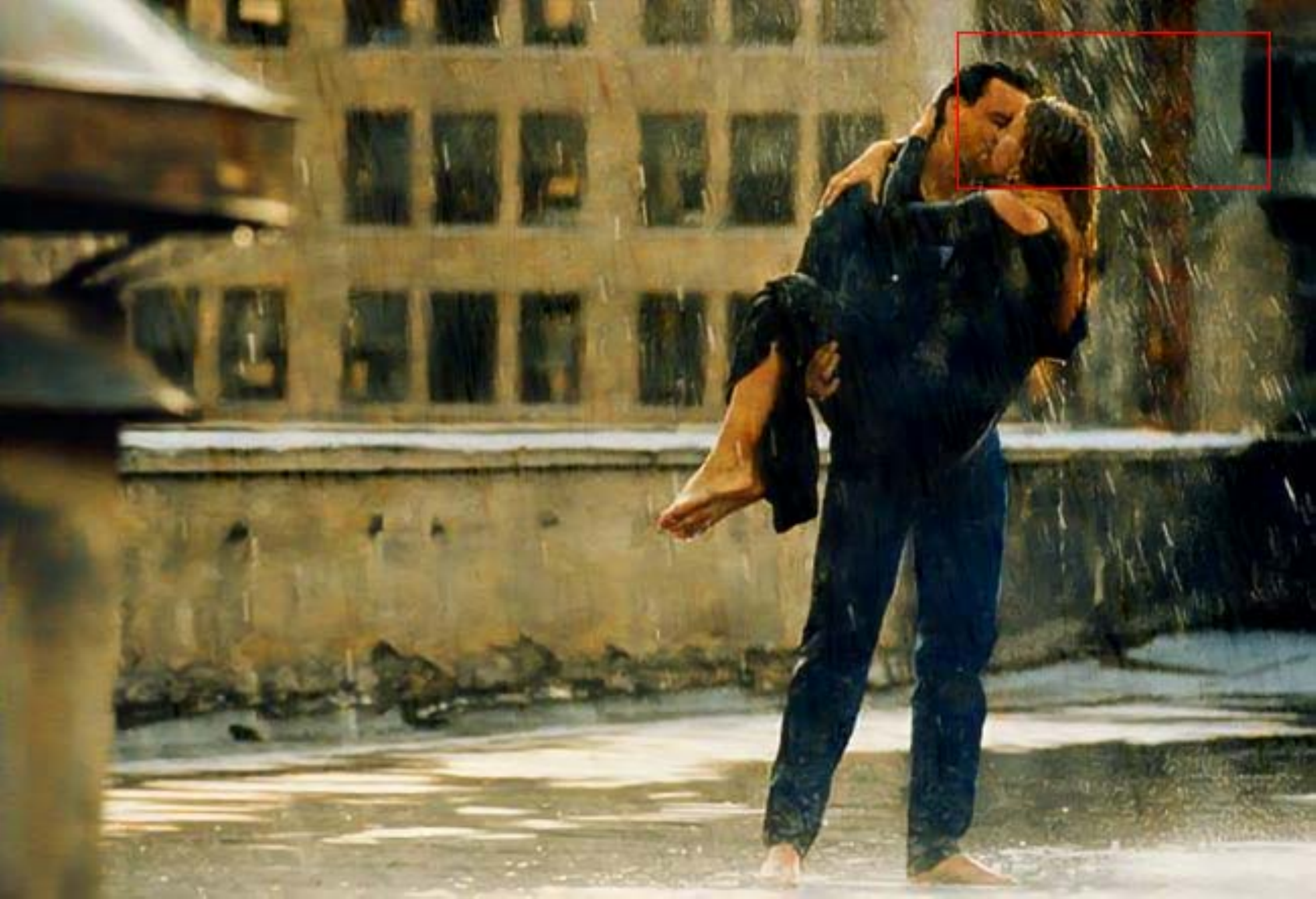} &\hspace{-4mm}
\includegraphics[width = 0.123\linewidth]{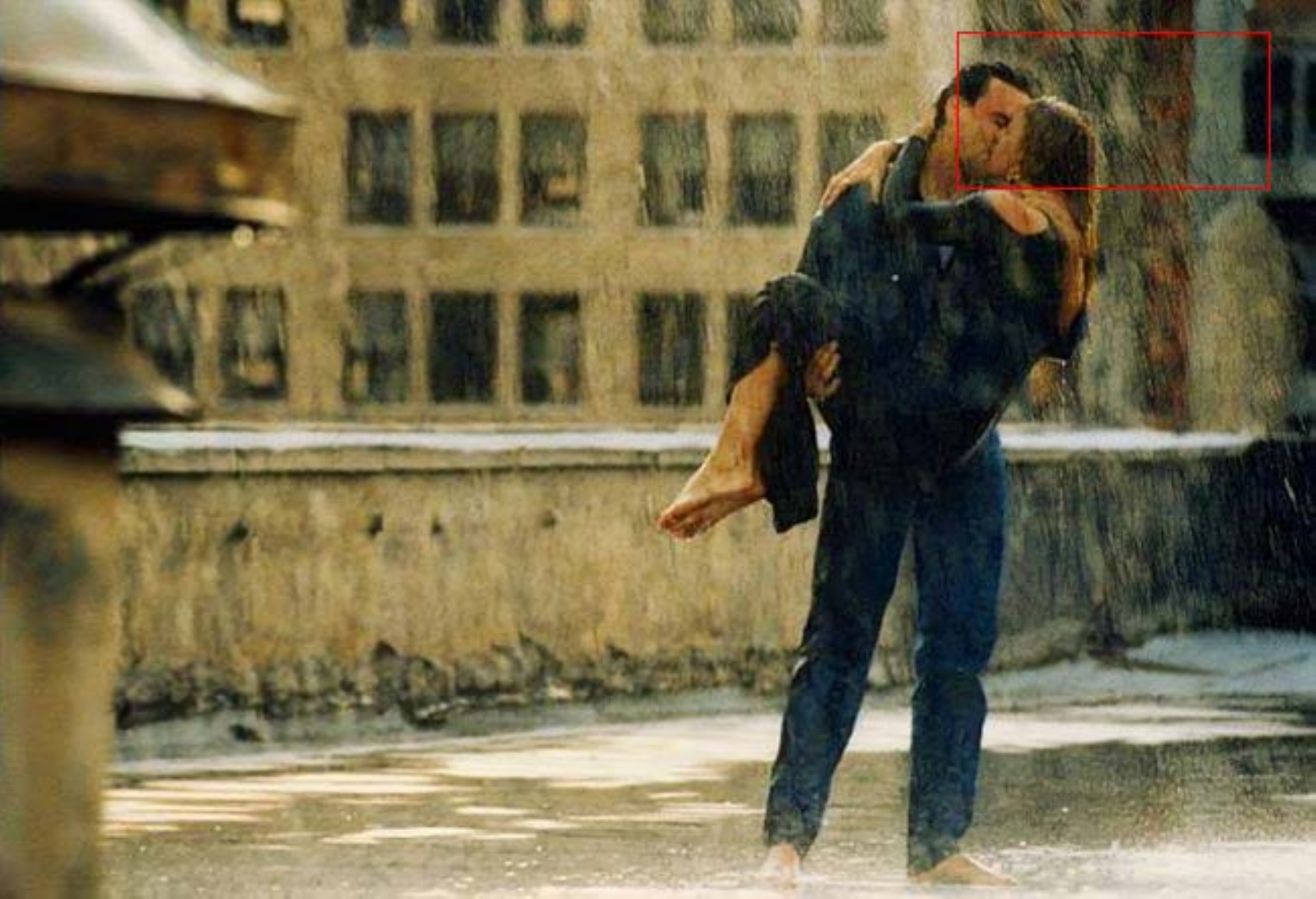}
\\

\includegraphics[width = 0.123\linewidth]{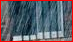} &\hspace{-4mm}
\includegraphics[width = 0.123\linewidth]{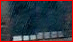} &\hspace{-4mm}
\includegraphics[width = 0.123\linewidth]{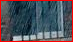} &\hspace{-4mm}
\includegraphics[width = 0.123\linewidth]{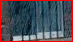} &\hspace{-4mm}
\includegraphics[width = 0.123\linewidth]{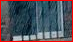} &\hspace{-4mm}
\includegraphics[width = 0.123\linewidth]{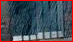} &\hspace{-4mm}
\includegraphics[width = 0.123\linewidth]{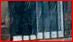} &\hspace{-4mm}
\includegraphics[width = 0.123\linewidth]{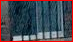}
\\
\includegraphics[width = 0.123\linewidth]{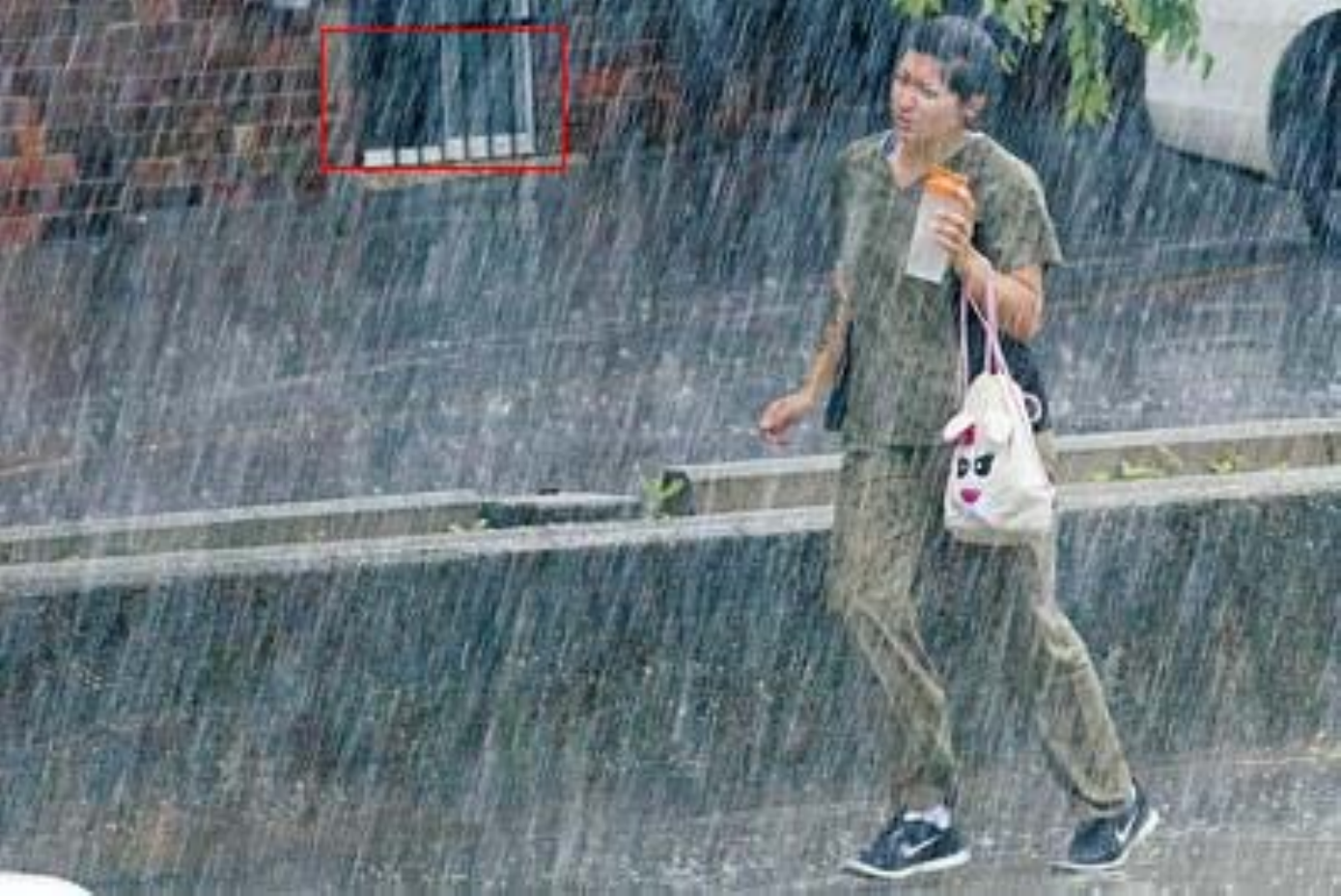} &\hspace{-4mm}
\includegraphics[width = 0.123\linewidth]{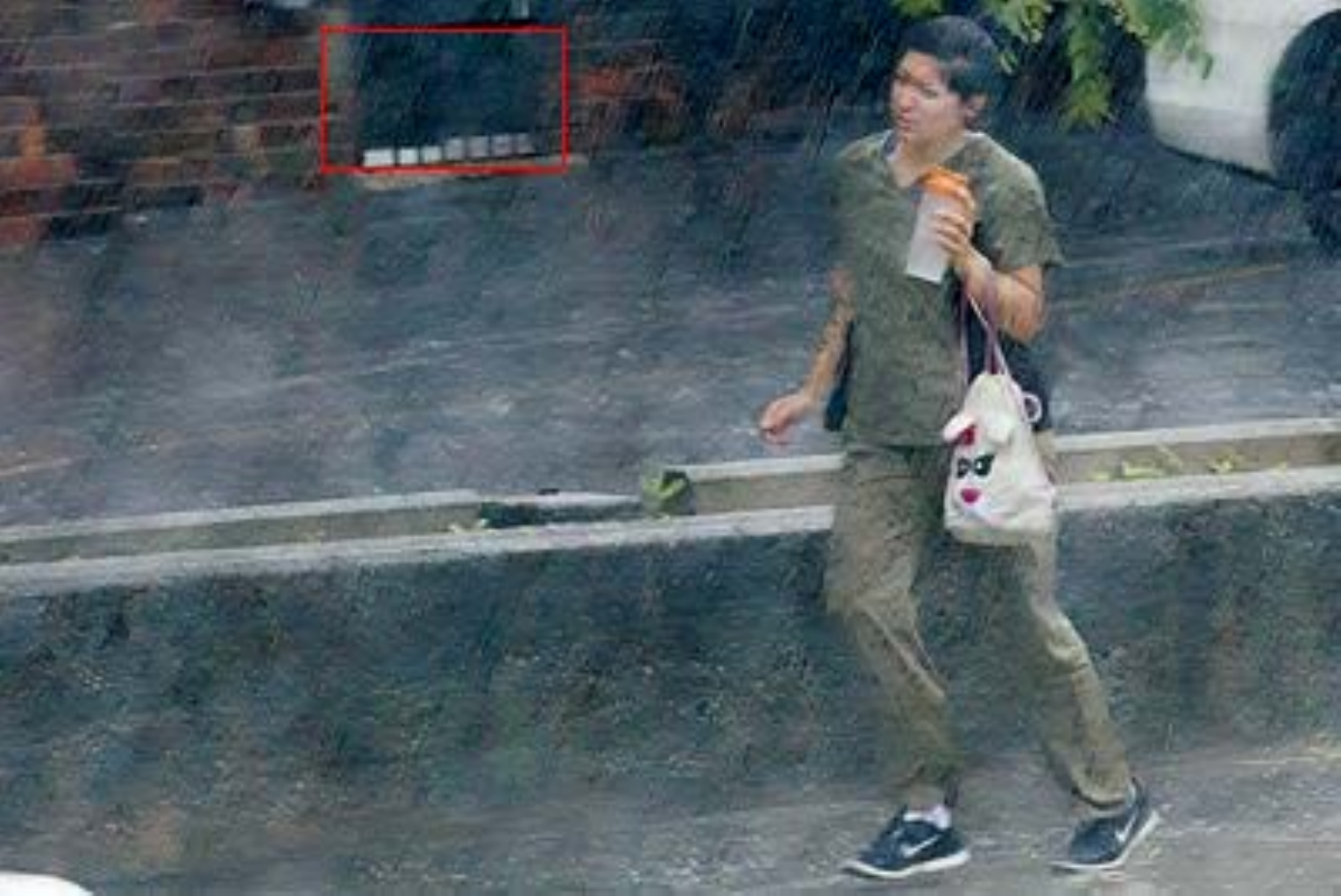} &\hspace{-4mm}
\includegraphics[width = 0.123\linewidth]{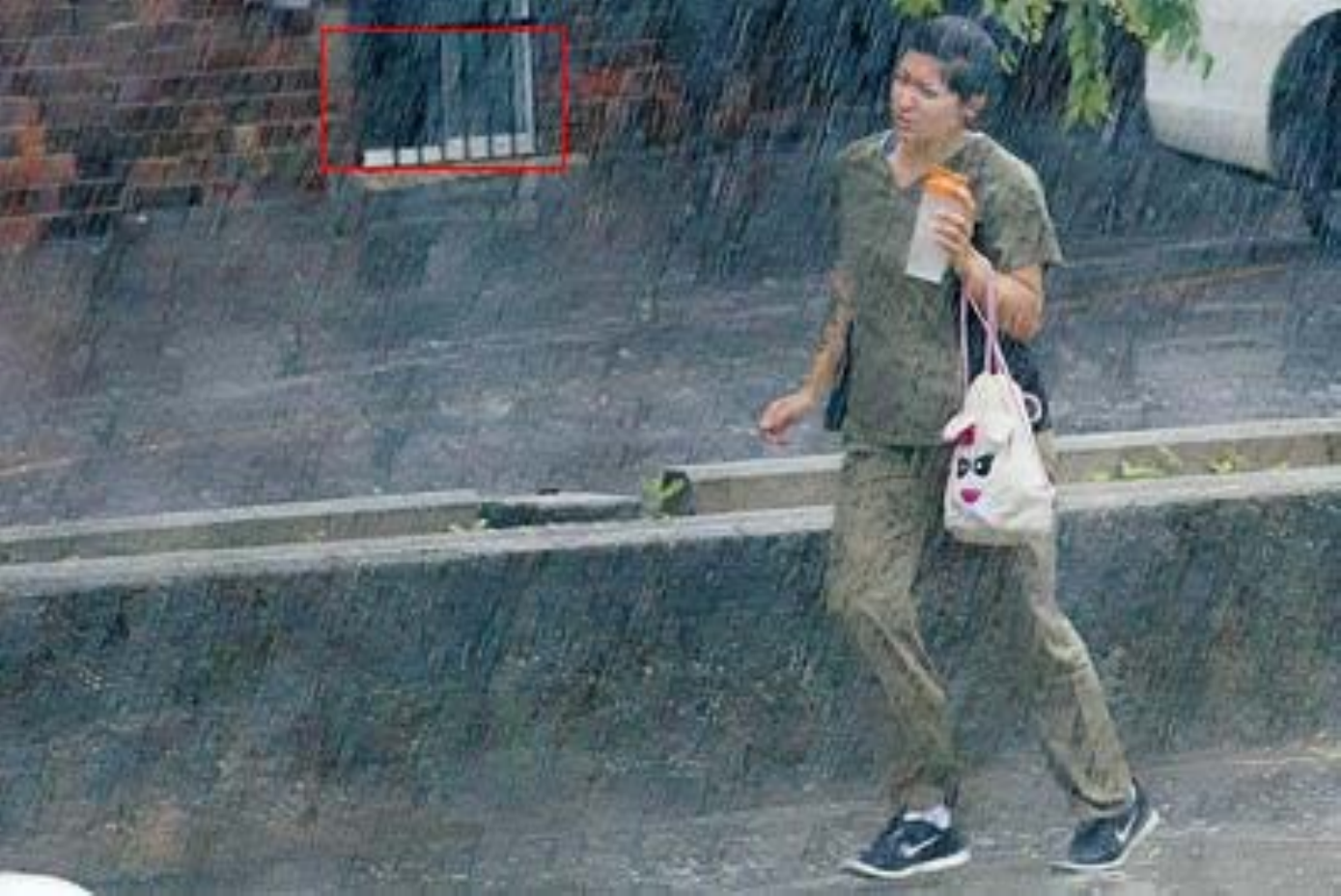} &\hspace{-4mm}
\includegraphics[width = 0.123\linewidth]{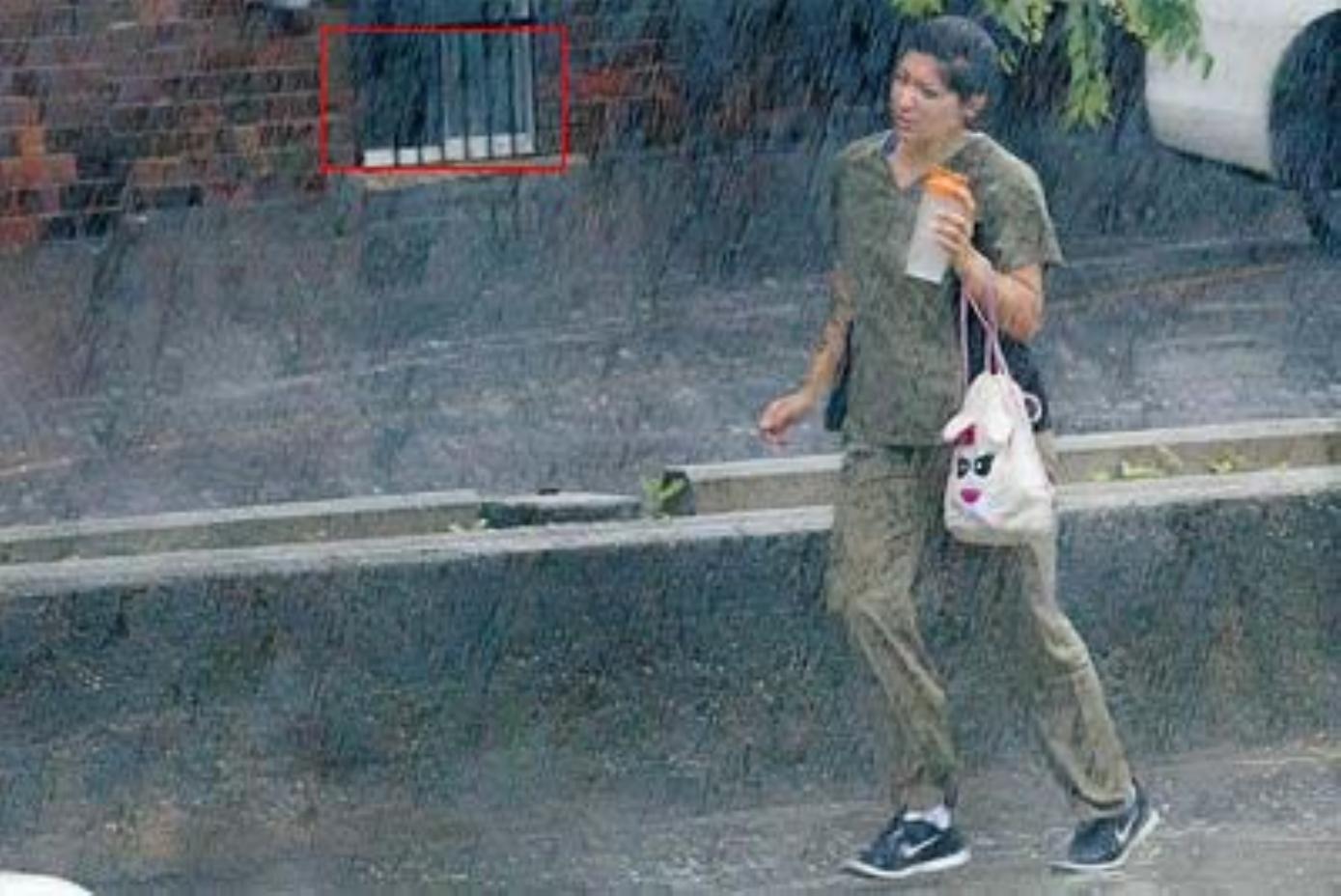} &\hspace{-4mm}
\includegraphics[width = 0.123\linewidth]{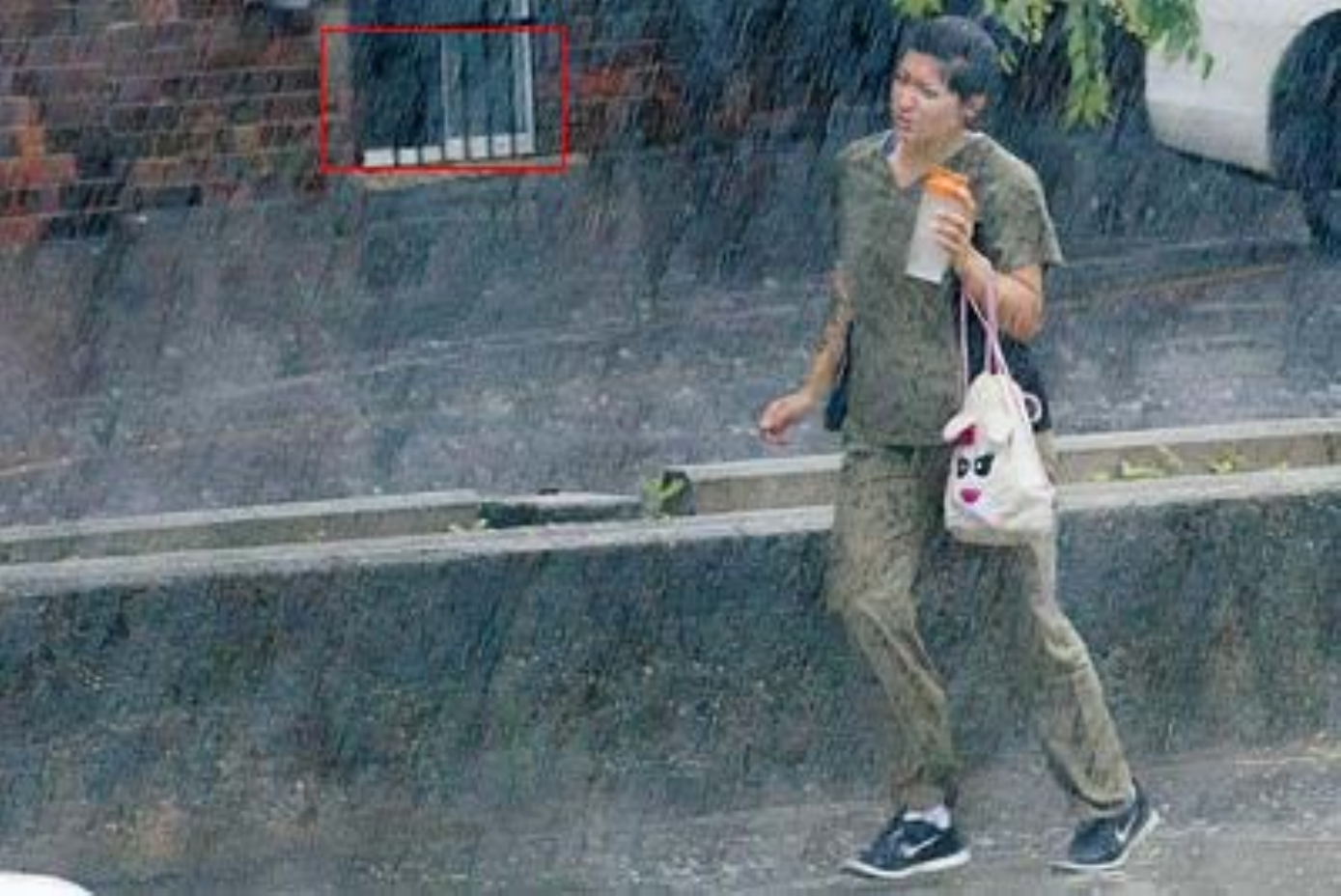} &\hspace{-4mm}
\includegraphics[width = 0.123\linewidth]{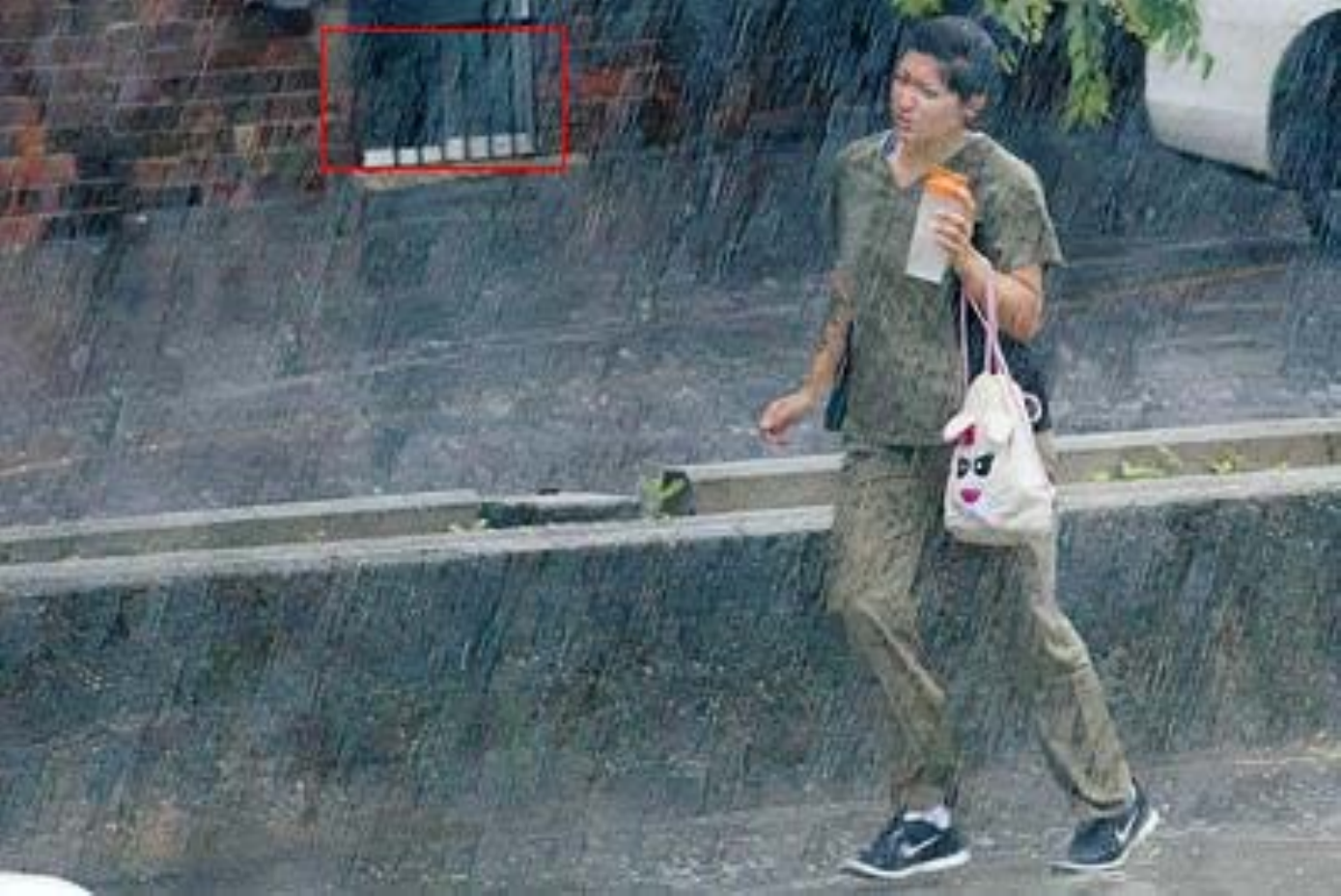} &\hspace{-4mm}
\includegraphics[width = 0.123\linewidth]{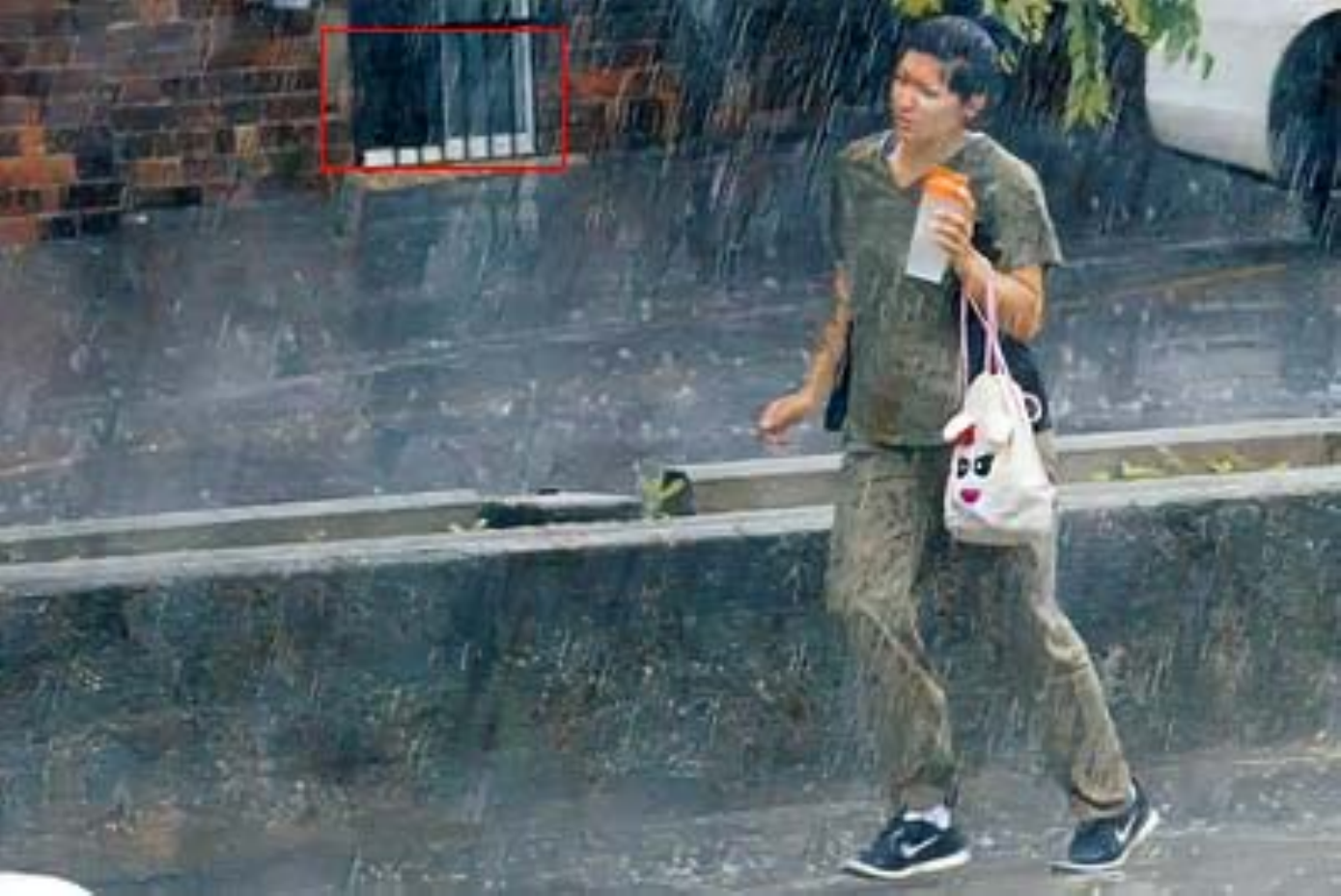} &\hspace{-4mm}
\includegraphics[width = 0.123\linewidth]{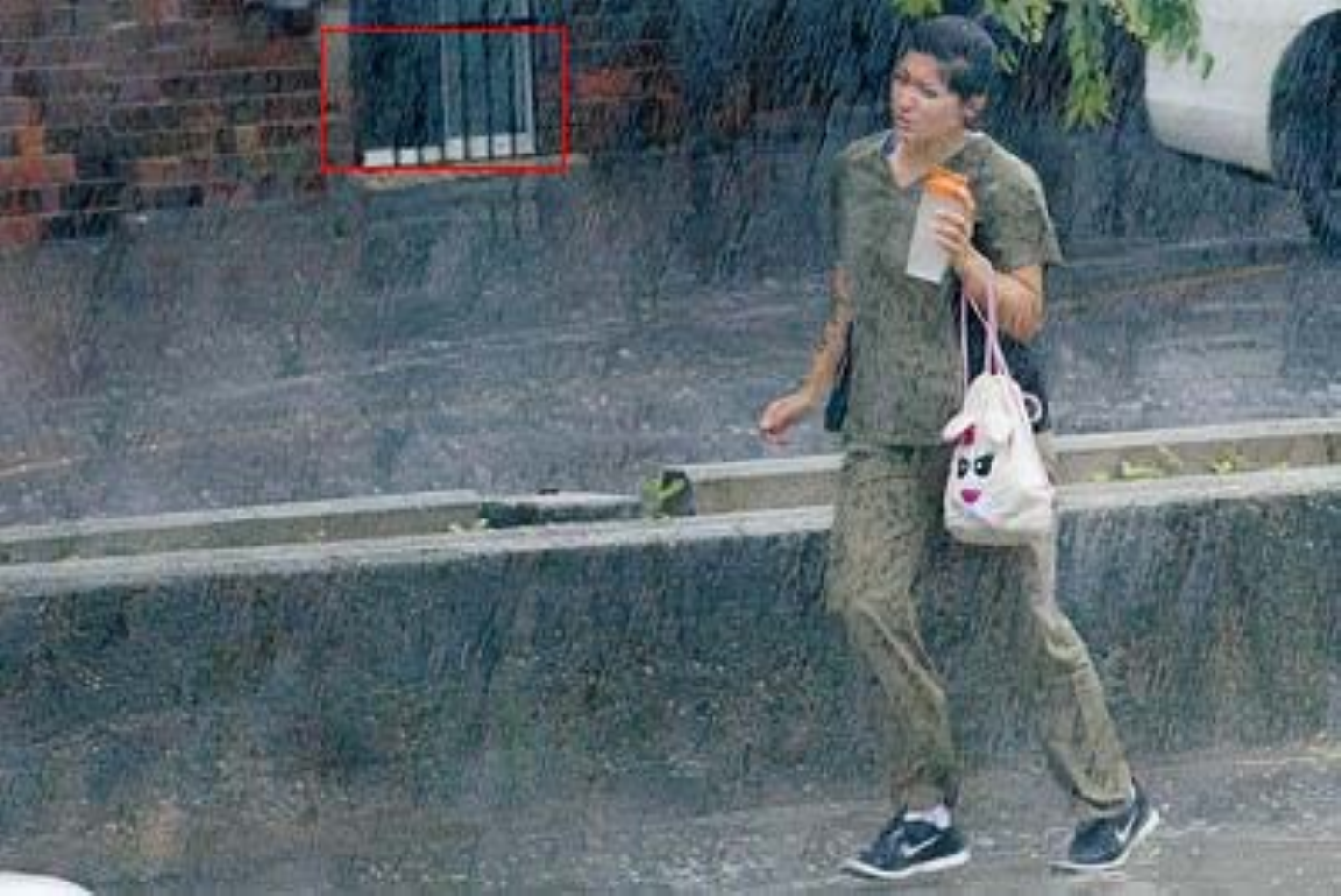}
\\
(a) Input&\hspace{-4mm} (b) DDN&\hspace{-4mm} (c) RESCAN&\hspace{-4mm} (d) NLEDN  &\hspace{-4mm} (e) REHEN&\hspace{-4mm} (f) PreNet&\hspace{-4mm} (g) SSIR&\hspace{-4mm} (h) DCSFN
\\
\end{tabular}
\end{center}
\caption{The results from real-world datasets.
For the first two examples, our method can restore clearer texture information, especially in cropped boxes, while other methods blur the texture.
For the last two examples, other methods hand down a number of rain, while our method is able to restore better rain-free images.
Again, we note that the SSIR~\cite{Derain-cvpr19-semi} is less effective to restore rain-free images.
}
\label{fig:he results from real-world datasets-2}
\end{figure*}

\subsubsection{Synthetic Datasets}
We carry out experiments to verify the effectiveness of our network on synthetic datasets: Rain 100L~\cite{derain_jorder_yang}, Rain 100H~\cite{derain_jorder_yang}, Rain1200~\cite{derain_zhang_did}.
Rain100L has light rain and therefore is relatively the easiest dataset.
The dataset has 1800 image pairs for training and 200 image pairs for testing.
Rain 100H has the same number of training and testing, while has heavy rain with different shapes, directions, and sizes and is also the most challenging dataset.
Rain1200 dataset has 12000 images for training and 1200 images for testing.
The testing set has different background images with training setting, assuring the experiment results are convincing.
\subsubsection{Real-world Datasets}
We collect some real-world rainy images from the Internet and select carefully some rainy images from \cite{derain_Comprehensive_Benchmark_Li_2019_CVPR}.
We use them as our real-world dataset.
\subsubsection{Measurements}
To verify the de-raining performance we use peak signal to noise ratio (PSNR)~\cite{PSNR_thu} and structure similarity index (SSIM)~\cite{SSIM_wang}.
The two measurements are used to compare the restored results with the corresponding ground-truth.
We use the two measurements to evaluate the deraining quality on synthetic datasets.
Since the performance on real-world images can not be evaluated in the same way due to without ground-truth of real-world rainy images, we evaluate the restored image visually.
Moreover, we also use the Natural Image Quality Evaluator (NIQE) and visual comparisons to evaluate the performance of our method. The NIQE calculates the no ground-truth image quality score for the real-world hazy image. The lower the image quality is, the higher NIQE is.
\subsection{Implementation Details}\label{sec:Implementing Details}

We set $L=16$, $K=3$ and $n=4$.
The number of channels is 32.
We randomly select $64 \times 64$ patch pairs from training image datasets as input.
We use $LeakyReLU$ with $\alpha = 0.2$ as our non-linear activation function.
The network is implemented using PyTorch and trained with Adam algorithm~\cite{adam} using two NVIDIA 1080ti GPUs.
We initialize the learning rate to 0.0005, and divide it by 10 at 480 epochs and 640 epochs, and stop the training at 800 epochs.

\begin{table}[!h]
	\centering
	\caption{The NIQE results. The best result is marked in bold.}
	\scalebox{0.99}{
		\begin{tabular}{lc}
			\toprule
			Methods & NIQE  \\
			\midrule
			DDN~\cite{derain_ddn_fu}  & 4.14     \\
			RESCAN~\cite{derain_rescan_li}& 3.98\\
			NLEDN~\cite{derain_nledn_li}& 3.62\\
			REHEN~\cite{derain-acmmm19-rehen}& 3.98\\
			SSIR~\cite{Derain-cvpr19-semi}&3.86\\
			PreNet~\cite{derain_prenet_Ren_2019_CVPR}& 4.13\\
			DCSFN (Ours)   & \textbf{3.54}\\
			\bottomrule
	\end{tabular}}
	\label{tab:The NIQE results.}
\end{table}

\begin{table*}[!t]
	\caption{Ablation study on different components.
		The $\checkmark$ symbol denotes that the corresponding component is adopted.}
	\label{tab:ablation study_Table1}
	\scalebox{0.99}{
		\centering
		\begin{tabular}{lcccc}
			\toprule
			Experiments                 &$M_{1}$        & $M_{2}$       & $M_{3}$     &$M_{4}$             \\
			\midrule
			w/o Skip Connection         & $\checkmark$  &              &             &                                    \\
			Skip Connection             &              &  $\checkmark$ & $\checkmark$& $\checkmark$           \\
			w/o Dense Connection        &               & $\checkmark$ &              &                             \\
			Dense Connection            &$\checkmark$   &              &  $\checkmark$ & $\checkmark$                \\
			Single                      &               &              &  $\checkmark$ &                                             \\
			Cross-scale                 & $\checkmark$  &$\checkmark$  &              & $\checkmark$                   \\
			\midrule
			PSNR                        & 27.86     & 27.78   & 26.89& 28.81        \\
			\midrule
			SSIM                         & 0.8979    & 0.8898  & 0.8770& 0.9040        \\
			\bottomrule
		\end{tabular}
	}
\end{table*}

\subsection{Results on Synthetic Datasets}\label{sec:Results on Synthetic Datasets}
We evaluate the proposed method on three widely used datasets: Rain100H, Rain100L, and Rain1200 in Tab.~\ref{tab: the results in synthetic datasets}.
Compared with other state-of-the-art methods, we achieve the newest state-of-the-art results on these datasets.
Especially, SSIR~\cite{Derain-cvpr19-semi}, a semi-supervised deraining network, is less effective on synthetic datasets.

We show some deraining examples on synthetic datasets in Fig.~\ref{fig:deraining-syn-example}.
We can observe that our method can restore better results from the global perspective and obtain clearer texture and cleaner background, especially in the cropped boxes.
We note that the SSIR fails to restore rain-free images, which illustrates their designed semi-supervised method cannot work on synthetic datasets.

\subsection{Results on Real-world Datasets}\label{sec:Results on Real-world Datasets}
We verify the robustness on real-world datasets in Fig.~\ref{fig:he results from real-world datasets-2}.
For the first two examples, our method can restore clearer texture information, especially in cropped boxes, while other methods blur the texture.
For the last two examples, other methods hand down a number of rain, while our method is able to restore better rain-free images.
Especially, we note that the SSIR~\cite{Derain-cvpr19-semi} can not restore rain-free and always hand down lots of rain.
The above examples illustrate our proposed method is a better deraining algorithm that can restore cleaner and clearer rain-free images and also can preserve better details and texture information.

We computer the NIQE values on our collected real-world datasets and the results are shown in Tab.~\ref{tab:The NIQE results.}.
We observe that our method has lower NIQE value, which illustrates our method is able to generate more natural deraining results.
This also demonstrates that the proposed method is a more robust derainer than others on real-world rainy conditions.

\begin{table}[!h]
	\centering
	\caption{Discussion on Inner-scale Connection.}
	\scalebox{0.85}{
		\begin{tabular}{ccccc}
			\toprule
			Metric &w/o inner-scale Connection& w  inner-scale Connection      \\
			\midrule
			PSNR  & 28.32& 28.81    \\
			\midrule
			SSIM  &0.9031 &  0.9040       \\
			\bottomrule
	\end{tabular}}
	\label{tab:Discussion on Inner-scale Connection.}
\end{table}

\subsection{Ablation Study}\label{sec:Ablation Study}
We analyze the proposed model by conducting various experiments.
\subsubsection{Analysis on Network Designment}
We analyze the network designment that consists of skip connection, dense connection, single model, and cross-scale network.
The results are illustrated in Tab.~\ref{tab:ablation study_Table1} and the abbreviation of these experiments are as follows:
\begin{itemize}
\item $M_{1}$: The proposed model without skip connection.
\item $M_{2}$: The proposed model without dense connection.
\item $M_{3}$: Single model that only consists of a single network.
\item $M_{4}$: Our proposed network with skip connection, dense connection, inner connection, and three sub-networks.
\end{itemize}
We can observe that both skip connection and dense connection have improvements for the deraining results.
Compared to the single model that is a single deraining network with a single scale network, the proposed cross-scale manner has significant improvement for the deraining performance.
This also shows our proposed cross-scale method is a better designment that a remarkable promotion to the deraining results.
\subsubsection{Discussion on Inner-scale Connection}
In this paper, we propose an inner-scale connection manner that improves the correlation between different scales.
Whether the inner-connection is meaningful for the deraining result is worth discussing.
We show the results in Tab.~\ref{tab:Discussion on Inner-scale Connection.} and we can see that the designed inner-scale connection can improve the deraining performance.
This demonstrates that our proposed inner-scale connection manner is meaningful.

\begin{table}[!h]
	\centering
	\caption{The analysis of the number of scale ($K$).}
	\scalebox{0.99}{
		\begin{tabular}{ccccc}
			\toprule
			Metric & K = 1& K = 2 & K = 3 & K = 4      \\
			\midrule
			PSNR  & 28.34& 28.35 & 28.81& 28.59    \\
			\midrule
			SSIM  &0.9006 &  0.9007 &0.9040 &  0.9036      \\
			\bottomrule
	\end{tabular}}
	\label{tab:The analysis on the number of scale.}
\end{table}

\begin{figure}[!h]
	\begin{center}
		\begin{tabular}{cc}
			\includegraphics[width = 0.5\linewidth]{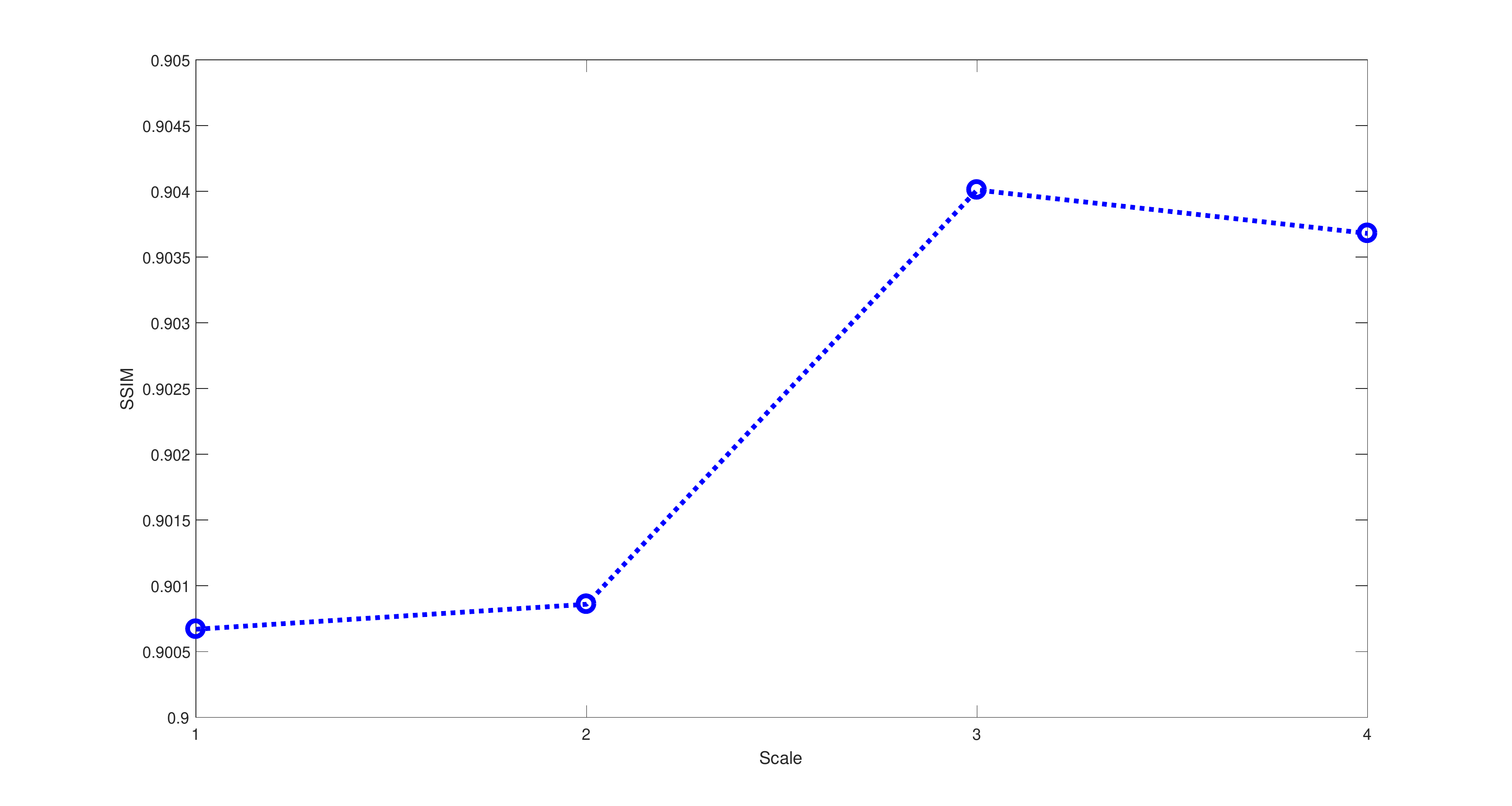} &\hspace{-4.5mm}
			\includegraphics[width = 0.5\linewidth]{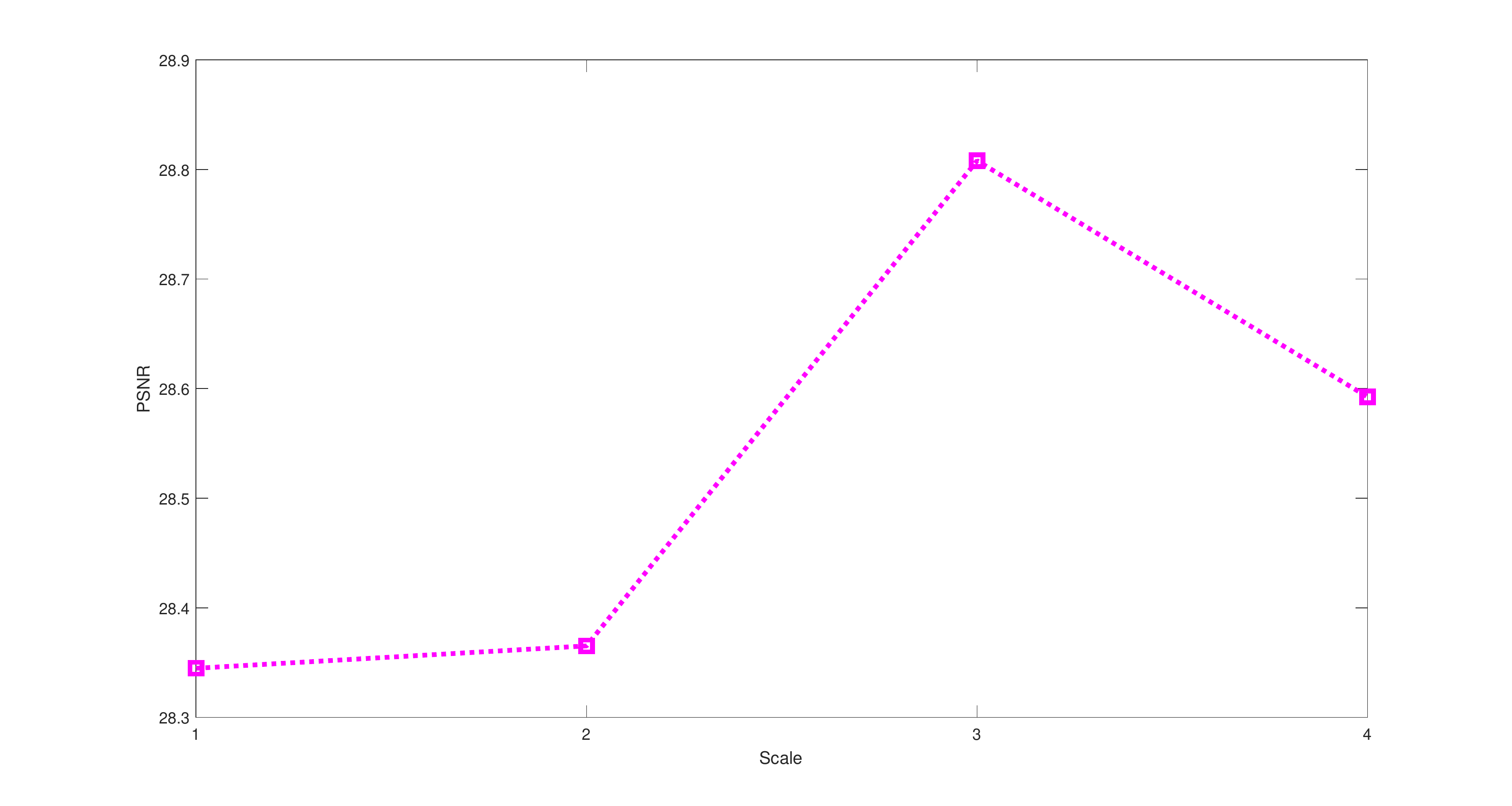}
			\\
			(a) PSNR  &\hspace{-4mm} (b) SSIM
			\\
		\end{tabular}
	\end{center}
	\caption{Curve graph about the number of scale.
	}
	\label{fig: Tcurve graph about the number of scale}
\end{figure}

\subsubsection{The analysis of the number of scales}
We provide an analysis of the number of scales in the inner-scale connection block.
The results are given in Tab.~\ref{tab:The analysis on the number of scale.}
Compare the single scale, the multi-scale manner improves deraining results and it reaches the best performance in terms of PSNR and SSIM when $K=3$.
So we select $K = 3$ as the number of the scale.
To better see the effect of the scale number on the results, we provide the curve graph about the number of scales in Fig.~\ref{fig: Tcurve graph about the number of scale}.
We can see that it has a significant improvement when $K = 3$.
\subsubsection{Analysis of recurrent unit}
\begin{figure}[!h]
\begin{center}
\begin{tabular}{cc}
\includegraphics[width = 0.5\linewidth]{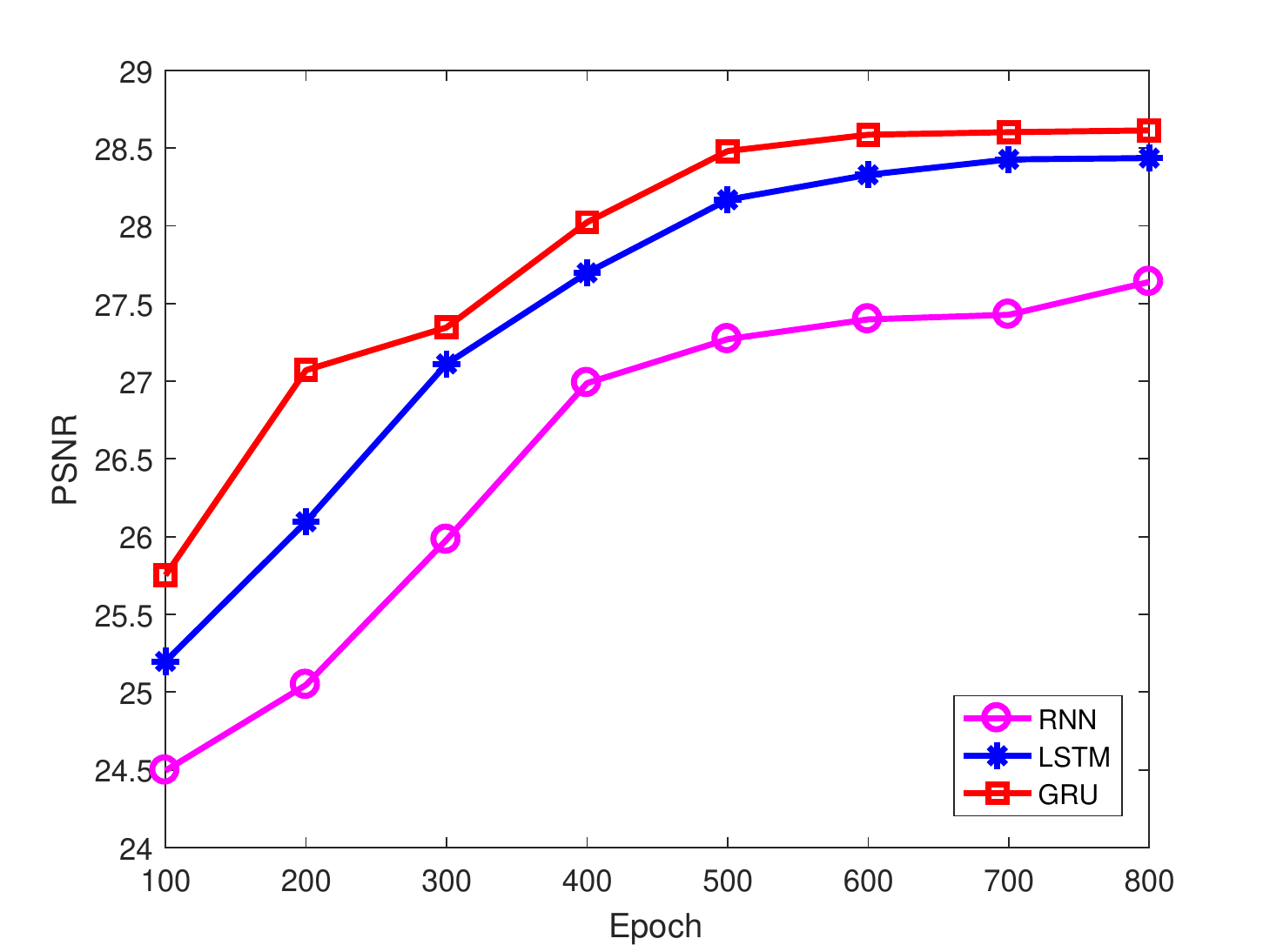} &\hspace{-4.5mm}
\includegraphics[width = 0.5\linewidth]{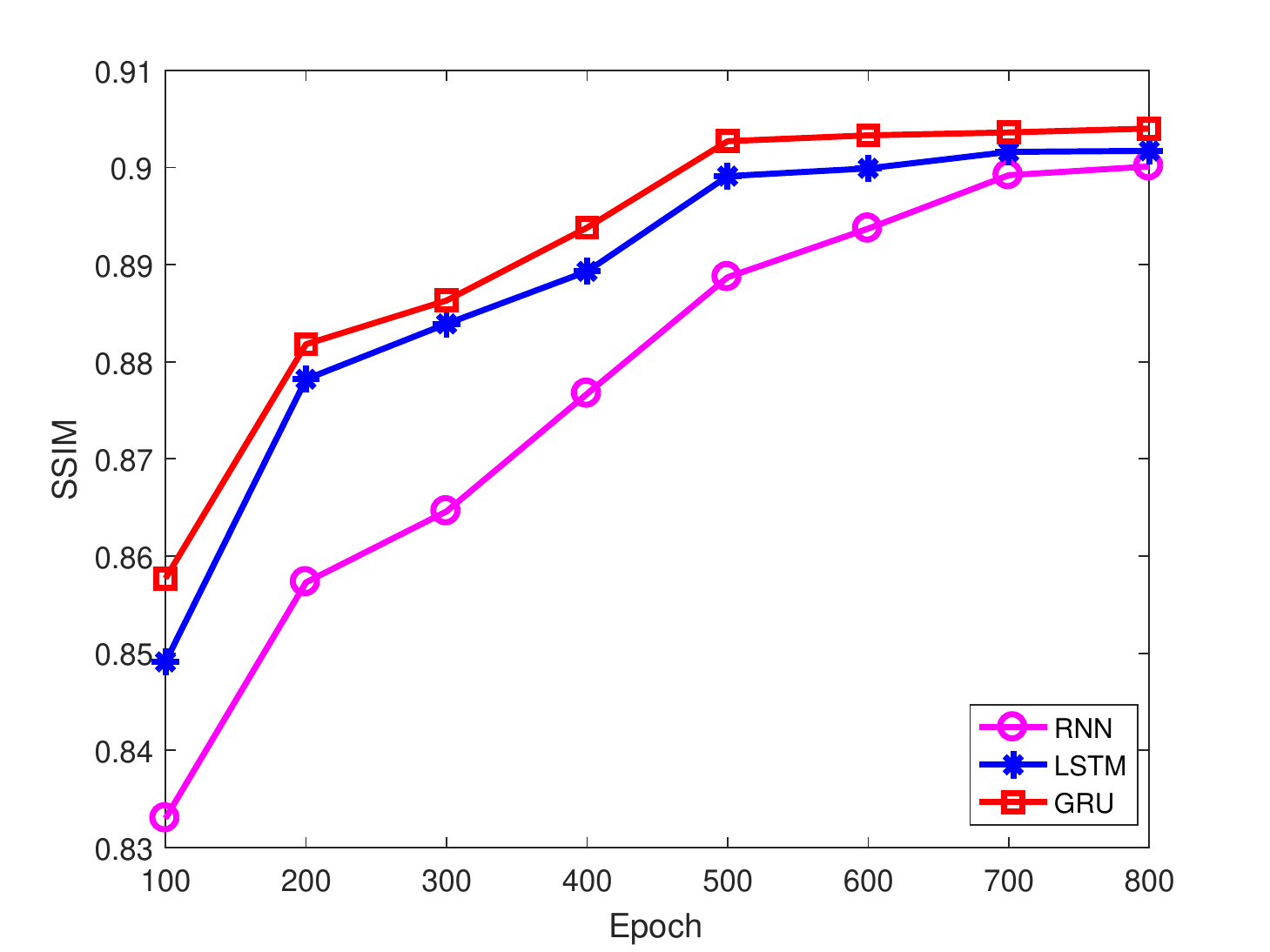}
\\
(a) PSNR  &\hspace{-4mm} (b) SSIM
\\
\end{tabular}
\end{center}
\caption{Curve graph about on recurrent unit.
}
\label{fig:recurrent unit.}
\end{figure}

There are three recurrent units can be selected: ConvRNN~\cite{conv_rnn}, ConvGRU~\cite{conv_GRU} and ConvLSTM~\cite{conv_lstm}.
It needs to explore which the best recurrent unit is.
We conduct three experiments on recurrent units, as shown in Tab.~\ref{tab:recurrent unit.} and Fig.~\ref{fig:recurrent unit.}.
We can see that the ConvGRU performs other recurrent units: ConvLSTM and ConvRNN.
So, we select ConvGRU as our default recurrent unit.
\begin{table}[!h]
\centering
\caption{Analysis on recurrent unit.}
\scalebox{0.99}{
\begin{tabular}{ccccc}
\toprule
Metric &RNN& LSTM &GRU      \\
\midrule
PSNR  & 27.63& 28.43 & 28.81    \\
\midrule
SSIM  &0.9001 &  0.9017 &0.9040      \\

\bottomrule
\end{tabular}}
\label{tab:recurrent unit.}
\end{table}
\subsubsection{Failure Case}\label{sec:Failure Case}
\begin{figure}[!h]
\begin{center}
\begin{tabular}{cc}
\includegraphics[width = 0.45\linewidth]{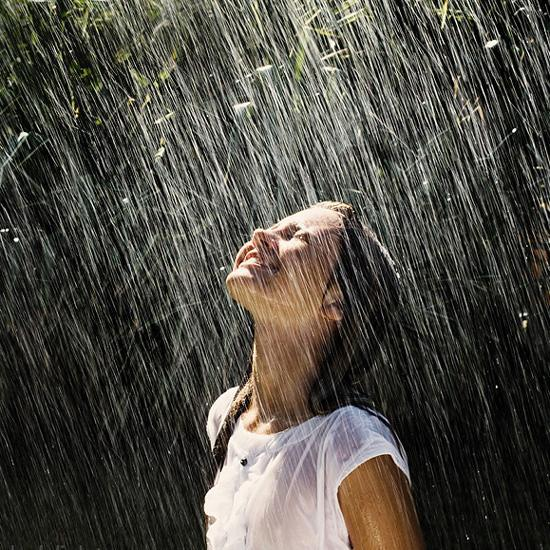} &\hspace{-1mm}
\includegraphics[width = 0.45\linewidth]{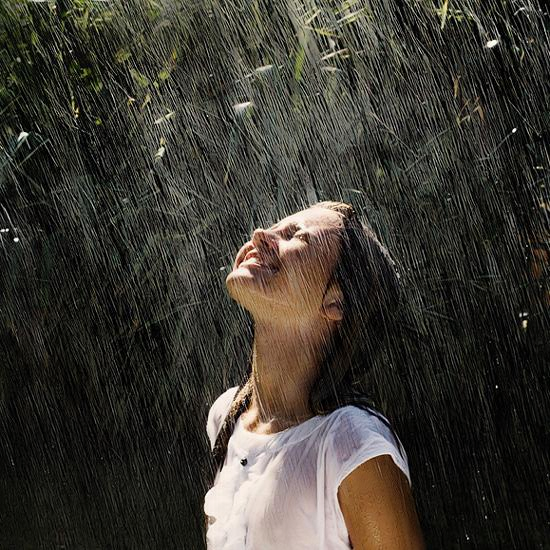}
\\
(a) Input  &\hspace{-4mm} (b) Ours
\\
\end{tabular}
\end{center}
\caption{A failure example.
}
\label{fig:A failure example.}
\end{figure}

We show a failure example in Fig.~\ref{fig:A failure example.}.
We can see that our method can not handle the complex rainy image with heavy rain streaks.
This may be that the samples of the synthetic dataset can not model the real-world image with other rainy condition so that the proposed model trained on synthetic datasets are not robust to real-world rainy images.
This also stimulates us to propose a more robust derainer in the future.
Maybe we can design a trainable model on real-world images to learn different rainy conditions.
\section{Conclusion}
In this paper, we propose a cross-scale fusion network (called DCSFN for short) to solve the image deraining problem.
Different from other multi-scale methods, an inner-scale connection manner has been explored and verified to improve the deraining performance.
Densely connected encoder and decoder with skip connections have been developed to maximize information flow and to enable the computation of long-range spatial dependencies as well as efficient usage of the feature activation of proceeding layers and boost the rain representation of the network.
A cross-scale fusion manner between subnetworks with different scales is proposed to inner-connect and make full use of information at
different scales.
Quantitative and qualitative experimental results demonstrate the superiority of the proposed method compared with several state-of-the-art deraining methods on Rain100H, Rain100L, and Rain1200 datasets.


\bibliographystyle{ACM-Reference-Format}
\bibliography{ijcai20}


\begin{thebibliography}{56}


\ifx \showCODEN    \undefined \def \showCODEN     #1{\unskip}     \fi
\ifx \showDOI      \undefined \def \showDOI       #1{#1}\fi
\ifx \showISBNx    \undefined \def \showISBNx     #1{\unskip}     \fi
\ifx \showISBNxiii \undefined \def \showISBNxiii  #1{\unskip}     \fi
\ifx \showISSN     \undefined \def \showISSN      #1{\unskip}     \fi
\ifx \showLCCN     \undefined \def \showLCCN      #1{\unskip}     \fi
\ifx \shownote     \undefined \def \shownote      #1{#1}          \fi
\ifx \showarticletitle \undefined \def \showarticletitle #1{#1}   \fi
\ifx \showURL      \undefined \def \showURL       {\relax}        \fi
\providecommand\bibfield[2]{#2}
\providecommand\bibinfo[2]{#2}
\providecommand\natexlab[1]{#1}
\providecommand\showeprint[2][]{arXiv:#2}

\bibitem[\protect\citeauthoryear{Cai, Xu, Jia, Qing, and Tao}{Cai
  et~al\mbox{.}}{2016}]%
        {dehaze_dehazenet_cai}
\bibfield{author}{\bibinfo{person}{Bolun Cai}, \bibinfo{person}{Xiangmin Xu},
  \bibinfo{person}{Kui Jia}, \bibinfo{person}{Chunmei Qing}, {and}
  \bibinfo{person}{Dacheng Tao}.} \bibinfo{year}{2016}\natexlab{}.
\newblock \showarticletitle{DehazeNet: An End-to-End System for Single Image
  Haze Removal}.
\newblock \bibinfo{journal}{\emph{{IEEE} TIP}} \bibinfo{volume}{25},
  \bibinfo{number}{11} (\bibinfo{year}{2016}), \bibinfo{pages}{5187--5198}.
\newblock


\bibitem[\protect\citeauthoryear{Chen, Tan, Hou, Chau, and Li}{Chen
  et~al\mbox{.}}{2018a}]%
        {video_derain_chen_cvpr18}
\bibfield{author}{\bibinfo{person}{Jie Chen}, \bibinfo{person}{Cheen{-}Hau
  Tan}, \bibinfo{person}{Junhui Hou}, \bibinfo{person}{Lap{-}Pui Chau}, {and}
  \bibinfo{person}{He Li}.} \bibinfo{year}{2018}\natexlab{a}.
\newblock \showarticletitle{Robust Video Content Alignment and Compensation for
  Rain Removal in a {CNN} Framework}. In \bibinfo{booktitle}{\emph{CVPR}}.
  \bibinfo{pages}{6286--6295}.
\newblock


\bibitem[\protect\citeauthoryear{Chen and Hsu}{Chen and Hsu}{2013}]%
        {derain_lowrank}
\bibfield{author}{\bibinfo{person}{Yi{-}Lei Chen} {and}
  \bibinfo{person}{Chiou{-}Ting Hsu}.} \bibinfo{year}{2013}\natexlab{}.
\newblock \showarticletitle{A Generalized Low-Rank Appearance Model for
  Spatio-temporally Correlated Rain Streaks}. In
  \bibinfo{booktitle}{\emph{ICCV}}. \bibinfo{pages}{1968--1975}.
\newblock


\bibitem[\protect\citeauthoryear{Chen, Wang, Peng, Zhang, Yu, and Sun}{Chen
  et~al\mbox{.}}{2018b}]%
        {pyramid_Pose_Estimation}
\bibfield{author}{\bibinfo{person}{Yilun Chen}, \bibinfo{person}{Zhicheng
  Wang}, \bibinfo{person}{Yuxiang Peng}, \bibinfo{person}{Zhiqiang Zhang},
  \bibinfo{person}{Gang Yu}, {and} \bibinfo{person}{Jian Sun}.}
  \bibinfo{year}{2018}\natexlab{b}.
\newblock \showarticletitle{Cascaded Pyramid Network for Multi-Person Pose
  Estimation}. In \bibinfo{booktitle}{\emph{CVPR}}.
  \bibinfo{pages}{7103--7112}.
\newblock


\bibitem[\protect\citeauthoryear{Cho, van Merrienboer, G{\"{u}}l{\c{c}}ehre,
  Bahdanau, Bougares, Schwenk, and Bengio}{Cho et~al\mbox{.}}{2014}]%
        {conv_GRU}
\bibfield{author}{\bibinfo{person}{Kyunghyun Cho}, \bibinfo{person}{Bart van
  Merrienboer}, \bibinfo{person}{{\c{C}}aglar G{\"{u}}l{\c{c}}ehre},
  \bibinfo{person}{Dzmitry Bahdanau}, \bibinfo{person}{Fethi Bougares},
  \bibinfo{person}{Holger Schwenk}, {and} \bibinfo{person}{Yoshua Bengio}.}
  \bibinfo{year}{2014}\natexlab{}.
\newblock \showarticletitle{Learning Phrase Representations using {RNN}
  Encoder-Decoder for Statistical Machine Translation}. In
  \bibinfo{booktitle}{\emph{EMNLP}}. \bibinfo{pages}{1724--1734}.
\newblock


\bibitem[\protect\citeauthoryear{Cui, Chang, Shan, Zhong, and Chen}{Cui
  et~al\mbox{.}}{2014}]%
        {SISR_cui}
\bibfield{author}{\bibinfo{person}{Zhen Cui}, \bibinfo{person}{Hong Chang},
  \bibinfo{person}{Shiguang Shan}, \bibinfo{person}{Bineng Zhong}, {and}
  \bibinfo{person}{Xilin Chen}.} \bibinfo{year}{2014}\natexlab{}.
\newblock \showarticletitle{Deep Network Cascade for Image Super-resolution}.
  In \bibinfo{booktitle}{\emph{ECCV}}. \bibinfo{pages}{49--64}.
\newblock


\bibitem[\protect\citeauthoryear{Dong, Loy, He, and Tang}{Dong
  et~al\mbox{.}}{2016}]%
        {SISR_dong}
\bibfield{author}{\bibinfo{person}{Chao Dong}, \bibinfo{person}{Chen~Change
  Loy}, \bibinfo{person}{Kaiming He}, {and} \bibinfo{person}{Xiaoou Tang}.}
  \bibinfo{year}{2016}\natexlab{}.
\newblock \showarticletitle{Image Super-Resolution Using Deep Convolutional
  Networks}.
\newblock \bibinfo{journal}{\emph{{IEEE} TPAMI}} \bibinfo{volume}{38},
  \bibinfo{number}{2} (\bibinfo{year}{2016}), \bibinfo{pages}{295--307}.
\newblock


\bibitem[\protect\citeauthoryear{Fan, Wu, Fu, Huang, and Ding}{Fan
  et~al\mbox{.}}{2018}]%
        {derain_GRN}
\bibfield{author}{\bibinfo{person}{Zhiwen Fan}, \bibinfo{person}{Huafeng Wu},
  \bibinfo{person}{Xueyang Fu}, \bibinfo{person}{Yue Huang}, {and}
  \bibinfo{person}{Xinghao Ding}.} \bibinfo{year}{2018}\natexlab{}.
\newblock \showarticletitle{Residual-Guide Network for Single Image Deraining}.
  In \bibinfo{booktitle}{\emph{{ACM} {MM}}}. \bibinfo{pages}{1751--1759}.
\newblock


\bibitem[\protect\citeauthoryear{Fu, Huang, Ding, Liao, and Paisley}{Fu
  et~al\mbox{.}}{2017a}]%
        {derain_clearing_fu}
\bibfield{author}{\bibinfo{person}{Xueyang Fu}, \bibinfo{person}{Jiabin Huang},
  \bibinfo{person}{Xinghao Ding}, \bibinfo{person}{Yinghao Liao}, {and}
  \bibinfo{person}{John Paisley}.} \bibinfo{year}{2017}\natexlab{a}.
\newblock \showarticletitle{Clearing the Skies: A Deep Network Architecture for
  Single-Image Rain Removal}.
\newblock \bibinfo{journal}{\emph{{IEEE} TIP}} \bibinfo{volume}{26},
  \bibinfo{number}{6} (\bibinfo{year}{2017}), \bibinfo{pages}{2944--2956}.
\newblock


\bibitem[\protect\citeauthoryear{Fu, Huang, Zeng, Huang, Ding, and Paisley}{Fu
  et~al\mbox{.}}{2017b}]%
        {derain_ddn_fu}
\bibfield{author}{\bibinfo{person}{Xueyang Fu}, \bibinfo{person}{Jiabin Huang},
  \bibinfo{person}{Delu Zeng}, \bibinfo{person}{Yue Huang},
  \bibinfo{person}{Xinghao Ding}, {and} \bibinfo{person}{John Paisley}.}
  \bibinfo{year}{2017}\natexlab{b}.
\newblock \showarticletitle{Removing Rain from Single Images via a Deep Detail
  Network}. In \bibinfo{booktitle}{\emph{CVPR}}. \bibinfo{pages}{1715--1723}.
\newblock


\bibitem[\protect\citeauthoryear{{Fu}, {Liang}, {Huang}, {Ding}, and
  {Paisley}}{{Fu} et~al\mbox{.}}{2019}]%
        {light_weight}
\bibfield{author}{\bibinfo{person}{X. {Fu}}, \bibinfo{person}{B. {Liang}},
  \bibinfo{person}{Y. {Huang}}, \bibinfo{person}{X. {Ding}}, {and}
  \bibinfo{person}{J. {Paisley}}.} \bibinfo{year}{2019}\natexlab{}.
\newblock \showarticletitle{Lightweight Pyramid Networks for Image Deraining}.
\newblock \bibinfo{journal}{\emph{{IEEE} TNNLS}} (\bibinfo{year}{2019}),
  \bibinfo{pages}{1--14}.
\newblock


\bibitem[\protect\citeauthoryear{Fu, Qi, Huang, Ding, Wu, and Paisley}{Fu
  et~al\mbox{.}}{2018}]%
        {derain_fu_deeptree}
\bibfield{author}{\bibinfo{person}{Xueyang Fu}, \bibinfo{person}{Qi Qi},
  \bibinfo{person}{Yue Huang}, \bibinfo{person}{Xinghao Ding},
  \bibinfo{person}{Feng Wu}, {and} \bibinfo{person}{John~W. Paisley}.}
  \bibinfo{year}{2018}\natexlab{}.
\newblock \showarticletitle{A Deep Tree-Structured Fusion Model for Single
  Image Deraining}.
\newblock \bibinfo{journal}{\emph{CoRR}}  \bibinfo{volume}{abs/1811.08632}
  (\bibinfo{year}{2018}).
\newblock


\bibitem[\protect\citeauthoryear{Hu, Shen, and Sun}{Hu et~al\mbox{.}}{2018}]%
        {se}
\bibfield{author}{\bibinfo{person}{Jie Hu}, \bibinfo{person}{Li Shen}, {and}
  \bibinfo{person}{Gang Sun}.} \bibinfo{year}{2018}\natexlab{}.
\newblock \showarticletitle{Squeeze-and-Excitation Networks}. In
  \bibinfo{booktitle}{\emph{CVPR}}. \bibinfo{pages}{7132--7141}.
\newblock


\bibitem[\protect\citeauthoryear{Huang, Liu, van~der Maaten, and
  Weinberger}{Huang et~al\mbox{.}}{2017}]%
        {Densenetwork_huang}
\bibfield{author}{\bibinfo{person}{Gao Huang}, \bibinfo{person}{Zhuang Liu},
  \bibinfo{person}{Laurens van~der Maaten}, {and} \bibinfo{person}{Kilian~Q.
  Weinberger}.} \bibinfo{year}{2017}\natexlab{}.
\newblock \showarticletitle{Densely Connected Convolutional Networks}. In
  \bibinfo{booktitle}{\emph{CVPR}}. \bibinfo{pages}{2261--2269}.
\newblock


\bibitem[\protect\citeauthoryear{Huynh-Thu and Ghanbari}{Huynh-Thu and
  Ghanbari}{2008}]%
        {PSNR_thu}
\bibfield{author}{\bibinfo{person}{Q. Huynh-Thu} {and} \bibinfo{person}{M.
  Ghanbari}.} \bibinfo{year}{2008}\natexlab{}.
\newblock \showarticletitle{Scope of validity of PSNR in image/video quality
  assessment}.
\newblock \bibinfo{journal}{\emph{Electronics Letters}} \bibinfo{volume}{44},
  \bibinfo{number}{13} (\bibinfo{year}{2008}), \bibinfo{pages}{800--801}.
\newblock


\bibitem[\protect\citeauthoryear{Kang, Lin, and Fu}{Kang et~al\mbox{.}}{2012}]%
        {derain_id_kang}
\bibfield{author}{\bibinfo{person}{Li{-}Wei Kang}, \bibinfo{person}{Chia{-}Wen
  Lin}, {and} \bibinfo{person}{Yu{-}Hsiang Fu}.}
  \bibinfo{year}{2012}\natexlab{}.
\newblock \showarticletitle{Automatic Single-Image-Based Rain Streaks Removal
  via Image Decomposition}.
\newblock \bibinfo{journal}{\emph{{IEEE} TIP}} \bibinfo{volume}{21},
  \bibinfo{number}{4} (\bibinfo{year}{2012}), \bibinfo{pages}{1742--1755}.
\newblock


\bibitem[\protect\citeauthoryear{Kim, Lee, Sim, and Kim}{Kim
  et~al\mbox{.}}{2013}]%
        {derain_nonlocalfilter_kim}
\bibfield{author}{\bibinfo{person}{Jin{-}Hwan Kim}, \bibinfo{person}{Chul Lee},
  \bibinfo{person}{Jae{-}Young Sim}, {and} \bibinfo{person}{Chang{-}Su Kim}.}
  \bibinfo{year}{2013}\natexlab{}.
\newblock \showarticletitle{Single-image deraining using an adaptive nonlocal
  means filter}. In \bibinfo{booktitle}{\emph{ICIP}}.
  \bibinfo{pages}{914--917}.
\newblock


\bibitem[\protect\citeauthoryear{Kingma and Ba}{Kingma and Ba}{2015}]%
        {adam}
\bibfield{author}{\bibinfo{person}{Diederik~P. Kingma} {and}
  \bibinfo{person}{Jimmy Ba}.} \bibinfo{year}{2015}\natexlab{}.
\newblock \showarticletitle{Adam: {A} Method for Stochastic Optimization}. In
  \bibinfo{booktitle}{\emph{ICLR}}.
\newblock


\bibitem[\protect\citeauthoryear{Lai, Huang, Ahuja, and Yang}{Lai
  et~al\mbox{.}}{2019}]%
        {pyramid_superresolution}
\bibfield{author}{\bibinfo{person}{Wei{-}Sheng Lai}, \bibinfo{person}{Jia{-}Bin
  Huang}, \bibinfo{person}{Narendra Ahuja}, {and} \bibinfo{person}{Ming{-}Hsuan
  Yang}.} \bibinfo{year}{2019}\natexlab{}.
\newblock \showarticletitle{Fast and Accurate Image Super-Resolution with Deep
  Laplacian Pyramid Networks}.
\newblock \bibinfo{journal}{\emph{{IEEE} Trans. Pattern Anal. Mach. Intell.}}
  \bibinfo{volume}{41}, \bibinfo{number}{11} (\bibinfo{year}{2019}),
  \bibinfo{pages}{2599--2613}.
\newblock


\bibitem[\protect\citeauthoryear{Li, Peng, Wang, Xu, and Feng}{Li
  et~al\mbox{.}}{2017}]%
        {dehaze_li_AOD}
\bibfield{author}{\bibinfo{person}{Boyi Li}, \bibinfo{person}{Xiulian Peng},
  \bibinfo{person}{Zhangyang Wang}, \bibinfo{person}{Jizheng Xu}, {and}
  \bibinfo{person}{Dan Feng}.} \bibinfo{year}{2017}\natexlab{}.
\newblock \showarticletitle{AOD-Net: All-in-One Dehazing Network}. In
  \bibinfo{booktitle}{\emph{ICCV}}. \bibinfo{pages}{4780--4788}.
\newblock


\bibitem[\protect\citeauthoryear{Li, He, Zhang, Chang, Dong, and Lin}{Li
  et~al\mbox{.}}{2018a}]%
        {derain_nledn_li}
\bibfield{author}{\bibinfo{person}{Guanbin Li}, \bibinfo{person}{Xiang He},
  \bibinfo{person}{Wei Zhang}, \bibinfo{person}{Huiyou Chang},
  \bibinfo{person}{Le Dong}, {and} \bibinfo{person}{Liang Lin}.}
  \bibinfo{year}{2018}\natexlab{a}.
\newblock \showarticletitle{Non-locally Enhanced Encoder-Decoder Network for
  Single Image De-raining}. In \bibinfo{booktitle}{\emph{{ACM} {MM}}}.
  \bibinfo{pages}{1056--1064}.
\newblock


\bibitem[\protect\citeauthoryear{Li, Xie, Zhao, Wei, Gu, Tao, and Meng}{Li
  et~al\mbox{.}}{2018c}]%
        {video_derain_cvpr18_li}
\bibfield{author}{\bibinfo{person}{Minghan Li}, \bibinfo{person}{Qi Xie},
  \bibinfo{person}{Qian Zhao}, \bibinfo{person}{Wei Wei},
  \bibinfo{person}{Shuhang Gu}, \bibinfo{person}{Jing Tao}, {and}
  \bibinfo{person}{Deyu Meng}.} \bibinfo{year}{2018}\natexlab{c}.
\newblock \showarticletitle{Video Rain Streak Removal by Multiscale
  Convolutional Sparse Coding}. In \bibinfo{booktitle}{\emph{CVPR}}.
  \bibinfo{pages}{6644--6653}.
\newblock


\bibitem[\protect\citeauthoryear{Li, Cheong, and Tan}{Li
  et~al\mbox{.}}{2019b}]%
        {derain_Heavy_Li_2019_CVPR}
\bibfield{author}{\bibinfo{person}{Ruoteng Li}, \bibinfo{person}{Loong{-}Fah
  Cheong}, {and} \bibinfo{person}{Robby~T. Tan}.}
  \bibinfo{year}{2019}\natexlab{b}.
\newblock \showarticletitle{Heavy Rain Image Restoration: Integrating Physics
  Model and Conditional Adversarial Learning}. In
  \bibinfo{booktitle}{\emph{CVPR}}. \bibinfo{pages}{1633--1642}.
\newblock


\bibitem[\protect\citeauthoryear{Li, Araujo, Ren, Wang, Tokuda, Junior,
  Cesar{-}Junior, Zhang, Guo, and Cao}{Li et~al\mbox{.}}{2019a}]%
        {derain_Comprehensive_Benchmark_Li_2019_CVPR}
\bibfield{author}{\bibinfo{person}{Siyuan Li}, \bibinfo{person}{Iago~Breno
  Araujo}, \bibinfo{person}{Wenqi Ren}, \bibinfo{person}{Zhangyang Wang},
  \bibinfo{person}{Eric~K. Tokuda}, \bibinfo{person}{Roberto~Hirata Junior},
  \bibinfo{person}{Roberto Cesar{-}Junior}, \bibinfo{person}{Jiawan Zhang},
  \bibinfo{person}{Xiaojie Guo}, {and} \bibinfo{person}{Xiaochun Cao}.}
  \bibinfo{year}{2019}\natexlab{a}.
\newblock \showarticletitle{Single Image Deraining: {A} Comprehensive Benchmark
  Analysis}. In \bibinfo{booktitle}{\emph{CVPR}}. \bibinfo{pages}{3838--3847}.
\newblock


\bibitem[\protect\citeauthoryear{Li, Wu, Lin, Liu, and Zha}{Li
  et~al\mbox{.}}{2018b}]%
        {derain_rescan_li}
\bibfield{author}{\bibinfo{person}{Xia Li}, \bibinfo{person}{Jianlong Wu},
  \bibinfo{person}{Zhouchen Lin}, \bibinfo{person}{Hong Liu}, {and}
  \bibinfo{person}{Hongbin Zha}.} \bibinfo{year}{2018}\natexlab{b}.
\newblock \showarticletitle{Recurrent Squeeze-and-Excitation Context
  Aggregation Net for Single Image Deraining}. In
  \bibinfo{booktitle}{\emph{ECCV}}. \bibinfo{pages}{262--277}.
\newblock


\bibitem[\protect\citeauthoryear{Li, Tan, Guo, Lu, and Brown}{Li
  et~al\mbox{.}}{2016}]%
        {derain_lp_li}
\bibfield{author}{\bibinfo{person}{Yu Li}, \bibinfo{person}{Robby~T. Tan},
  \bibinfo{person}{Xiaojie Guo}, \bibinfo{person}{Jiangbo Lu}, {and}
  \bibinfo{person}{Michael~S. Brown}.} \bibinfo{year}{2016}\natexlab{}.
\newblock \showarticletitle{Rain Streak Removal Using Layer Priors}. In
  \bibinfo{booktitle}{\emph{CVPR}}. \bibinfo{pages}{2736--2744}.
\newblock


\bibitem[\protect\citeauthoryear{Lin, Doll{\'{a}}r, Girshick, He, Hariharan,
  and Belongie}{Lin et~al\mbox{.}}{2017}]%
        {feature_pyramid_objectdetection}
\bibfield{author}{\bibinfo{person}{Tsung{-}Yi Lin}, \bibinfo{person}{Piotr
  Doll{\'{a}}r}, \bibinfo{person}{Ross~B. Girshick}, \bibinfo{person}{Kaiming
  He}, \bibinfo{person}{Bharath Hariharan}, {and} \bibinfo{person}{Serge~J.
  Belongie}.} \bibinfo{year}{2017}\natexlab{}.
\newblock \showarticletitle{Feature Pyramid Networks for Object Detection}. In
  \bibinfo{booktitle}{\emph{CVPR}}. \bibinfo{pages}{936--944}.
\newblock


\bibitem[\protect\citeauthoryear{Liu, Yang, Yang, and Guo}{Liu
  et~al\mbox{.}}{2018}]%
        {video_derain_cvpr18_liu}
\bibfield{author}{\bibinfo{person}{Jiaying Liu}, \bibinfo{person}{Wenhan Yang},
  \bibinfo{person}{Shuai Yang}, {and} \bibinfo{person}{Zongming Guo}.}
  \bibinfo{year}{2018}\natexlab{}.
\newblock \showarticletitle{Erase or Fill? Deep Joint Recurrent Rain Removal
  and Reconstruction in Videos}. In \bibinfo{booktitle}{\emph{CVPR}}.
  \bibinfo{pages}{3233--3242}.
\newblock


\bibitem[\protect\citeauthoryear{Liu, Wu, Li, Zhang, and Wu}{Liu
  et~al\mbox{.}}{2020}]%
        {multi-scale-face}
\bibfield{author}{\bibinfo{person}{Zhilei Liu}, \bibinfo{person}{Yunpeng Wu},
  \bibinfo{person}{Le Li}, \bibinfo{person}{Cuicui Zhang}, {and}
  \bibinfo{person}{Baoyuan Wu}.} \bibinfo{year}{2020}\natexlab{}.
\newblock \showarticletitle{Joint Face Completion and Super-resolution using
  Multi-scale Feature Relation Learning}.
\newblock \bibinfo{journal}{\emph{CoRR}}  \bibinfo{volume}{abs/2003.00255}
  (\bibinfo{year}{2020}).
\newblock
\showeprint[arxiv]{2003.00255}
\urldef\tempurl%
\url{https://arxiv.org/abs/2003.00255}
\showURL{%
\tempurl}


\bibitem[\protect\citeauthoryear{Long, Shelhamer, and Darrell}{Long
  et~al\mbox{.}}{2015}]%
        {semanticsegmentation_fcn}
\bibfield{author}{\bibinfo{person}{Jonathan Long}, \bibinfo{person}{Evan
  Shelhamer}, {and} \bibinfo{person}{Trevor Darrell}.}
  \bibinfo{year}{2015}\natexlab{}.
\newblock \showarticletitle{Fully convolutional networks for semantic
  segmentation}. In \bibinfo{booktitle}{\emph{CVPR}}.
  \bibinfo{pages}{3431--3440}.
\newblock


\bibitem[\protect\citeauthoryear{Luo, Xu, and Ji}{Luo et~al\mbox{.}}{2015}]%
        {derain_dsc_luo}
\bibfield{author}{\bibinfo{person}{Yu Luo}, \bibinfo{person}{Yong Xu}, {and}
  \bibinfo{person}{Hui Ji}.} \bibinfo{year}{2015}\natexlab{}.
\newblock \showarticletitle{Removing Rain from a Single Image via
  Discriminative Sparse Coding}. In \bibinfo{booktitle}{\emph{ICCV}}.
  \bibinfo{pages}{3397--3405}.
\newblock


\bibitem[\protect\citeauthoryear{Mandic and Chambers}{Mandic and
  Chambers}{2001}]%
        {conv_rnn}
\bibfield{author}{\bibinfo{person}{Danilo~P. Mandic} {and}
  \bibinfo{person}{Jonathon Chambers}.} \bibinfo{year}{2001}\natexlab{}.
\newblock \showarticletitle{Recurrent Neural Networks for Prediction: Learning
  Algorithms,Architectures and Stability}.
\newblock \bibinfo{journal}{\emph{Adaptive Learning Systems for Signal
  Processing Communications Control}} (\bibinfo{year}{2001}).
\newblock


\bibitem[\protect\citeauthoryear{Mustaniemi, Kannala, S{\"{a}}rkk{\"{a}},
  Matas, and Heikkil{\"{a}}}{Mustaniemi et~al\mbox{.}}{2018}]%
        {deblur_Mustaniemi}
\bibfield{author}{\bibinfo{person}{Janne Mustaniemi}, \bibinfo{person}{Juho
  Kannala}, \bibinfo{person}{Simo S{\"{a}}rkk{\"{a}}}, \bibinfo{person}{Jiri
  Matas}, {and} \bibinfo{person}{Janne Heikkil{\"{a}}}.}
  \bibinfo{year}{2018}\natexlab{}.
\newblock \showarticletitle{Inertial-aided Motion Deblurring with Deep
  Networks}. In \bibinfo{booktitle}{\emph{CoRR}},
  Vol.~\bibinfo{volume}{abs/1810.00986}.
\newblock
\showeprint[arxiv]{1810.00986}


\bibitem[\protect\citeauthoryear{Pan, Liu, Sun, Zhang, Liu, Ren, Li, Tang, Lu,
  Tai, and Yang}{Pan et~al\mbox{.}}{2018b}]%
        {dual_cnn}
\bibfield{author}{\bibinfo{person}{Jinshan Pan}, \bibinfo{person}{Sifei Liu},
  \bibinfo{person}{Deqing Sun}, \bibinfo{person}{Jiawei Zhang},
  \bibinfo{person}{Yang Liu}, \bibinfo{person}{Jimmy S.~J. Ren},
  \bibinfo{person}{Zechao Li}, \bibinfo{person}{Jinhui Tang},
  \bibinfo{person}{Huchuan Lu}, \bibinfo{person}{Yu{-}Wing Tai}, {and}
  \bibinfo{person}{Ming{-}Hsuan Yang}.} \bibinfo{year}{2018}\natexlab{b}.
\newblock \showarticletitle{Learning Dual Convolutional Neural Networks for
  Low-Level Vision}. In \bibinfo{booktitle}{\emph{CVPR}}.
  \bibinfo{pages}{3070--3079}.
\newblock


\bibitem[\protect\citeauthoryear{Pan, Liu, Dong, Zhang, Ren, Tang, Tai, and
  Yang}{Pan et~al\mbox{.}}{2018a}]%
        {derain_Physics_gan}
\bibfield{author}{\bibinfo{person}{Jinshan Pan}, \bibinfo{person}{Yang Liu},
  \bibinfo{person}{Jiangxin Dong}, \bibinfo{person}{Jiawei Zhang},
  \bibinfo{person}{Jimmy S.~J. Ren}, \bibinfo{person}{Jinhui Tang},
  \bibinfo{person}{Yu{-}Wing Tai}, {and} \bibinfo{person}{Ming{-}Hsuan Yang}.}
  \bibinfo{year}{2018}\natexlab{a}.
\newblock \showarticletitle{Physics-Based Generative Adversarial Models for
  Image Restoration and Beyond}.
\newblock \bibinfo{journal}{\emph{CoRR}}  \bibinfo{volume}{abs/1808.00605}
  (\bibinfo{year}{2018}).
\newblock
\showeprint[arxiv]{1808.00605}


\bibitem[\protect\citeauthoryear{Ranjan and Black}{Ranjan and Black}{2017}]%
        {Pyramid_Ranjan_Optical_Flow}
\bibfield{author}{\bibinfo{person}{Anurag Ranjan} {and}
  \bibinfo{person}{Michael~J. Black}.} \bibinfo{year}{2017}\natexlab{}.
\newblock \showarticletitle{Optical Flow Estimation Using a Spatial Pyramid
  Network}. In \bibinfo{booktitle}{\emph{CVPR}}. \bibinfo{pages}{2720--2729}.
\newblock


\bibitem[\protect\citeauthoryear{Ren, Zuo, Hu, Zhu, and Meng}{Ren
  et~al\mbox{.}}{2019}]%
        {derain_prenet_Ren_2019_CVPR}
\bibfield{author}{\bibinfo{person}{Dongwei Ren}, \bibinfo{person}{Wangmeng
  Zuo}, \bibinfo{person}{Qinghua Hu}, \bibinfo{person}{Pengfei Zhu}, {and}
  \bibinfo{person}{Deyu Meng}.} \bibinfo{year}{2019}\natexlab{}.
\newblock \showarticletitle{Progressive Image Deraining Networks: {A} Better
  and Simpler Baseline}. In \bibinfo{booktitle}{\emph{CVPR}}.
  \bibinfo{pages}{3937--3946}.
\newblock


\bibitem[\protect\citeauthoryear{Ren, Liu, Zhang, Pan, Cao, and Yang}{Ren
  et~al\mbox{.}}{2016}]%
        {dehaze_mscnn_ren}
\bibfield{author}{\bibinfo{person}{Wenqi Ren}, \bibinfo{person}{Si Liu},
  \bibinfo{person}{Hua Zhang}, \bibinfo{person}{Jin{-}shan Pan},
  \bibinfo{person}{Xiaochun Cao}, {and} \bibinfo{person}{Ming{-}Hsuan Yang}.}
  \bibinfo{year}{2016}\natexlab{}.
\newblock \showarticletitle{Single Image Dehazing via Multi-scale Convolutional
  Neural Networks}. In \bibinfo{booktitle}{\emph{ECCV}}.
  \bibinfo{pages}{154--169}.
\newblock


\bibitem[\protect\citeauthoryear{Wang, Yang, Xu, Chen, Zhang, and Lau}{Wang
  et~al\mbox{.}}{2019}]%
        {derain_2019_CVPR_spa}
\bibfield{author}{\bibinfo{person}{Tianyu Wang}, \bibinfo{person}{Xin Yang},
  \bibinfo{person}{Ke Xu}, \bibinfo{person}{Shaozhe Chen},
  \bibinfo{person}{Qiang Zhang}, {and} \bibinfo{person}{Rynson W.~H. Lau}.}
  \bibinfo{year}{2019}\natexlab{}.
\newblock \showarticletitle{Spatial Attentive Single-Image Deraining With a
  High Quality Real Rain Dataset}. In \bibinfo{booktitle}{\emph{CVPR}}.
  \bibinfo{pages}{12270--12279}.
\newblock


\bibitem[\protect\citeauthoryear{Wang, Girshick, Gupta, and He}{Wang
  et~al\mbox{.}}{2017a}]%
        {nonlocalnetwork_wang}
\bibfield{author}{\bibinfo{person}{Xiaolong Wang}, \bibinfo{person}{Ross~B.
  Girshick}, \bibinfo{person}{Abhinav Gupta}, {and} \bibinfo{person}{Kaiming
  He}.} \bibinfo{year}{2017}\natexlab{a}.
\newblock \showarticletitle{Non-local Neural Networks}. In
  \bibinfo{booktitle}{\emph{CoRR}}, Vol.~\bibinfo{volume}{abs/1711.07971}.
\newblock
\showeprint[arxiv]{1711.07971}


\bibitem[\protect\citeauthoryear{Wang, Shrivastava, and Gupta}{Wang
  et~al\mbox{.}}{2017c}]%
        {objectdetection_wang}
\bibfield{author}{\bibinfo{person}{Xiaolong Wang}, \bibinfo{person}{Abhinav
  Shrivastava}, {and} \bibinfo{person}{Abhinav Gupta}.}
  \bibinfo{year}{2017}\natexlab{c}.
\newblock \showarticletitle{A-Fast-RCNN: Hard Positive Generation via Adversary
  for Object Detection}. In \bibinfo{booktitle}{\emph{CVPR}}.
  \bibinfo{pages}{3039--3048}.
\newblock


\bibitem[\protect\citeauthoryear{Wang, Liu, Chen, and Zeng}{Wang
  et~al\mbox{.}}{2017b}]%
        {derain_Wang_Hierarchical}
\bibfield{author}{\bibinfo{person}{Yinglong Wang}, \bibinfo{person}{Shuaicheng
  Liu}, \bibinfo{person}{Chen Chen}, {and} \bibinfo{person}{Bing Zeng}.}
  \bibinfo{year}{2017}\natexlab{b}.
\newblock \showarticletitle{A Hierarchical Approach for Rain or Snow Removing
  in a Single Color Image}.
\newblock \bibinfo{journal}{\emph{{IEEE} TIP}} \bibinfo{volume}{26},
  \bibinfo{number}{8} (\bibinfo{year}{2017}), \bibinfo{pages}{3936--3950}.
\newblock


\bibitem[\protect\citeauthoryear{Wang, Bovik, Sheikh, and Simoncelli}{Wang
  et~al\mbox{.}}{2004}]%
        {SSIM_wang}
\bibfield{author}{\bibinfo{person}{Zhou Wang}, \bibinfo{person}{Alan~C. Bovik},
  \bibinfo{person}{Hamid~R. Sheikh}, {and} \bibinfo{person}{Eero~P.
  Simoncelli}.} \bibinfo{year}{2004}\natexlab{}.
\newblock \showarticletitle{Image quality assessment: from error visibility to
  structural similarity}.
\newblock \bibinfo{journal}{\emph{{IEEE} TIP}} \bibinfo{volume}{13},
  \bibinfo{number}{4} (\bibinfo{year}{2004}), \bibinfo{pages}{600--612}.
\newblock


\bibitem[\protect\citeauthoryear{Wei, Meng, Zhao, Xu, and Wu}{Wei
  et~al\mbox{.}}{2019}]%
        {Derain-cvpr19-semi}
\bibfield{author}{\bibinfo{person}{Wei Wei}, \bibinfo{person}{Deyu Meng},
  \bibinfo{person}{Qian Zhao}, \bibinfo{person}{Zongben Xu}, {and}
  \bibinfo{person}{Ying Wu}.} \bibinfo{year}{2019}\natexlab{}.
\newblock \showarticletitle{Semi-Supervised Transfer Learning for Image Rain
  Removal}. In \bibinfo{booktitle}{\emph{CVPR}}. \bibinfo{pages}{3877--3886}.
\newblock


\bibitem[\protect\citeauthoryear{Yang, Liu, and Feng}{Yang
  et~al\mbox{.}}{2019}]%
        {video_derain_Yang_2019_CVPR}
\bibfield{author}{\bibinfo{person}{Wenhan Yang}, \bibinfo{person}{Jiaying Liu},
  {and} \bibinfo{person}{Jiashi Feng}.} \bibinfo{year}{2019}\natexlab{}.
\newblock \showarticletitle{Frame-Consistent Recurrent Video Deraining With
  Dual-Level Flow}. In \bibinfo{booktitle}{\emph{CVPR}}.
\newblock


\bibitem[\protect\citeauthoryear{Yang, Tan, Feng, Liu, Guo, and Yan}{Yang
  et~al\mbox{.}}{2017}]%
        {derain_jorder_yang}
\bibfield{author}{\bibinfo{person}{Wenhan Yang}, \bibinfo{person}{Robby~T.
  Tan}, \bibinfo{person}{Jiashi Feng}, \bibinfo{person}{Jiaying Liu},
  \bibinfo{person}{Zongming Guo}, {and} \bibinfo{person}{Shuicheng Yan}.}
  \bibinfo{year}{2017}\natexlab{}.
\newblock \showarticletitle{Deep Joint Rain Detection and Removal from a Single
  Image}. In \bibinfo{booktitle}{\emph{CVPR}}. \bibinfo{pages}{1685--1694}.
\newblock


\bibitem[\protect\citeauthoryear{{Yang}, {Tan}, {Feng}, {Liu}, {Yan}, and
  {Guo}}{{Yang} et~al\mbox{.}}{2019}]%
        {derain_pami_yang}
\bibfield{author}{\bibinfo{person}{W. {Yang}}, \bibinfo{person}{R.~T. {Tan}},
  \bibinfo{person}{J. {Feng}}, \bibinfo{person}{J. {Liu}}, \bibinfo{person}{S.
  {Yan}}, {and} \bibinfo{person}{Z. {Guo}}.} \bibinfo{year}{2019}\natexlab{}.
\newblock \showarticletitle{Joint Rain Detection and Removal from a Single
  Image with Contextualized Deep Networks}.
\newblock \bibinfo{journal}{\emph{{IEEE} TPAMI}} (\bibinfo{year}{2019}).
\newblock


\bibitem[\protect\citeauthoryear{Yang and Lu}{Yang and Lu}{2019}]%
        {derain-acmmm19-rehen}
\bibfield{author}{\bibinfo{person}{Youzhao Yang} {and} \bibinfo{person}{Hong
  Lu}.} \bibinfo{year}{2019}\natexlab{}.
\newblock \showarticletitle{Single Image Deraining via Recurrent Hierarchy
  Enhancement Network}. In \bibinfo{booktitle}{\emph{{ACM} {MM}}}.
  \bibinfo{pages}{1814--1822}.
\newblock


\bibitem[\protect\citeauthoryear{Yu, Fan, Yang, Xu, Wang, Wang, and Huang}{Yu
  et~al\mbox{.}}{2018}]%
        {SISR_yu}
\bibfield{author}{\bibinfo{person}{Jiahui Yu}, \bibinfo{person}{Yuchen Fan},
  \bibinfo{person}{Jianchao Yang}, \bibinfo{person}{Ning Xu},
  \bibinfo{person}{Zhaowen Wang}, \bibinfo{person}{Xinchao Wang}, {and}
  \bibinfo{person}{Thomas Huang}.} \bibinfo{year}{2018}\natexlab{}.
\newblock \showarticletitle{Wide Activation for Efficient and Accurate Image
  Super-Resolution}.
\newblock \bibinfo{journal}{\emph{CoRR}}  \bibinfo{volume}{abs/1808.08718}.
\newblock
\showeprint[arxiv]{1808.08718}


\bibitem[\protect\citeauthoryear{Zaremba, Sutskever, and Vinyals}{Zaremba
  et~al\mbox{.}}{2014}]%
        {conv_lstm}
\bibfield{author}{\bibinfo{person}{Wojciech Zaremba}, \bibinfo{person}{Ilya
  Sutskever}, {and} \bibinfo{person}{Oriol Vinyals}.}
  \bibinfo{year}{2014}\natexlab{}.
\newblock \showarticletitle{Recurrent Neural Network Regularization}.
\newblock \bibinfo{journal}{\emph{CoRR}}  \bibinfo{volume}{abs/1409.2329}
  (\bibinfo{year}{2014}).
\newblock
\showeprint[arxiv]{1409.2329}


\bibitem[\protect\citeauthoryear{Zhang and Patel}{Zhang and Patel}{2017}]%
        {derain_zhang_Sparse_and_Low-Rank}
\bibfield{author}{\bibinfo{person}{He Zhang} {and} \bibinfo{person}{Vishal~M.
  Patel}.} \bibinfo{year}{2017}\natexlab{}.
\newblock \showarticletitle{Convolutional Sparse and Low-Rank Coding-Based Rain
  Streak Removal}. In \bibinfo{booktitle}{\emph{WACV}}.
  \bibinfo{pages}{1259--1267}.
\newblock


\bibitem[\protect\citeauthoryear{Zhang and Patel}{Zhang and Patel}{2018a}]%
        {dehaze_zhang_dcdpn}
\bibfield{author}{\bibinfo{person}{He Zhang} {and} \bibinfo{person}{Vishal~M.
  Patel}.} \bibinfo{year}{2018}\natexlab{a}.
\newblock \showarticletitle{Densely Connected Pyramid Dehazing Network}. In
  \bibinfo{booktitle}{\emph{CVPR}}. \bibinfo{pages}{3194--3203}.
\newblock


\bibitem[\protect\citeauthoryear{Zhang and Patel}{Zhang and Patel}{2018b}]%
        {derain_zhang_did}
\bibfield{author}{\bibinfo{person}{He Zhang} {and} \bibinfo{person}{Vishal~M.
  Patel}.} \bibinfo{year}{2018}\natexlab{b}.
\newblock \showarticletitle{Density-Aware Single Image De-Raining Using a
  Multi-Stream Dense Network}. In \bibinfo{booktitle}{\emph{CVPR}}.
  \bibinfo{pages}{695--704}.
\newblock


\bibitem[\protect\citeauthoryear{{Zhang}, {Sindagi}, and {Patel}}{{Zhang}
  et~al\mbox{.}}{2019}]%
        {derain_cgan_zhang}
\bibfield{author}{\bibinfo{person}{H. {Zhang}}, \bibinfo{person}{V. {Sindagi}},
  {and} \bibinfo{person}{V.~M. {Patel}}.} \bibinfo{year}{2019}\natexlab{}.
\newblock \showarticletitle{Image De-raining Using a Conditional Generative
  Adversarial Network}.
\newblock \bibinfo{journal}{\emph{{IEEE} TCSVT}} (\bibinfo{year}{2019}).
\newblock


\bibitem[\protect\citeauthoryear{Zhang, Wang, Qi, Wang, Feng, and Lu}{Zhang
  et~al\mbox{.}}{2018}]%
        {tracking_zhang}
\bibfield{author}{\bibinfo{person}{Yunhua Zhang}, \bibinfo{person}{Lijun Wang},
  \bibinfo{person}{Jinqing Qi}, \bibinfo{person}{Dong Wang},
  \bibinfo{person}{Mengyang Feng}, {and} \bibinfo{person}{Huchuan Lu}.}
  \bibinfo{year}{2018}\natexlab{}.
\newblock \showarticletitle{Structured Siamese Network for Real-Time Visual
  Tracking}. In \bibinfo{booktitle}{\emph{ECCV}}. \bibinfo{pages}{355--370}.
\newblock


\bibitem[\protect\citeauthoryear{Zhu, Fu, Lischinski, and Heng}{Zhu
  et~al\mbox{.}}{2017}]%
        {derain_zhu_bilayer}
\bibfield{author}{\bibinfo{person}{Lei Zhu}, \bibinfo{person}{Chi{-}Wing Fu},
  \bibinfo{person}{Dani Lischinski}, {and} \bibinfo{person}{Pheng{-}Ann Heng}.}
  \bibinfo{year}{2017}\natexlab{}.
\newblock \showarticletitle{Joint Bi-layer Optimization for Single-Image Rain
  Streak Removal}. In \bibinfo{booktitle}{\emph{ICCV}}.
  \bibinfo{pages}{2545--2553}.
\newblock


\end{thebibliography}

%
%
%
%
%
%
%
%

\end{document}